\documentclass[conference]{IEEEtran}
\usepackage{amsmath,amssymb,amsfonts,graphicx,amsthm,nicefrac,mathtools,bm}
\usepackage[bookmarks=false]{hyperref} 
\usepackage[numbers, sort&compress]{natbib}
\usepackage[english]{babel}
\usepackage{times}
\usepackage{bbm,mdframed}
\usepackage[dvipsnames]{xcolor}
\usepackage{xspace}
\usepackage{rotating,multirow} 
\usepackage{tikz}
\usetikzlibrary{arrows}
\usetikzlibrary{shapes} 
\usepackage{algpseudocode,algorithm,algorithmicx} 
\usepackage{fancyhdr} 
\usepackage{tabularx}
\usepackage{subcaption}
\usepackage{multirow}
\usepackage{float}
\usepackage{soul}
\usepackage{color}
\usepackage{enumitem}
\usepackage{comment}
\usepackage{leading}
\usepackage[noabbrev]{cleveref}

\newcommand{\Prob}{\ensuremath{\mathbb{P}}}

\newcommand{\Exp}{\ensuremath{\mathbb{E}}}

\newcommand{\real}{\ensuremath{\mathbb{R}}}

\newcommand{\defn}{\ensuremath{: \, = }}

\newcommand{\grad}{\nabla S(x_t)}
\newcommand{\shortgrad}{\nabla_t} 
\newcommand{\diff}{d_t} 
\definecolor{darkbrown}{rgb}{0.4, 0.26, 0.13}

\makeatletter
\newcommand\footnoteref[1]{\protected@xdef\@thefnmark{\ref{#1}}\@footnotemark}
\makeatother

\newtheorem{theorem}{Theorem}

\makeatletter \renewcommand*{\@fnsymbol}[1]{\ensuremath{\ifcase#1\or
    \dagger\or \dagger\or \ddagger\or \mathsection\or
    \mathparagraph\or \|\or **\or \dagger\dagger \or \ddagger\ddagger
    \else\@ctrerr\fi}} \makeatother

\renewcommand{\defn}{\vcentcolon=}
\newcommand{\origsym}{\star}
\newcommand{\xorig}{\ensuremath{{x^\origsym}}}
\newcommand{\corig}{\ensuremath{{c^\origsym}}}
\newcommand{\ctarget}{\ensuremath{{c^\dagger}}}

\newcommand{\Class}{\ensuremath{C}}
\newcommand{\Disc}{\ensuremath{F}}
\newcommand{\sign}{\operatorname{sign}}

\newcommand{\bd}{\operatorname{bd}}

\newcommand{\interstep}{\ensuremath{\xi}}

\newcommand{\resid}{\ensuremath{r}}

\newcommand{\inprod}[2]{\ensuremath{\big \langle #1 , \, #2 \big \rangle}}

\ifCLASSINFOpdf
\else
\fi

\usepackage{dblfloatfix}

\begin{document}
%
\title{HopSkipJumpAttack: A Query-Efficient Decision-Based Attack}

\author{ 
Jianbo Chen$^{*}$ \hspace{5mm} Michael I. Jordan$^*$ \hspace{5mm} Martin J. Wainwright$^{*, \dagger}$ \\
University of California, Berkeley$^{*}$\hspace{5mm}Voleon Group$^\dagger$\\
  \texttt{\{jianbochen@, jordan@cs., wainwrig@\}berkeley.edu}
}

%


\maketitle

\begin{abstract}
  The goal of a decision-based adversarial attack on a trained model
  is to generate adversarial examples based solely on observing output
  labels returned by the targeted model. We develop HopSkipJumpAttack,
  a family of algorithms based on a novel estimate of the gradient
  direction using binary information at the decision boundary. The
  proposed family includes both untargeted and targeted attacks
  optimized for $\ell_2$ and $\ell_\infty$ similarity metrics
  respectively. Theoretical analysis is provided for the proposed
  algorithms and the gradient direction estimate. Experiments show
  HopSkipJumpAttack requires significantly fewer model queries than
  several state-of-the-art decision-based adversarial attacks. {It also achieves competitive performance in attacking several widely-used defense mechanisms.}
\end{abstract}


%
\IEEEpeerreviewmaketitle

\section{Introduction}\label{sec:intro}

Although deep neural networks have achieved state-of-the-art
performance on a variety of tasks, they have been shown to be
vulnerable to \emph{adversarial examples}---that is, maliciously
perturbed examples that are almost identical to original samples in
human perception, but cause models to make incorrect
decisions~\cite{szegedy2013intriguing}. The vulnerability of neural
networks to adversarial examples implies a security risk in
applications with real-world consequences, such as self-driving cars,
robotics, financial services, and criminal justice; in addition, it
highlights fundamental differences between human learning and existing
machine-based systems. The study of adversarial examples is thus
necessary to identify the limitation of current machine learning
algorithms, provide a metric for robustness, investigate the potential
risk, and suggest ways to improve the robustness of models.

Recent years have witnessed a flurry of research on the design of new
algorithms for generating adversarial
examples~\cite{szegedy2013intriguing, goodfellow2014explaining,
  kurakin2016adversarial, moosavi2016deepfool,papernot2016limitations,
  carlini2017towards,madry2018towards, chen2017zoo, ilyas2018black,
  ilyas2018prior, liu2016delving, papernot2016transferability,
  papernot2017practical, brendel2018decisionbased,
  brunner2018guessing, cheng2018queryefficient}. Adversarial examples
can be categorized according to at least three different criteria: the
similarity metric, the attack goal, and the threat model. Commonly
used similarity metrics are $\ell_p$-distances between adversarial and
original examples with $p \in \{0, 2, \infty\}$.  The goal of attack
is either untargeted or targeted.  The goal of an untargeted attack is
to perturb the input so as to cause any type of misclassification,
whereas the goal of a targeted attack is to alter the decision of the
model to a pre-specific target class.  Changing the loss function
allows for switching between two types of
attacks~\cite{kurakin2016adversarial, papernot2016limitations,
  carlini2017towards}.

\begin{figure}[!bt]
\centering \includegraphics[width=1.0\linewidth]{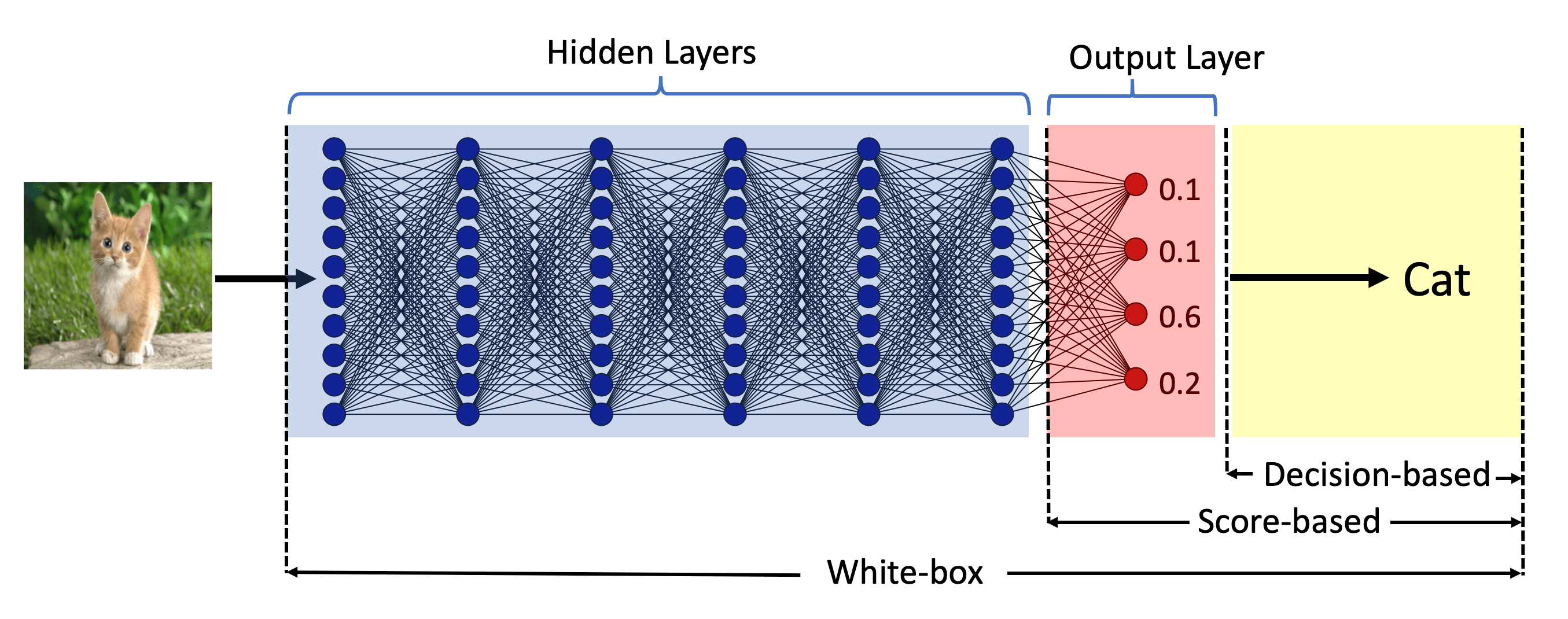}
\caption{An illustration of accessible components of the target model for each of 
the three threat models. A white-box threat model assumes access to the whole model; a score-based threat model assumes access to the output layer; a decision-based threat model assumes access to the predicted label alone.}
\label{fig:threat}
\end{figure}

Perhaps the most important criterion in practice is the \emph{threat
  model}, of which there are two primary types: white-box and
black-box. In the white-box setting, an attacker has complete access
to the model, including its structure and weights. Under this setting,
the generation of adversarial examples is often formulated as an
optimization problem, which is solved either via treating
misclassification loss as a regularization
\cite{szegedy2013intriguing, carlini2017towards} or via tackling the
dual as a constrained optimization problem
\cite{goodfellow2014explaining, kurakin2016adversarial,
  madry2018towards}.
In the black-box setting, an attacker can only access outputs of the
target model. Based on whether one has access to the full probability
or the label of a given input, black-box attacks are further divided
into score-based and decision-based. See Figure~\ref{fig:threat} for 
an illustration of accessible components of the target model for each of 
the three threat models. \citet{chen2017zoo} and \citet{ilyas2018black, 
ilyas2018prior} introduced score-based methods
using zeroth-order gradient estimation to craft adversarial examples.

The most practical threat model is that in which an attacker has
access to decisions alone.  A widely studied type of the
decision-based attack is transfer-based attack. \citet{liu2016delving}
showed that adversarial examples generated on an ensemble of deep
neural networks from a white-box attack can be transferred to an
unseen neural network. \citet{papernot2016transferability,
  papernot2017practical} proposed to train a substitute model by
querying the target model. 
{However, transfer-based attack often requires a carefully-designed substitute
model, or even access to part of the training data. Moreover, they can be defended against via training on a data
set augmented by adversarial examples from multiple static pre-trained
models~\cite{tramer2018ensemble}.
} In recent work, \citet{brendel2018decisionbased} proposed Boundary Attack, which
generates adversarial examples via rejection sampling. While relying
neither on training data nor on the assumption of transferability,
this attack method achieves comparable performance with
state-of-the-art white-box attacks such as C\&W
attack~\cite{carlini2017towards}. {One limitation of Boundary Attack, however, is that it was formulated only for $\ell_2$-distance. Moreover, it requires a relatively large number of model queries, rendering it impractical for real-world applications.}


{
It is more realistic to evaluate the vulnerability of a machine learning system under the decision-based attack with a limited budget of model queries. Online image classification platforms often set a limit on the allowed number of queries within a certain time period. For example, the cloud vision API from Google currently allow 1,800 requests per minute. Query inefficiency thus leads to clock-time inefficiency and prevents an attacker from carrying out large-scale attacks. A system may also be set to recognize the behavior of feeding a large number of similar queries within a small amount of time as a fraud, which will automatically filter out query-inefficient decision-based attacks. Last but not least, a smaller query budget directly implies less cost in evaluation and research. Query-efficient algorithms help save the cost of evaluating the robustness of public platforms, which incur a cost for each query made by the attacker. It also helps facilitate research in adversarial vulnerability, as such a decision-based attack which does not require access to model details may be used as a simple and efficient first step in evaluating new defense mechanisms, as we will see in Section~\ref{sec:defense} and Appendix~\ref{app:nondiff}.
}

In this paper, we study decision-based attacks under an optimization
framework, and propose a novel family of algorithms for generating
both targeted and untargeted adversarial examples that are optimized
for minimum distance with respect to either the $\ell_2$-distance or
$\ell_\infty$ distance. The family of algorithms is iterative in
nature, with each iteration involving three steps: estimation of the
gradient direction, step-size search via geometric progression, and
Boundary search via a binary search. {Theoretical analysis has been carried out for the optimization framework and the gradient direction estimate, which not only provides insights for choosing hyperparamters, but also motivating essential steps in the proposed algorithms.
}
We refer to the algorithm as
HopSkipJumpAttack\footnote{A hop, skip, and a jump originally referred to an exercise or game involving these movements dating from the early 1700s, but by the mid-1800s it was also being used figuratively for the short distance so covered.}.
In summary, our contributions are the following:
\vspace{-0.3cm}
\begin{itemize}[leftmargin=*]
    \item We propose a novel unbiased estimate of gradient direction
      at the decision boundary based solely on access to model decisions,
      and propose ways to control the error from deviation from the boundary.
    \item We design a family of algorithms, HopSkipJumpAttack, based
      on the proposed estimate and our analysis, which is
      hyperparameter-free, query-efficient and equipped with a
      convergence analysis.
    \item We demonstrate the superior efficiency of our algorithm over {several state-of-the-art decision-based attacks} through extensive experiments.
    \item {Through the evaluation of several defense mechanisms such as defensive distillation, region-based classification, adversarial training and input binarization, we suggest our attack can be used as a simple and efficient first step for researchers to evaluate new defense mechanisms.}
\end{itemize}
\textbf{Roadmap. } In Section~\ref{sec:related}, we describe previous
work on decision-based adversarial attacks and their relationship to
our algorithm. We also discuss the connection of our algorithm to
zeroth-order optimization. In Section~\ref{sec:framework}, we propose
and analyze a novel iterative algorithm which requires access to the
gradient information. Each step carries out a gradient update from the
boundary, and then projects back to the boundary again. In
Section~\ref{sec:gradient}, we introduce a novel asymptotically
unbiased gradient-direction estimate at the boundary, and a
binary-search procedure to approach the boundary. We also discuss how
to control errors with deviation from the boundary. The analysis motivates a decision-based algorithm, HopSkipJumpAttack
(Algorithm~\ref{alg:attack}). Experimental results are provided in
Section~\ref{sec:exp}. We conclude in Section~\ref{SecDiscussion} with
a discussion of future work.


\section{Related work}
\label{sec:related}

\subsection{Decision-based attacks}
Most related to our work is the Boundary Attack method introduced
by~\citet{brendel2018decisionbased}. Boundary Attack is an iterative
algorithm based on rejective sampling, initialized at an image that
lies in the target class. At each step, a perturbation is sampled from
a proposal distribution, which reduces the distance of the perturbed
image towards the original input. If the perturbed image still lies in
the target class, the perturbation is kept. Otherwise, the
perturbation is dropped. Boundary Attack achieves performance
comparable to state-of-the-art white-box attacks on deep neural
networks for image classification. The key obstacle to its practical
application is, however, the demand for a large number of model
queries. In practice, the required number of model queries for crafting an adversarial example directly determines the level of the threat imposed by a decision-based attack. 
One source of inefficiency in Boundary
Attack is the rejection of perturbations which deviate from the target
class. In our algorithm, the perturbations are used for estimation of a
gradient direction.


Several other decision-based attacks have been proposed to improve efficiency. \citet{brunner2018guessing} introduced Biased Boundary Attack, which biases the sampling procedure by combining low-frequency random noise with the gradient from a substitute model. Biased Boundary Attack is able to significantly reduce the number of model queries. However, it relies on the transferability between the
substitute model and the target model, as with other transfer-based
attacks. Our algorithm does not rely on the additional assumption of
transferability. Instead, it achieves a significant improvement over Boundary Attack through the exploitation of discarded information into the gradient-direction estimation. \citet{ilyas2018black} proposed Limited attack in the label-only setting, which directly performs projected gradient descent by estimating gradients based on novel proxy scores. \citet{cheng2018queryefficient} introduced Opt attack, which transforms the original problem to a continuous version, and solves the new problem via randomized zeroth-order gradient update. 
{Our algorithm approaches the original problem directly via a novel gradient-direction estimate, leading to improved query efficiency over both Limited Attack and Opt Attack.} The majority of model queries in HopSkipJumpAttack come in mini-batches, which also leads to improved clock-time efficiency over Boundary Attack.

\subsection{Zeroth-order optimization} 

Zeroth-order optimization refers to the problem of optimizing a
function $f$ based only on access to function values $f(x)$, as
opposed to gradient values $\nabla f(x)$.  Such problems have been
extensively studied in the convex optimization and bandit
literatures. \citet{flaxman2005online} studied one-point randomized
estimate of gradient for bandit convex
optimization. \citet{agarwal2011stochastic} and
\citet{nesterov2017random} demonstrated that faster convergence can be
achieved by using two function evaluations for estimating the
gradient. \citet{duchi2015optimal} established optimal rates of convex
zeroth-order optimization via mirror descent with two-point gradient
estimates. Zeroth-order algorithms have been applied to the generation
of adversarial examples under the score-based threat
model~\cite{chen2017zoo, ilyas2018black, ilyas2018prior}.  Subsequent
work~\cite{liu2018zeroth} proposed and analyzed an algorithm based on variance-reduced
stochastic gradient estimates.


We formulate decision-based attack as an optimization
problem. A core component of our proposed algorithm is a gradient-direction estimate, the design of which is motivated by zeroth-order optimization. However, the problem of decision-based attack is more challenging than zeroth-order optimization, essentially because we only have binary information from output labels of the target model, rather than function values.


\section{An optimization framework}\label{sec:framework}

In this section, we describe an optimization framework for finding
adversarial instances for an $m$-ary classification model of the
following type.  The first component is a \emph{discriminant function}
$\Disc: \real^d \rightarrow \real^m$ that accepts an input $x \in [0,
  1]^d$ and produces an output \mbox{$y \in \Delta_m \defn \left\{ y
  \in [0,1]^m \mid \sum_{c=1}^m y_c = 1 \right \}$.}  The output
vector $y = (\Disc_1(x), \ldots, \Disc_m(x))$ can be viewed as a
probability distribution over the label set $[m] = \{1, \ldots, m \}$.
Based on the function $F$, the classifier $\Class :\real^d \rightarrow
[m]$ assigns input $x$ to the class with maximum probability---that
is,
$$\Class(x) \defn \arg \max_{c \in [m]} F_c(x).$$

We study adversaries of both the untargeted and targeted
varieties.  Given some input $\xorig$, the goal of an \emph{untargeted
  attack} is to change the original classifier decision $\corig \defn
\Class(\xorig)$ to any $c \in [m] \backslash \{\corig\}$, whereas the
goal of a \emph{targeted attack} is to change the decision to some
pre-specified $\ctarget \in [m] \backslash \{\corig\}$.  Formally, if
we define the function $S_\xorig: \real^d \rightarrow \real$ via
\begin{align}
S_\xorig(x') \defn \begin{cases} \max \limits_{c \neq \corig} F_c(x')
  - F_{\corig}(x') & \mbox{(Untargeted)} \\
  F_\ctarget(x') - \max \limits_{c \neq \ctarget} F_c(x') & \mbox{(Targeted)}
\end{cases}
\end{align}
then a perturbed image $x'$ is a successful attack if and only if
$S_\xorig(x') > 0$. The boundary between successful and unsuccessful
perturbed images is 
$$\bd(S_\xorig) \defn \left \{z \in [0,1]^d \, \mid
\, S_\xorig(z) = 0 \right \}.$$
As an indicator of successful perturbation, we introduce the Boolean-valued function
$\phi_\xorig:[0,1]^d \to \{-1, 1\}$ via
\begin{align*}
  \phi_\xorig(x') & \defn \sign \left( S_\xorig(x') \right) \; = \;
  \begin{cases} 
  1 & \mbox{if $S_\xorig(x') > 0$,} 
  \\ -1 & \mbox{otherwise.}
  \end{cases}
\end{align*}
This function is accessible in the decision-based setting, as it can
be computed by querying the classifier $\Class$ alone.  The goal of an
adversarial attack is to generate a perturbed sample $x'$ such that
$\phi_\xorig(x') = 1$, while keeping $x'$ close to the original sample
$\xorig$. This can be formulated as the optimization problem
\begin{align}
  \label{prob:constrained}
\min_{x'} d(x', \xorig) \quad \mbox{such that} \quad \phi_\xorig(x') =
1,
\end{align}
where $d$ is a distance function that quantifies similarity.  Standard
choices of $d$ studied in past
work~\cite{papernot2016limitations,goodfellow2014explaining,
  carlini2017towards} include the usual $\ell_p$-norms, for $p \in
\{0, 2, \infty \}$.

\subsection{An iterative algorithm for $\ell_2$ distance}
Consider the case of the optimization problem~\eqref{prob:constrained}
with the $\ell_2$-norm $d(x,\xorig) = \|x - \xorig\|_2$.  We first
specify an iterative algorithm that is given access to the gradient
$\nabla S_\xorig$.  Given an initial vector $x_0$ such that
$S_\xorig(x_0) > 0$ and a stepsize sequence $\{\interstep_t \}_{t
  \geq 0}$, it performs the update
\begin{align}
\label{eq:update_l2}
x_{t+1}= \alpha_t \xorig + (1-\alpha_t) \left \{ x_t + \interstep_t
\frac{\nabla S_\xorig(x_t)}{\|\nabla S_\xorig(x_t)\|_2} \right \},
\end{align} 
where $\interstep_t$ is a positive step size. 
Here the line search parameter $\alpha_t\in [0, 1]$ is chosen such that
$S_\xorig(x_{t+1})=0$---that is, so that the next iterate $x_{t+1}$
lies on the boundary. The motivation for this choice is that our
gradient-direction estimate in Section~\ref{sec:gradient} is only
valid near the boundary.


We now analyze this algorithm with the assumption that we have access to the gradient of  $S_\xorig$ in the setting of binary classification.
Assume that the function $S_\xorig$ is
twice differentiable with a locally Lipschitz gradient, meaning that there exists $L>0$ such that for all $x, y\in\{z: \|z - \xorig\|_2 \leq \|x_0 - \xorig\|_2\}$, we have
\begin{align}
  \label{EqnLocalLipschitz}
  \|\nabla S_\xorig (x) - \nabla S_\xorig(y)\|_2\leq L\|x - y\|_2, 
\end{align}
In addition, we assume the
gradient is bounded away from zero on the boundary: there exists a
positive $\tilde C > 0$ such that $\|\nabla S_\xorig(z)\|> \tilde C$ for any $z \in
\bd(S_\xorig)$. 

We analyze the behavior of the updates~\eqref{eq:update_l2} in terms
of the angular measure
\begin{align*}
\resid(x_t, \xorig) & \defn \cos \angle \left(x_t - \xorig, \nabla S_\xorig(x_t) \right)\\
&~=\frac{ \inprod{x_t - \xorig}{\nabla
    S_\xorig(x_t)}} {\|x_t - \xorig \|_2\| \nabla S_\xorig(x_t)\|_2},
\end{align*}
corresponding to the cosine of the angle between $x_t - \xorig$ and
the gradient $\nabla S_\xorig(x_t)$.  Note that the condition
$\resid(x, \xorig) = 1$ holds if and only if $x$ is a stationary point
of the optimization~\eqref{prob:constrained}.  The following theorem
guarantees that, with a suitable step size, the updates converge to
such a stationary point:
\begin{theorem}
\label{thm:convergence}
Under the previously stated conditions on $S_\xorig$, suppose that we
compute the updates~\eqref{eq:update_l2} with step size
\mbox{$\interstep_t = \|x_t - \xorig \|_2 t^{-q}$} for some $q \in
\left(\frac{1}{2}, 1 \right)$.  Then there is a universal
constant $c$ such that
\begin{align}
0 \; \leq \; 1 - \resid(x_t, \xorig) & \leq c\; t^{q-1} \quad \mbox{for $t =1, 2,\ldots$.}
\end{align}
In particular, the algorithm converges to a stationary point of
problem~\eqref{prob:constrained}.
\end{theorem}
\noindent {Theorem~\ref{thm:convergence} suggests a scheme for choosing the step size in the algorithm that we present in the next section. An experimental evaluation of the proposed scheme is carried out in Appendix~\ref{app:sensitivity}. The proof of the theorem is constructed by establishing the relationship between the objective value $d(x_t, \xorig)$ and $\resid(x_t, \xorig)$, with a second-order Taylor approximation to the boundary. See Appendix~\ref{app:thm_convergence} for details.} 


\subsection{Extension to $\ell_\infty$-distance}

We now describe how to extend these updates so as to minimize the
$\ell_\infty$-distance.  Consider the $\ell_2$-projection of a point
$x$ onto the sphere of radius $\alpha_t$ centered at $\xorig$:
\begin{align}
\label{eq:l2}
\Pi^2_{\xorig, \alpha_t}(x) &\defn \underset{\|y - \xorig\|_2 \leq
  \alpha_t}{\arg \min} \|y - x\|_2 = \alpha_t \xorig + (1 - \alpha_t)
x.
\end{align}
In terms of this operator, our $\ell_2$-based
update~\eqref{eq:update_l2} can be rewritten in the equivalent form
\begin{align}\label{eq:update_l2_proj}
  x_{t+1} = \Pi_{\xorig, \alpha_t}^2 \left( x_t + \interstep_t
  \frac{\nabla S_\xorig(x_t)}{\|\nabla S_\xorig(x_t)\|_2} \right).
\end{align}

This perspective allows us to extend the algorithm to other
$\ell_p$-norms for $p \neq 2$. For instance, in the case $p = \infty$,
we can define the $\ell_\infty$-projection operator
$\Pi_{\xorig,\alpha}^\infty$.  It performs a per-pixel clip within a
neighborhood of $\xorig$, such that the $i$th entry of
$\Pi_{\xorig,\alpha}^\infty(x)$ is
\begin{align*}
\Pi_{\xorig, \alpha}^\infty(x)_i &\defn \max \left \{ \min\{x^\origsym_i, x^\origsym_i + c \right \}, x_i - c\},
\end{align*}
where $c \defn \alpha\|x - \xorig \|_\infty$.  We propose the
$\ell_\infty$-version of our algorithm by carrying out the following
update iteratively:
\begin{align}
\label{eq:linf}
x_{t+1} & = \Pi_{\xorig, \alpha_t}^\infty \big(x_t + \interstep_t
\text{sign} (\nabla S_\xorig(x_t))\big),
\end{align}
where $\alpha_t$ is chosen such that $S_\xorig(x_{t+1}) = 0$, and
``sign'' returns the element-wise sign of a vector.   We use the sign of the gradient for faster convergence
in practice, similar to previous work~\cite{goodfellow2014explaining,
  kurakin2016adversarial, madry2018towards}.


\section{A decision-based algorithm based on a novel gradient estimate}
\label{sec:gradient}

We now extend our procedures to the decision-based setting, in which
we have access \emph{only} to the Boolean-valued function
$\phi_\xorig(x) = \sign(S_\xorig(x))$---that is, the method cannot
observe the underlying discriminant function $\Disc$ or its gradient.  In
this section, we introduce a gradient-direction estimate based on
$\phi_\xorig$ when $x_t \in \bd(S_\xorig)$ (so that $S_\xorig(x_t) =
0$ by definition). We proceed to discuss how to approach the
boundary. Then we discuss how to control the error of our estimate
with a deviation from the boundary. We will summarize the analysis
with a decision-based algorithm.

\subsection{At the boundary} Given an iterate $x_t \in \bd(S_\xorig)$
we propose to approximate the direction of the gradient $\nabla
S_\xorig(x_t)$ via the Monte Carlo estimate
\begin{align}
\label{eq:grad_approx}
\widetilde{\nabla S}(x_t, \delta) & \defn \frac{1}{B} \sum_{b=1}^B
\phi_\xorig(x_t + \delta u_b) u_b,
\end{align}
where $\{u_b \}_{b=1}^B$ are i.i.d. draws from the uniform
distribution over the $d$-dimensional sphere, and $\delta$ is small
positive parameter. (The dependence of this estimator on the fixed
centering point $\xorig$ is omitted for notational simplicity.)

The perturbation parameter $\delta$ is necessary, but introduces a
form of bias in the estimate.  Our first result controls this bias,
and shows that $\widetilde{\nabla S}(x_t, \delta)$ is asymptotically
unbiased as \mbox{$\delta \rightarrow 0^+$.}

\begin{theorem}
\label{thm:unbiased}
For a boundary point $x_t$, suppose that $S_\xorig$ has $L$-Lipschitz
gradients in a neighborhood of $x_t$.  Then the cosine of the angle
between $\widetilde{\nabla S}(x_t, \delta)$ and $\nabla S_\xorig(x_t)$
is bounded as
\begin{align}
\cos \angle \left( \Exp[ \widetilde{\nabla S}(x_t,
  \delta)], \nabla S_\xorig(x_t) \right)
\geq 1 - \frac {9L^2\delta^2d^2}{8\|\nabla S(x_t)\|_2^2}.
\end{align}
In particular, we have
\begin{align}
\lim_{\delta \to 0} \cos \angle \left( \Exp[ \widetilde{\nabla S}(x_t,
  \delta)], \nabla S_\xorig(x_t) \right) & = 1,
\end{align}
showing that the estimate is asymptotically unbiased as an estimate of
direction.
\end{theorem}
\noindent {We remark that Theorem~\ref{thm:unbiased} only establishes the asymptotic behavior of the proposed estiamte at the boundary. This also motivates the boundary search step in our algorithm to be discussed in Seciton~\ref{sec:approaching}.
The proof of Theorem~\ref{thm:unbiased} starts from dividing the unit sphere into three
components: the upper cap along the direction of gradient, the
lower cap opposite to the direction of gradient, and the annulus in
between. The error from the annulus can be bounded when $\delta$ is small.}
See Appendix~\ref{app:thm1} for the proof of this theorem. 
As will be seen in the sequel, the size of perturbation $\delta$
should be chosen proportionally to $d^{-1}$; see
Section~\ref{sec:deviation} for details.


\subsection{Approaching the boundary}\label{sec:approaching}

The proposed estimate~\eqref{eq:grad_approx} is only valid at the
boundary. We now describe how we approach the boundary via a binary
search.  Let $\tilde x_t$ denote the updated sample before the
operator $\Pi^p_{x, \alpha_t}$ is applied:
\begin{align}
\label{eq:update}
&\tilde x_t \defn x_t + \interstep_t v_t(x_t, \delta_t), \text{ such that }\\
&
v_t(x_t,\delta_t) =
    \begin{cases}
        \widehat{\nabla S}(x_t, \delta_t) / \|\widehat{\nabla S}(x_t,
        \delta_t)\|_2, \text{ if } p= 2, \\
        \text{sign}(\widehat{\nabla S}(x_t, \delta_t)), \text{ if } p
        = \infty,
    \end{cases}\nonumber
\end{align}
where $\widehat{\nabla S}$ will be introduced later in
equation~\eqref{eq:grad_approx_baseline}, as a variance-reduced
version of $\widetilde{\nabla S}$, and $\delta_t$ is the size of
perturbation at the $t$-th step.

We hope $\tilde x_t$ is at the opposite side of the boundary to $x$ so
that the binary search can be carried out. Therefore, we initialize at
$\tilde x_0$ at the target side with \mbox{$\phi_\xorig(\tilde x_0) =
  1$,} and set \mbox{$x_0 \defn \Pi_{x,\alpha_0}^p(\tilde x_0)$,}
where $\alpha_0$ is chosen via a binary search between $0$ and $1$ to
approach the boundary, stopped at $x_0$ lying on the target side with
$\phi_\xorig(x_0)=1$.  At the $t$-th iteration, we start at $x_t$
lying at the target side $\phi_\xorig(x_t)=1$. The step size is
initialized as
\begin{align}
  \label{eq:stepsize}
\interstep_t:= \|x_t-\xorig\|_p / \sqrt t,
\end{align}
as suggested by Theorem~\ref{thm:convergence} in the $\ell_2$ case,
and is decreased by half until $\phi_\xorig(\tilde x_t)= 1$, which we
call \emph{geometric progression} of $\interstep_t$. Having found an
appropriate $\tilde x_t$, we choose the projection radius $\alpha_t$
via a binary search between $0$ and $1$ to approach the boundary,
which stops at $x_{t+1}$ with $\phi_\xorig(x_{t+1}) = 1$. See
Algorithm~\ref{alg:bs} for the complete binary search, where the
binary search threshold $\theta$ is set to be some small constant.
\begin{algorithm}[H]
       \caption{Bin-Search}
       \label{alg:bs}
       \begin{algorithmic}
       \Require Samples $x', x$, with a binary function $\phi$, such
       that $\phi(x')=1, \phi(x) = 0$, threshold $\theta$, constraint
       $\ell_p$.  \Ensure A sample $x''$ near the boundary.  \State Set
       $\alpha_l = 0$ and $\alpha_u= 1$.  \While{$|\alpha_l -
         \alpha_u|>\theta$}
             \State Set $\alpha_m \leftarrow \frac {\alpha_l + \alpha_u} 2$.
            \If{$\phi(\Pi_{x,\alpha_m}(x')) = 1$}
                \State Set $\alpha_u \leftarrow \alpha_m$.
            \Else
                \State Set $\alpha_l \leftarrow \alpha_m$.
            \EndIf            
           \EndWhile
           \State Output $x'' = \Pi_{x,\alpha_u}(x')$.
       \end{algorithmic}
 \end{algorithm}

\begin{figure}[!bt]
\centering \includegraphics[width=1.0\linewidth]{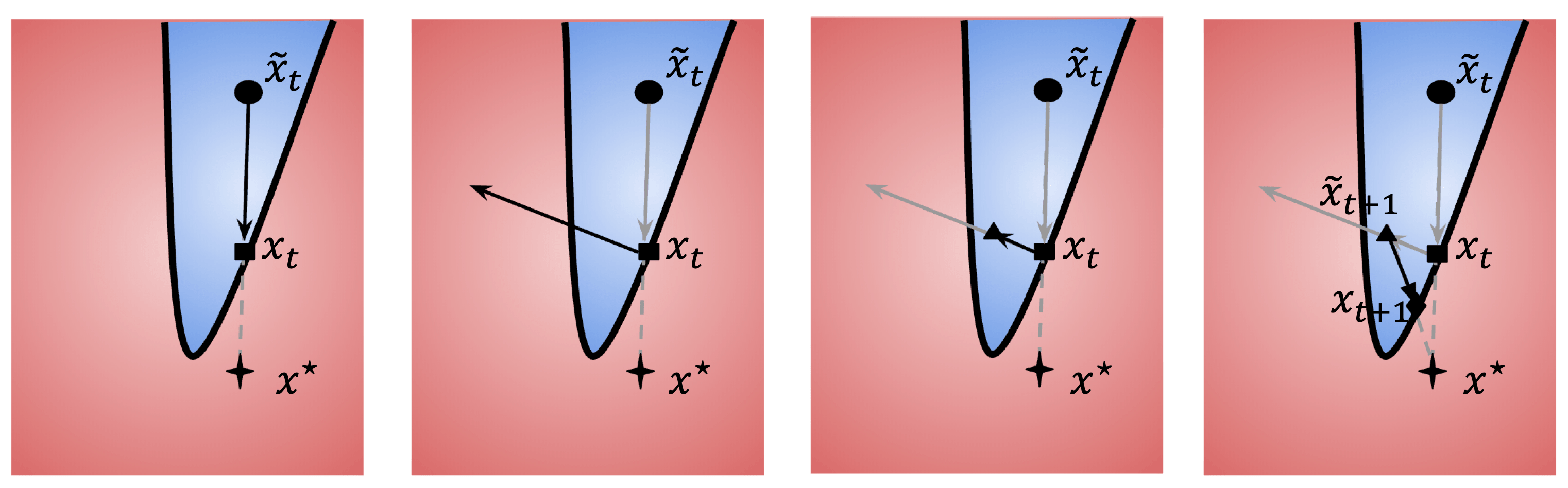}
\caption{Intuitive explanation of HopSkipJumpAttack. (a) Perform a
  binary search to find the boundary, and then update $\tilde x_t \to
  x_t$. (b) Estimate the gradient at the boundary point $x_t$. (c)
  Geometric progression and then update $x_t \to \tilde x_{t+1}$. (d)
  Perform a binary search, and then update $\tilde x_{t+1}\to
  x_{t+1}$.}
\label{fig:visualize}
\end{figure}

\subsection{Controlling errors of deviations from the boundary}\label{sec:deviation}
Binary search never places $x_{t+1}$ exactly onto the boundary. We
analyze the error of the gradient-direction estimate,
and propose two approaches for reducing the error.

\paragraph{Appropriate choice of the size of random perturbation} First, the size of random perturbation $\delta_t$ for estimating the
gradient direction is chosen as a function of image size $d$ and the
binary search threshold $\theta$. This is different from numerical
differentiation, where the optimal choice of $\delta_t$ is at the
scale of round-off errors (e.g., \cite{kincaid2009numerical}). Below
we characterize the error incurred by a large $\delta_t$ as a function
of distance between $\tilde x_t$ and the boundary, and derive the
appropriate choice of $\xi_t$ and $\delta_t$.  In fact, with a Taylor
approximation of $S_\xorig$ at $x_t$, we have
\begin{align*}
S_\xorig(x_t + \delta_t u) = S_\xorig(x_t) + \delta_t \inprod{\nabla
  S_\xorig(x_t)}{u}+ \mathcal O(\delta_t^2).
\end{align*}
At the boundary $S_\xorig(x_t)=0$, the error of gradient approximation
scales at $\mathcal O (\delta_t^2)$, which is minimized by reducing
$\delta_t$ to the scale of rooted round-off error.  However, the
outcome $x_t$ of a finite-step binary search lies close to, but not
exactly on the boundary.

When $\delta_t$ is small enough such that second-order terms can be
omitted, the first-order Taylor approximation implies that
$\phi_\xorig(x_t + \delta_t u) = -1$ if and only if $x_t + \delta_t u$
lies on the spherical cap $\mathcal C$, with 
$$
\mathcal C\defn \Big\{u \mid \Big\langle\frac{\nabla S_\xorig(x_t)}{\|\nabla S_\xorig(x_t)\|_2}, u\Big\rangle <
-\delta_t^{-1}\frac{S_\xorig(x_t)}{\|\nabla S_\xorig(x_t)\|_2}\Big\}.
$$  
On the other hand, the probability mass of $u$ concentrates on the equator in
a high-dimensional sphere, which is characterized by the following
inequality~\cite{ledoux2001concentration}:
\begin{align}
  \label{eq:concentration}
\Prob (u\in \mathcal C)\leq \frac{2}{c}\exp\{-\frac {c^2} 2\},
\mbox{where } c = \frac{\sqrt{d-2}S_\xorig(x_t)}{\delta_t
  \|\nabla S_\xorig(x_t)\|_2}.
\end{align}

A Taylor expansion of
$x_t$ at $x_t'\defn\Pi_\partial^2(x_t)$ yields
\begin{align*}
S_\xorig(x_t) &= \nabla S_\xorig(x_t')^T(x_t - x_t')
+ \mathcal O (\|x_t - x_t'\|_2^2)\\ 
&= \nabla
S_\xorig(x_t)^T(x_t - x_t') + \mathcal O (\|x_t
- x_t'\|_2^2).
\end{align*}
By the Cauchy-Schwarz inequality and the definition of
$\ell_2$-projection, we have
\begin{align*}
\lefteqn{|\nabla S_\xorig(x_t)^T(x_t - x_t')|} \\
&\leq \|\nabla S_\xorig(x_t)\|_2\|x_t - \Pi_\partial^2(x_t)\|_2  \\
&\leq \begin{cases}\|\nabla S_\xorig(x_t)\|_2\theta\|\tilde x_{t-1}-\xorig\|_p, \text{ if } p=2,\\
\|\nabla S_\xorig(x_t)\|_2\theta\|\tilde x_{t-1}-\xorig\|_p\sqrt{d}, \text{ if } p=\infty.
\end{cases}
\end{align*}
This yields
\begin{align*}
c = \mathcal O(\frac{d^q\theta \|\tilde
  x_{t-1} - \xorig\|_p}{\delta_t}),
\end{align*}
where $q = 1 - (1/p)$ is the dual exponent. 
In order to avoid a loss of accuracy from concentration of measure, we
let $\delta_t=d^q\theta\|\tilde x_{t-1} - \xorig\|_2$. 
To make the approximation error independent of dimension $d$, 
we set $\theta$ at
the scale of $d^{-q-1}$, so that $\delta_t$ is proportional to $d^{-1}$, as suggested by Theorem~\ref{thm:unbiased}. This leads to a logarithmic dependence on dimension for the number of model queries.  
In
practice, we set
\begin{align}
  \label{eq:theta}
\theta = d^{-q-1};\text{ } \delta_t = d^{-1}\|\tilde x_{t-1} - \xorig\|_p.
\end{align}

\begin{algorithm}[!bt]
       \caption{HopSkipJumpAttack}
       \label{alg:attack}
       \begin{algorithmic}
       \Require Classifier $C$, a sample $x$, constraint $\ell_p$,
       initial batch size $B_0$, iterations $T$.  
       \Ensure Perturbed image $x_t$.  
       \State Set $\theta$ (Equation~\eqref{eq:theta}).
       \State Initialize at $\tilde x_0$ with $\phi_\xorig(\tilde x_0) =
       1$. 
       \State Compute $d_0 = \|\tilde x_0 - \xorig\|_p$.  
       \For{$t$ in $1,2,\dots,
         T-1$} 
          \State (\textbf{Boundary search})
          \State $x_t = \textsc{Bin-Search}(\tilde x_{t-1}, x,
          \theta, \phi_\xorig, p) $ 
          \State (\textbf{Gradient-direction estimation}) 
          \State Sample $B_t=
          B_0\sqrt{t}$ unit vectors $u_1,\dots, u_{B_t}$.
          \State Set $\delta_t$ (Equation~\eqref{eq:theta}).
          \State Compute $v_t(x_t, \delta_t)$
          (Equation~\eqref{eq:update}).  
          \State (\textbf{Step size search}) 
            \State Initialize step size
          $\interstep_t = \|x_t - \xorig\|_p/\sqrt{t}$.  
          \While {$\phi_\xorig(x_t + \varepsilon_t v_t) = 0$}
              \State $\interstep_t \leftarrow \interstep_t / 2$.
          \EndWhile          
          \State Set $\tilde x_t = x_t + \interstep_t v_t$.
          \State Compute $d_t = \|\tilde x_t - \xorig\|_p$.  
        \EndFor 
        \State Output $x_t =
        \textsc{Bin-Search}(\tilde x_{t-1}, x, \theta,
        \phi_\xorig, p)$.
       \end{algorithmic}
 \end{algorithm}

\begin{table*}[!bt]
\caption{{Median distance at various model queries. The smaller median
  distance at a given model query is bold-faced. BA and HSJA stand for Boundary Attack and HopSkipJumpAttack respectively.}}
\begin{center}
\begin{tabular}{||c|c|c|c||c|c|c||c|c|c||c|c|c||}
 \hline
\multirow{3}{*}{Distance} & \multirow{3}{*}{Data} & \multirow{3}{*}{Model} &\multirow{3}{*}{Objective}& \multicolumn{9}{c||}{Model Queries}\\
 \cline{5-13}
 &&&& \multicolumn{3}{c||}{1K} & \multicolumn{3}{c||}{5K}& \multicolumn{3}{c||}{20K}\\
 \cline{5-13}
 &&&& BA & Opt & HSJA & BA & Opt & HSJA & BA & Opt & HSJA  \\
 \hline
 \hline
\multirow{12}{*}{$\ell_2$}&\multirow{2}{*}{MNIST} & \multirow{2}{*}{CNN} & Untargeted 
&6.14&6.79&\textbf{2.46}&5.45&3.76&\textbf{1.67}&1.50&2.07&\textbf{1.48}\\
\cline{4-13}
& & & Targeted
&5.41&4.84&\textbf{3.26}&5.38&3.90&\textbf{2.24}&1.98&2.49&\textbf{1.96}\\
\cline{2-13}
& \multirow{4}{*}{CIFAR10} & \multirow{2}{*}{ResNet} & Untargeted
&2.78&2.07&\textbf{0.56}&2.34&0.77&\textbf{0.21}&0.27&0.29&\textbf{0.13}\\
\cline{4-13}
& & & Targeted
&7.83&8.21&\textbf{2.53}&5.91&4.76&\textbf{0.41}&0.59&1.06&\textbf{0.21}\\
\cline{3-13}
&& \multirow{2}{*}{DenseNet} & Untargeted
&2.57&1.78&\textbf{0.48}&2.12&0.67&\textbf{0.18}&0.21&0.28&\textbf{0.12}\\
\cline{4-13}
& & & Targeted
&7.70&7.65&\textbf{1.75}&5.33&3.47&\textbf{0.34}&0.35&0.78&\textbf{0.19}\\
\cline{2-13}
& \multirow{4}{*}{CIFAR100} & \multirow{2}{*}{ResNet} & Untargeted 
&1.34&1.20&\textbf{0.20}&1.12&0.41&\textbf{0.08}&0.10&0.14&\textbf{0.06}\\
\cline{4-13}
& & & Targeted 
&9.30&12.43&\textbf{6.12}&7.40&8.34&\textbf{0.92}&1.61&4.06&\textbf{0.29}\\
\cline{3-13}
&& \multirow{2}{*}{DenseNet} & Untargeted 
&1.47&1.22&\textbf{0.25}&1.23&0.34&\textbf{0.11}&0.12&0.13&\textbf{0.08}\\
\cline{4-13}
& & & Targeted 
&8.83&11.72&\textbf{5.10}&6.76&8.22&\textbf{0.75}&0.91&2.89&\textbf{0.26}\\
\cline{2-13}
&\multirow{2}{*}{ImageNet} & \multirow{2}{*}{ResNet} & Untargeted 
&36.86&33.60&\textbf{9.75}&31.95&13.91&\textbf{2.30}&2.71&5.26&\textbf{0.84}\\
\cline{4-13}
& & & Targeted
&87.49&84.38&\textbf{71.99}&82.91&71.83&\textbf{38.79}&40.92&53.78&\textbf{10.95}\\
\cline{1-13}
\hline\hline
\multirow{12}{*}{$\ell_\infty$}&\multirow{2}{*}{MNIST} & \multirow{2}{*}{CNN} & Untargeted 
&0.788&0.641&\textbf{0.235}&0.700&0.587&\textbf{0.167}&0.243&0.545&\textbf{0.136}\\
\cline{4-13}
& & & Targeted
&0.567&0.630&\textbf{0.298}&0.564&0.514&\textbf{0.211}&0.347&0.325&\textbf{0.175}\\
\cline{2-13}
 &\multirow{4}{*}{CIFAR10} & \multirow{2}{*}{ResNet} & Untargeted 
&0.127&0.128&\textbf{0.023}&0.105&0.096&\textbf{0.008}&0.019&0.073&\textbf{0.005}\\
\cline{4-13}
& & & Targeted 
&0.379&0.613&\textbf{0.134}&0.289&0.353&\textbf{0.028}&0.038&0.339&\textbf{0.010}\\
\cline{3-13}
&& \multirow{2}{*}{DenseNet} & Untargeted 
&0.114&0.119&\textbf{0.017}&0.095&0.078&\textbf{0.007}&0.017&0.063&\textbf{0.004}\\
\cline{4-13}
& & & Targeted 
&0.365&0.629&\textbf{0.130}&0.249&0.359&\textbf{0.022}&0.025&0.338&\textbf{0.008}\\
\cline{2-13}
 &\multirow{4}{*}{CIFAR100} & \multirow{2}{*}{ResNet} & Untargeted 
&0.061&0.077&\textbf{0.009}&0.051&0.055&\textbf{0.004}&0.008&0.040&\textbf{0.002}\\
\cline{4-13}
 && & Targeted 
&0.409&0.773&\textbf{0.242}&0.371&0.472&\textbf{0.124}&0.079&0.415&\textbf{0.019}\\
\cline{3-13}
&& \multirow{2}{*}{DenseNet} & Untargeted 
&0.065&0.076&\textbf{0.010}&0.055&0.038&\textbf{0.005}&0.010&0.030&\textbf{0.003}\\
\cline{4-13}
 && & Targeted 
&0.388&0.750&\textbf{0.248}&0.314&0.521&\textbf{0.096}&0.051&0.474&\textbf{0.017}\\
\cline{2-13}
&\multirow{2}{*}{ImageNet} & \multirow{2}{*}{ResNet} & Untargeted 
&0.262&0.287&\textbf{0.057}&0.234&0.271&\textbf{0.017}&0.030&0.248&\textbf{0.007}\\
\cline{4-13}
& & & Targeted
&0.615&0.872&\textbf{0.329}&0.596&0.615&\textbf{0.219}&0.326&0.486&\textbf{0.091}\\
 \hline
 \hline
\end{tabular}
\end{center}
\label{table:efficiency}
\end{table*}

\paragraph{A baseline for variance reduction in gradient-direction estimation}
Another source of error comes from the variance of the estimate, where
we characterize variance of a random vector $v\in\mathbb R^d$ by the
trace of its covariance operator: $\text{Var}(v) : =
\sum_{i=1}^d\text{Var}(v_i)$.  When $x_t$ deviates from the boundary
and $\delta_t$ is not exactly zero, there is an uneven distribution of
perturbed samples at the two sides of the boundary:
$$|\Exp [\phi_\xorig(x_t
  + \delta_t u)]| > 0,$$ 
as we can see from Equation~\eqref{eq:concentration}. To attempt to control the variance,
we introduce a baseline $\overline{\phi_\xorig}$ into the estimate:
$$
\overline{\phi_\xorig} := \frac{1}{B}\sum_{b=1}^B
\phi_\xorig(x_t + \delta u_b),
$$
which yields the following estimate:
\begin{align}
  \label{eq:grad_approx_baseline}
\widehat{\nabla S}(x_t, \delta) :=
\frac{1}{B-1}\sum_{b=1}^B(\phi_\xorig(x_t + \delta u_b) -
\overline{\phi_\xorig})u_b.
\end{align}
It can be easily observed that this estimate is equal to the previous estimate in expectation, and thus still asymptotically unbiased at the boundary: When $x_t\in \bd(S_\xorig)$, we have
\begin{align*}
  &\cos \angle \left( \Exp[ \widehat{\nabla S}(x_t,
  \delta)], \nabla S_\xorig(x_t) \right)
\geq 1 - \frac {9L^2\delta^2d^2}{8\|\nabla S(x_t)\|_2^2}, \\
&~~~~~~\lim_{\delta \to 0} \cos \angle \left( \Exp[ \widehat{\nabla S}(x_t,
  \delta)], \nabla S_\xorig(x_t) \right) = 1.
\end{align*}
Moreover, the introduction of the baseline reduces the variance when
$\mathbb E [\phi_\xorig(x_t + \delta u)]$ deviates from zero. In
particular, the following theorem shows that whenever $|\mathbb E
[\phi_\xorig(x_t + \delta u)]| = \Omega (B^{-\frac{1}{2}})$, the
introduction of a baseline reduces the variance.

%
\begin{theorem}
\label{thm:var}

Defining $\sigma^2 \defn \text{Var}(\phi_\xorig(x_t+\delta u)u)$ as
the variance of one-point estimate, we have
\begin{align*}
\text{Var}(\widehat{\nabla S}(x_t, \delta)) <
\text{Var}(\widetilde{\nabla S}(x_t, \delta))
(1 - \psi),
\end{align*}
where
\begin{align*}
\psi = \frac {2}{\sigma^2 (B-1)}\big(2B\mathbb E
     [\phi_\xorig(x_t + \delta u)]^2-1\big) - \frac{2B-1}{(B-1)^2}.
\end{align*}
\end{theorem}
\noindent See Appendix~\ref{app:thm2} for the proof. We also present an experimental evaluation of our gradient-direction
estimate when the sample deviates from the boundary in {Appendix~\ref{app:sensitivity}}, where we show our proposed choice of
$\delta_t$ and the introduction of baseline yield a performance gain
in estimating gradient.
\subsection{HopSkipJumpAttack} \label{sec:alg}
We now combine the above analysis into an iterative algorithm,
HopSkipJumpAttack. It is initialized with a sample in the target class
for untargeted attack, and with a sample blended with uniform noise
that is misclassified for targeted attack. Each iteration of the
algorithm has three components. First, the iterate from the last
iteration is pushed towards the boundary via a binary search
(Algorithm~\ref{alg:bs}). Second, the gradient direction is estimated
via Equation~\eqref{eq:grad_approx_baseline}. Third, the updating step size along the gradient direction is initialized as Equation~\eqref{eq:stepsize} based on Theorem~\ref{thm:convergence}, and is decreased via geometric progression until perturbation becomes
successful. The next iteration starts with projecting the perturbed sample back to the boundary again. The complete procedure is summarized in Algorithm~\ref{alg:attack}. Figure~\ref{fig:visualize} provides an intuitive visualization of the three steps in $\ell_2$. For all experiments, we initialize the batch size at $100$ and increase it with $\sqrt t$ linearly, so that the variance of the estimate reduces with $t$. When the input domain is bounded in practice, a clip is performed at each step by default.

\begin{figure*}[h]
\centering
\includegraphics[width=0.23\linewidth]{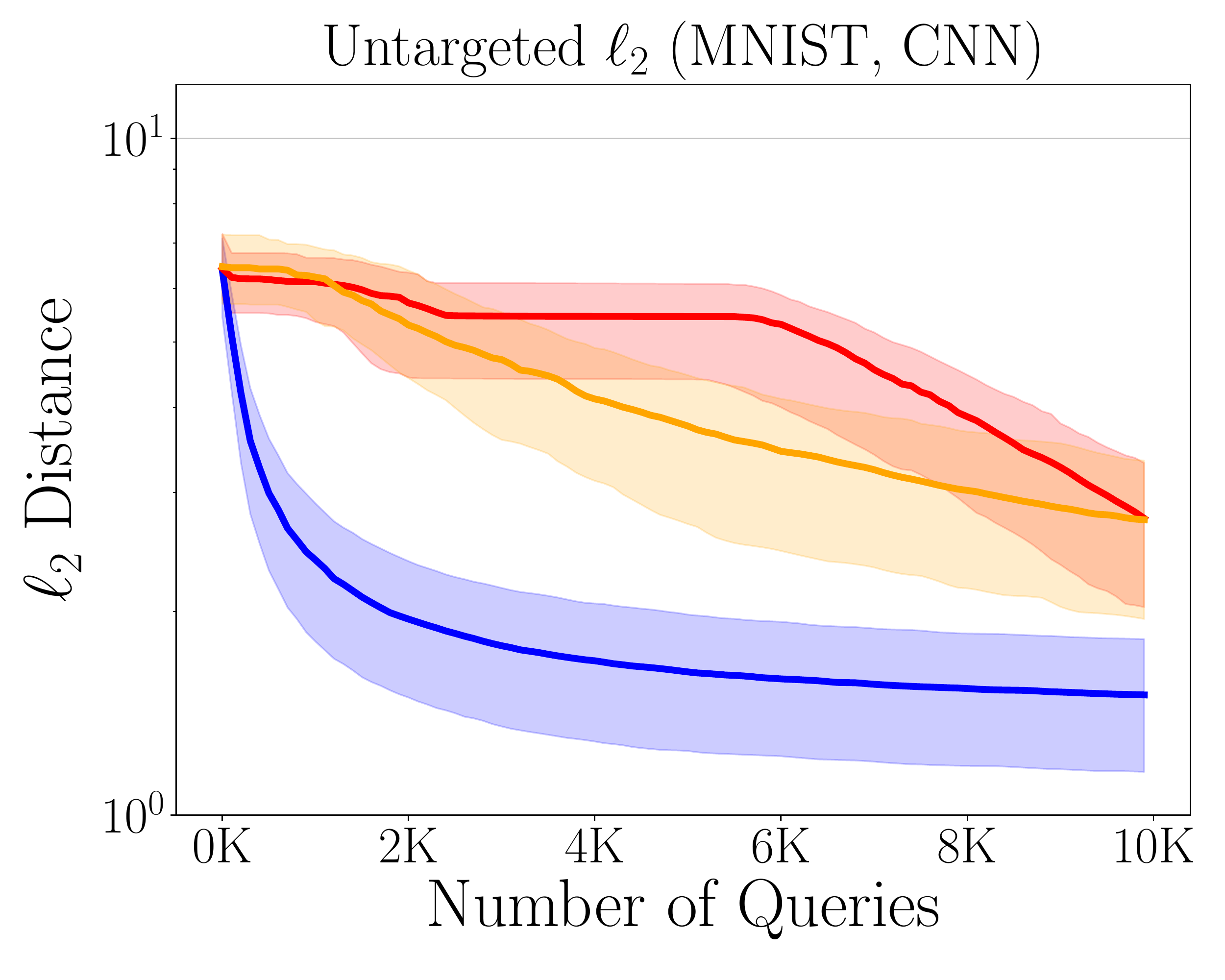} 
\includegraphics[width=0.23\linewidth]{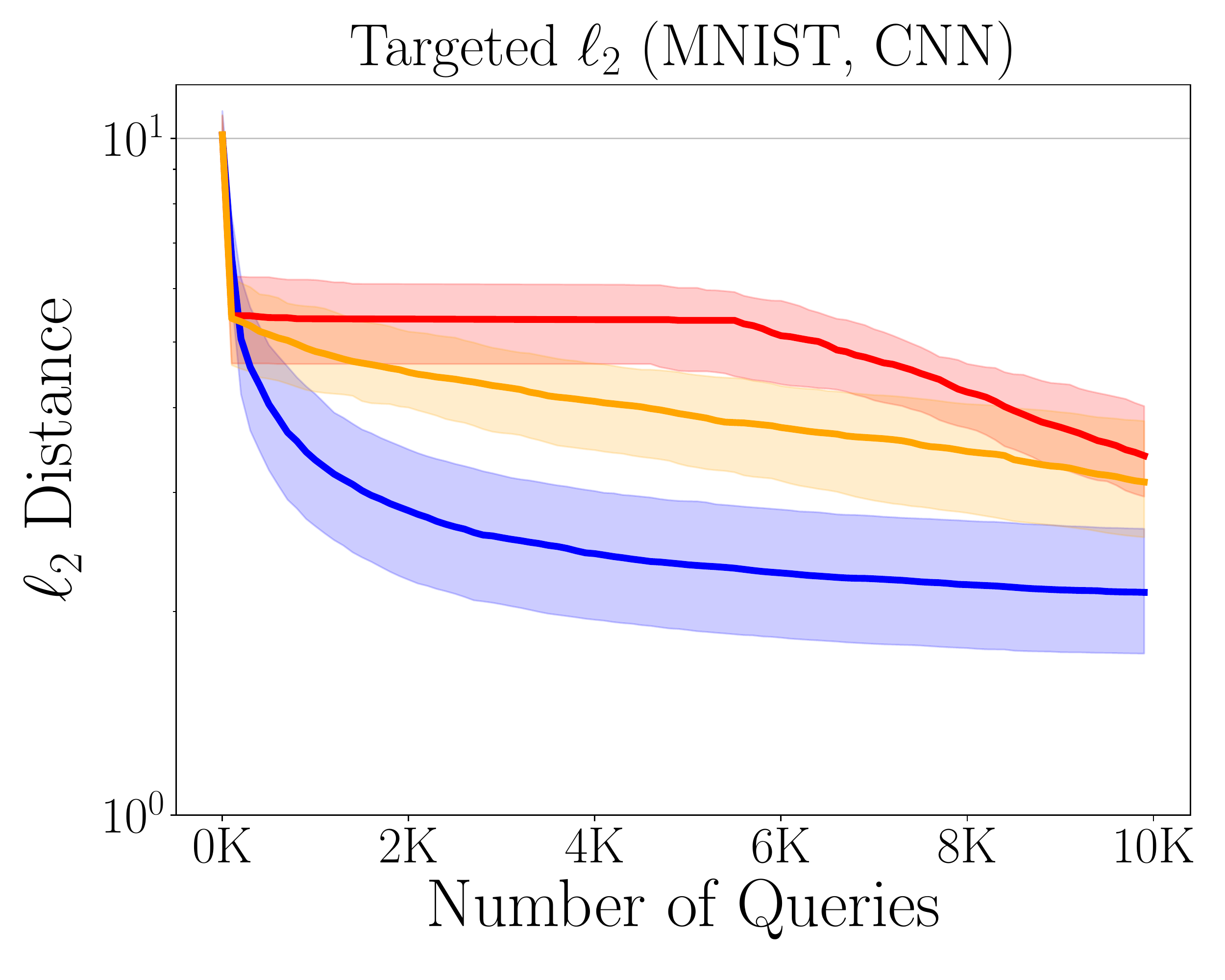} 
\includegraphics[width=0.23\linewidth]{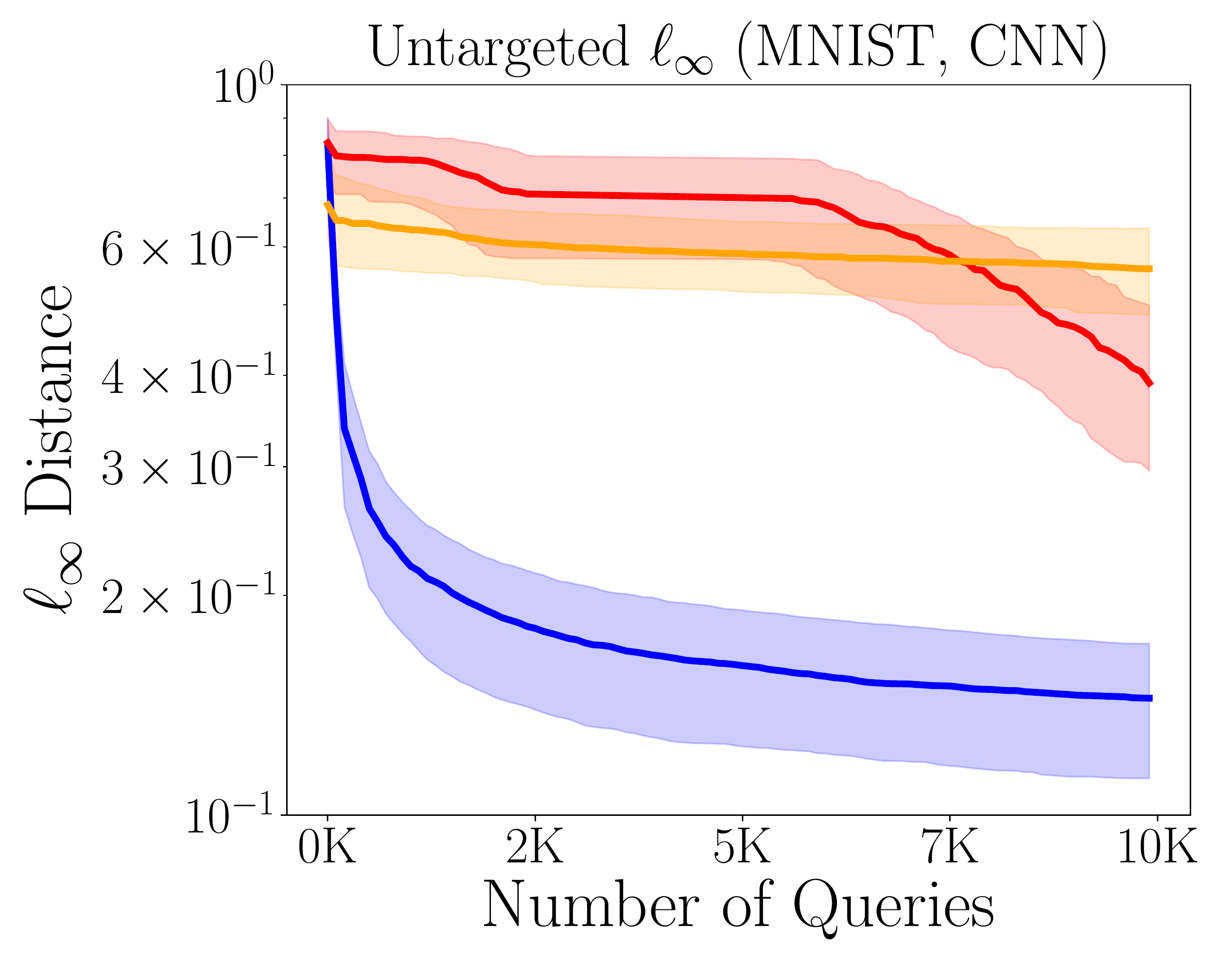} 
\includegraphics[width=0.23\linewidth]{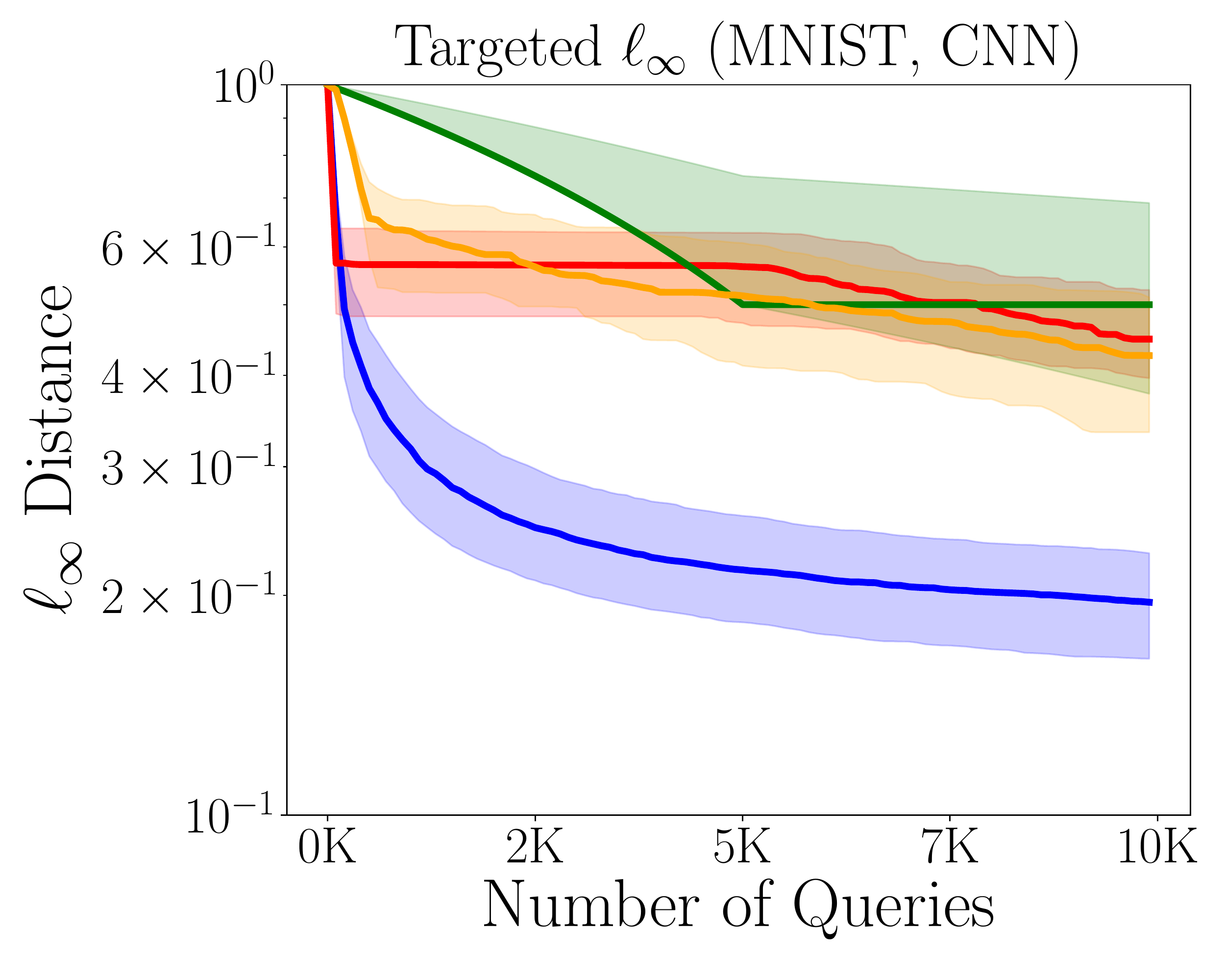} 

\includegraphics[width=0.23\linewidth]{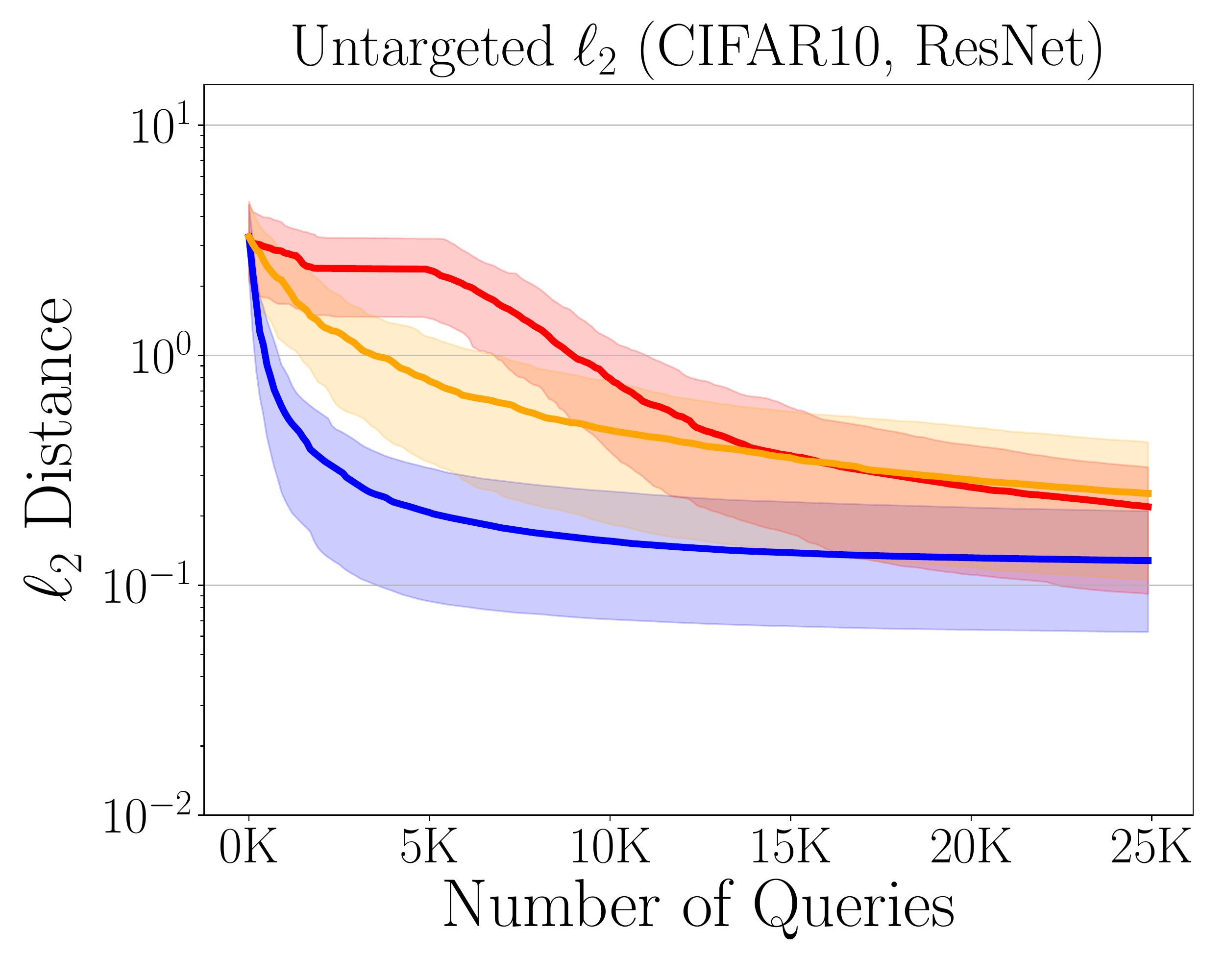}
\includegraphics[width=0.23\linewidth]{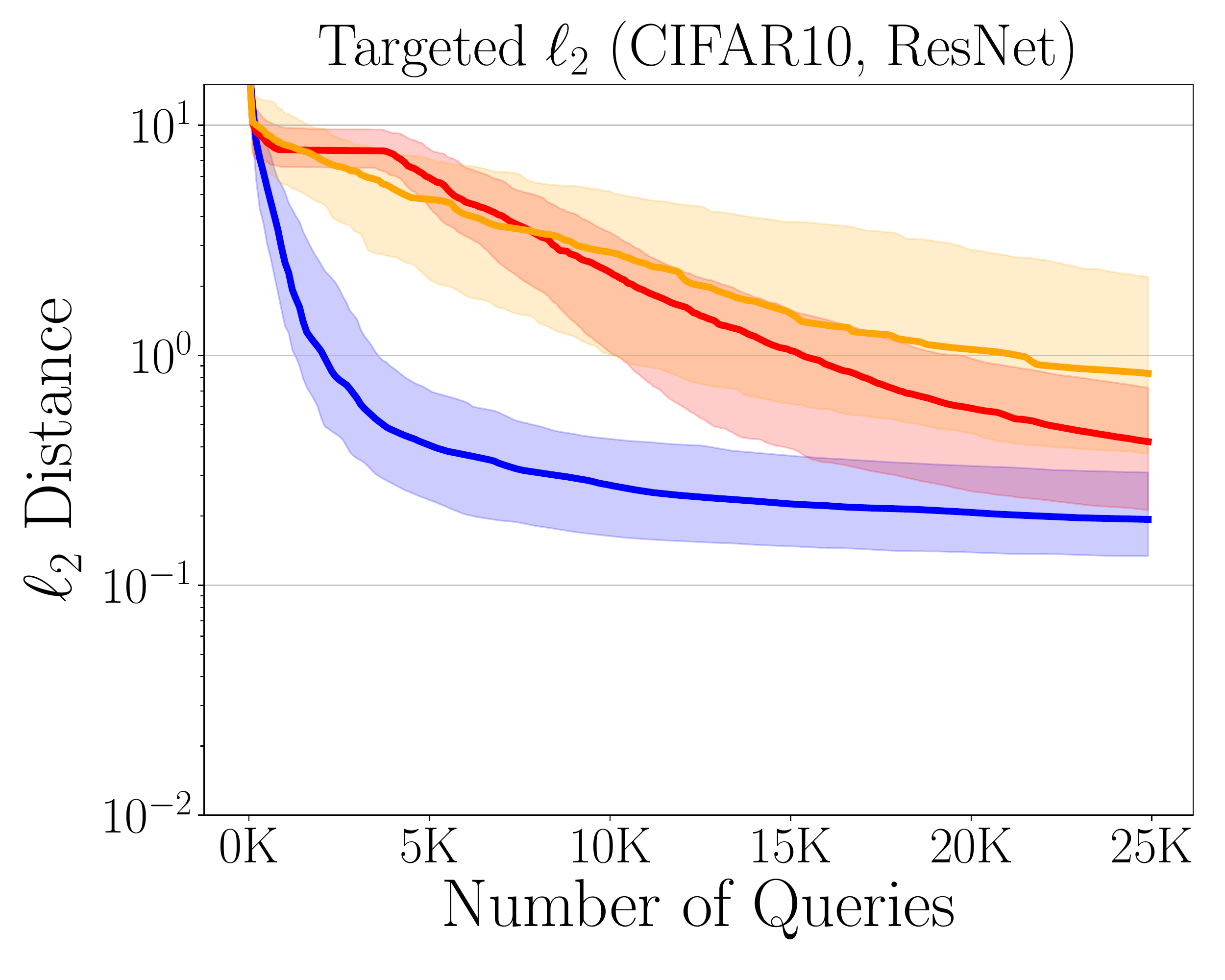}
\includegraphics[width=0.23\linewidth]{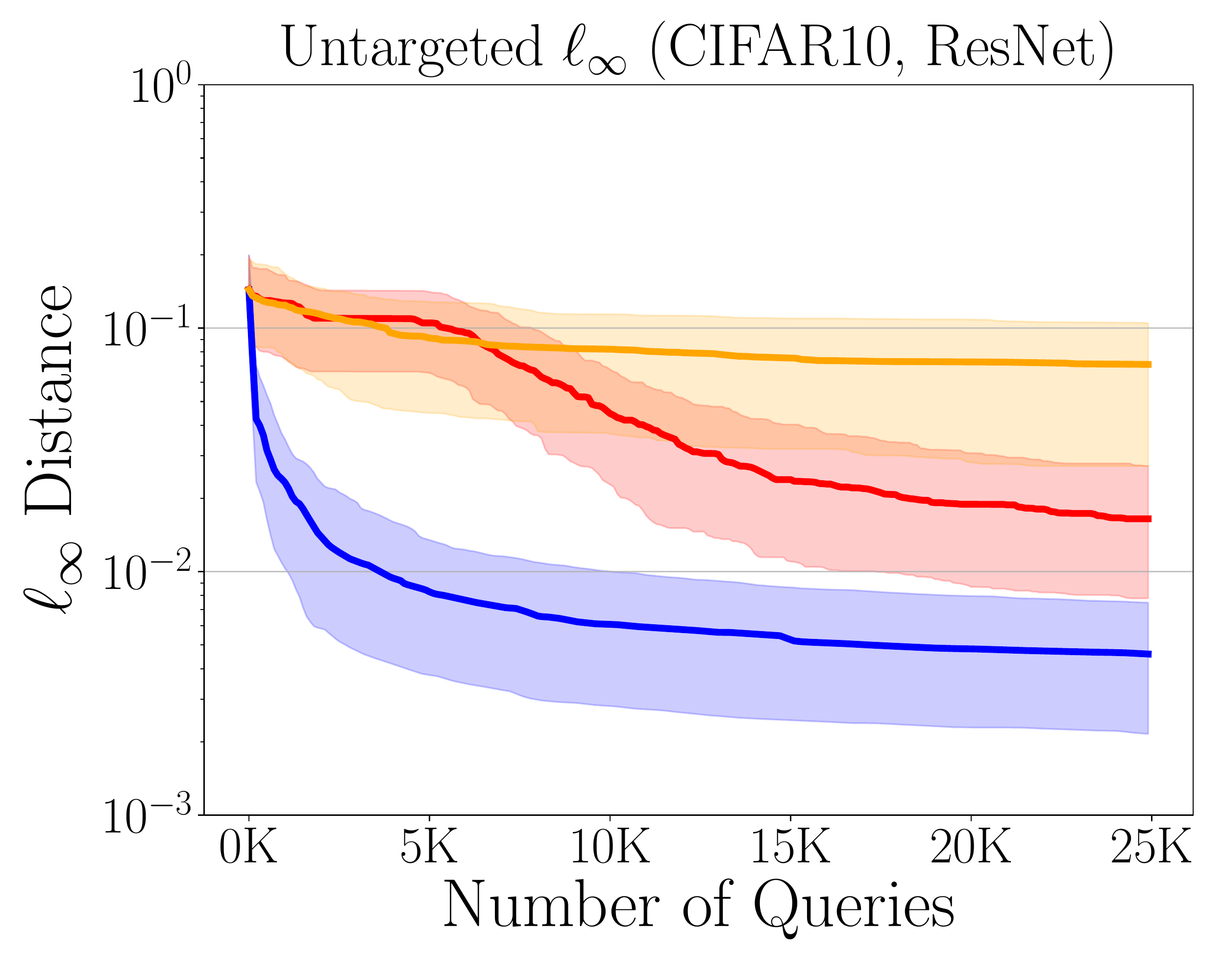}
\includegraphics[width=0.23\linewidth]{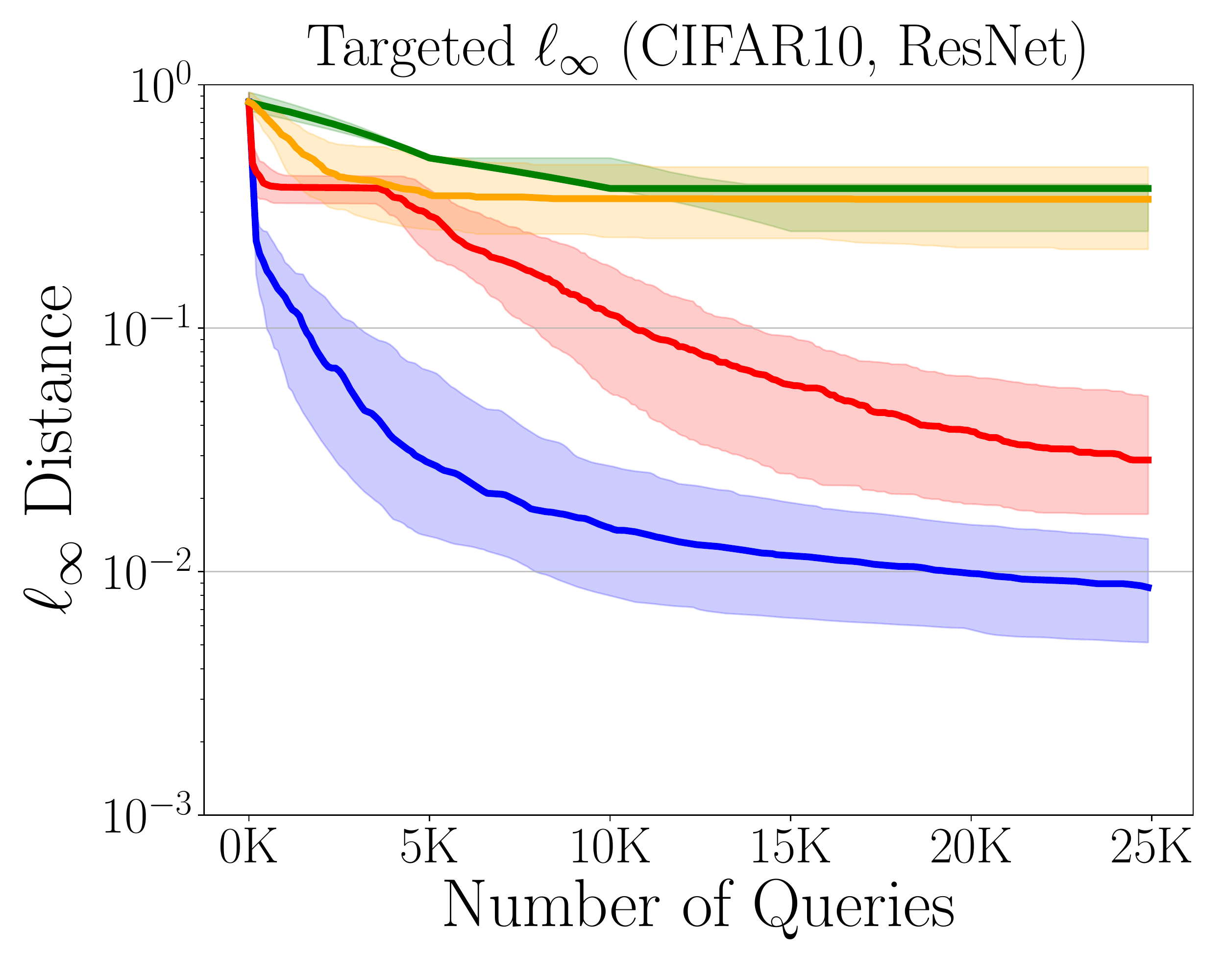}

\includegraphics[width=0.23\linewidth]{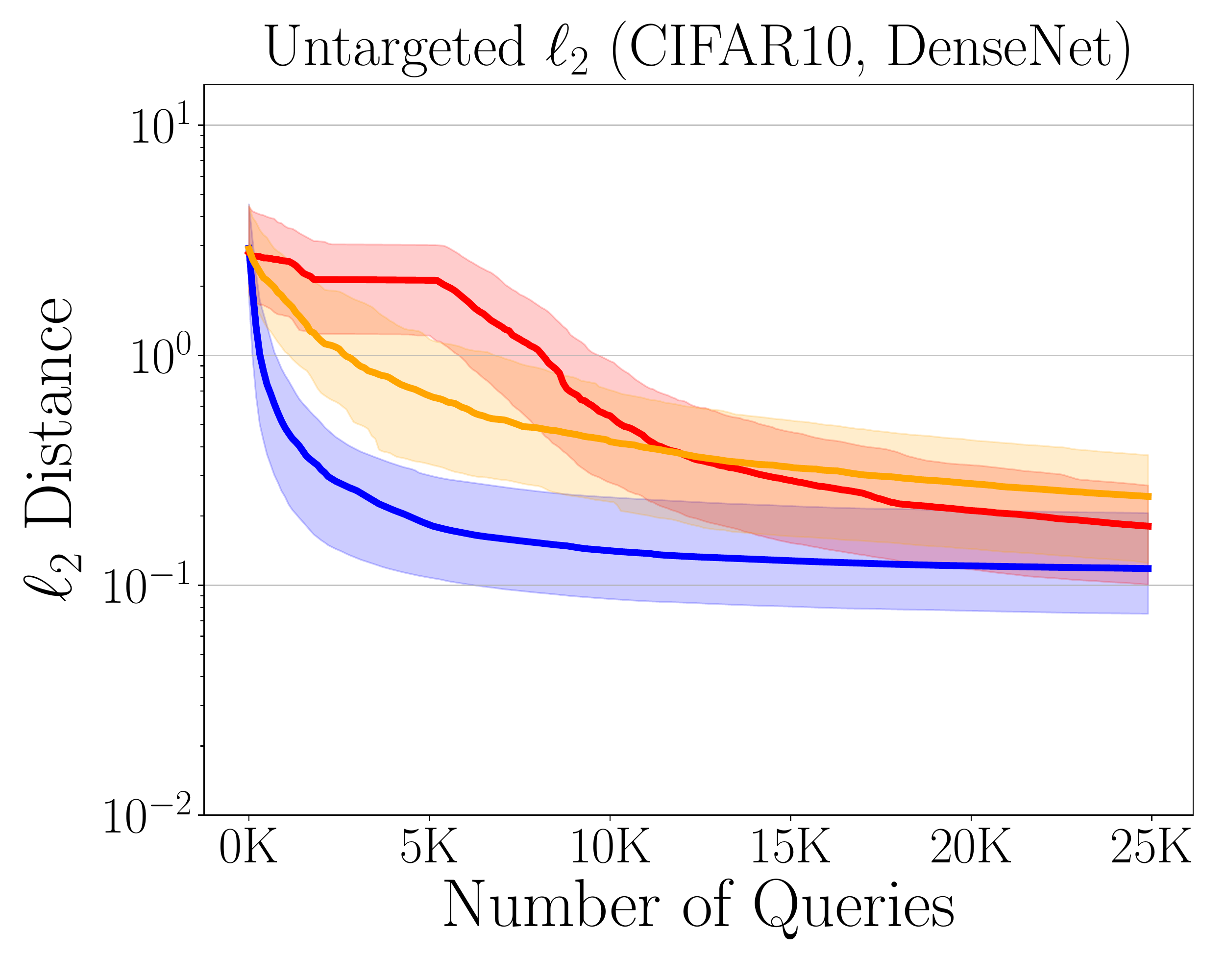} 
\includegraphics[width=0.23\linewidth]{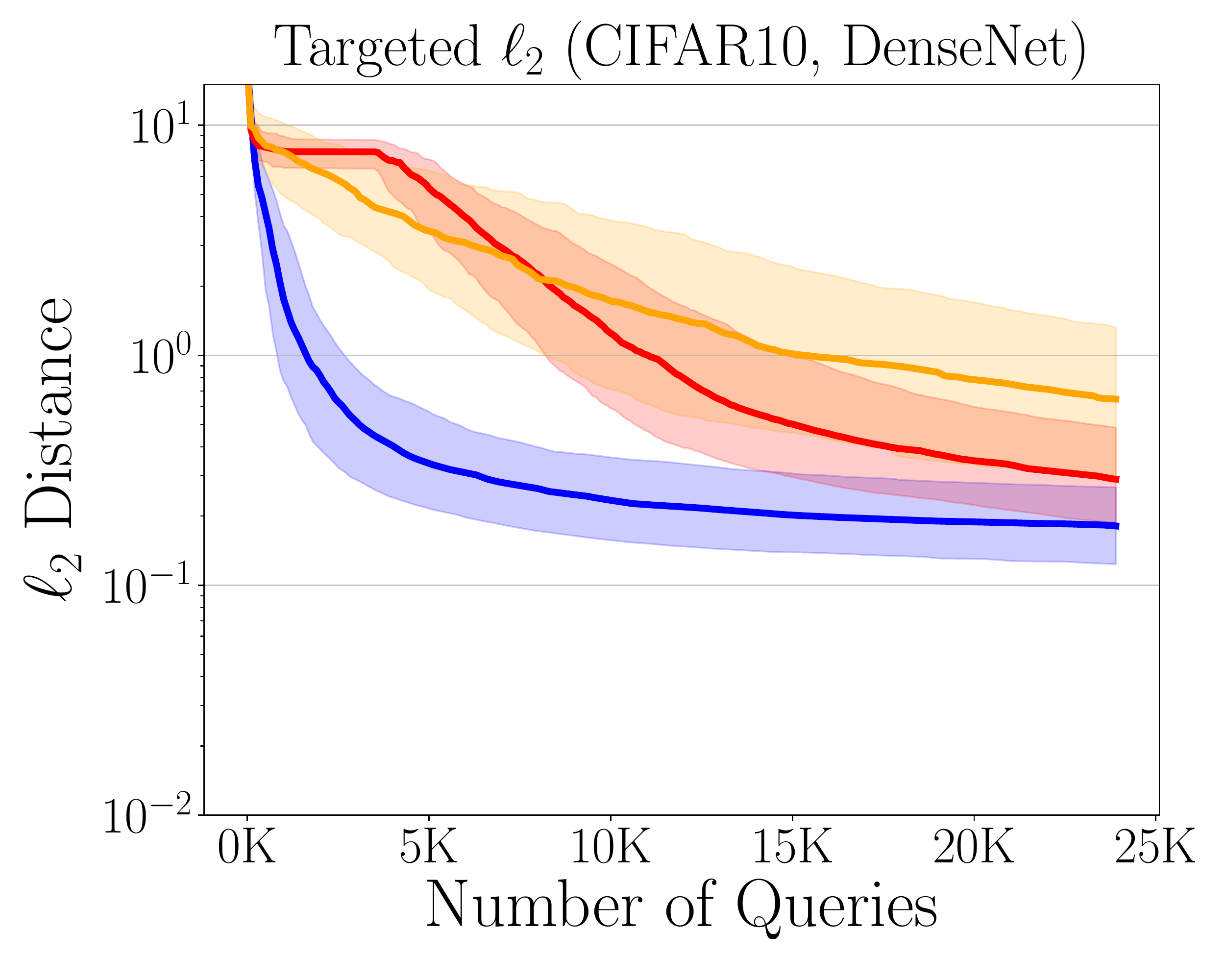}
\includegraphics[width=0.23\linewidth]{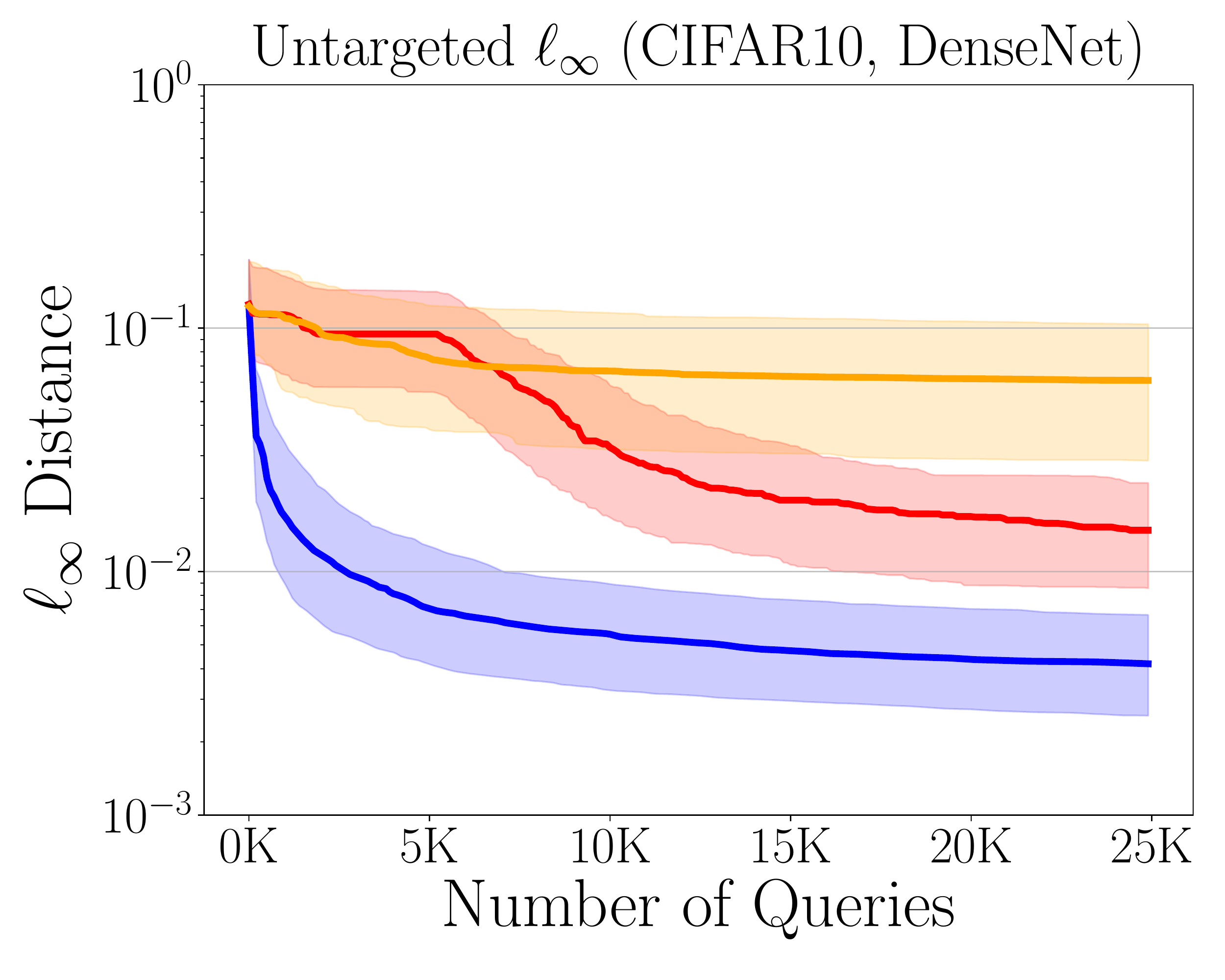} 
\includegraphics[width=0.23\linewidth]{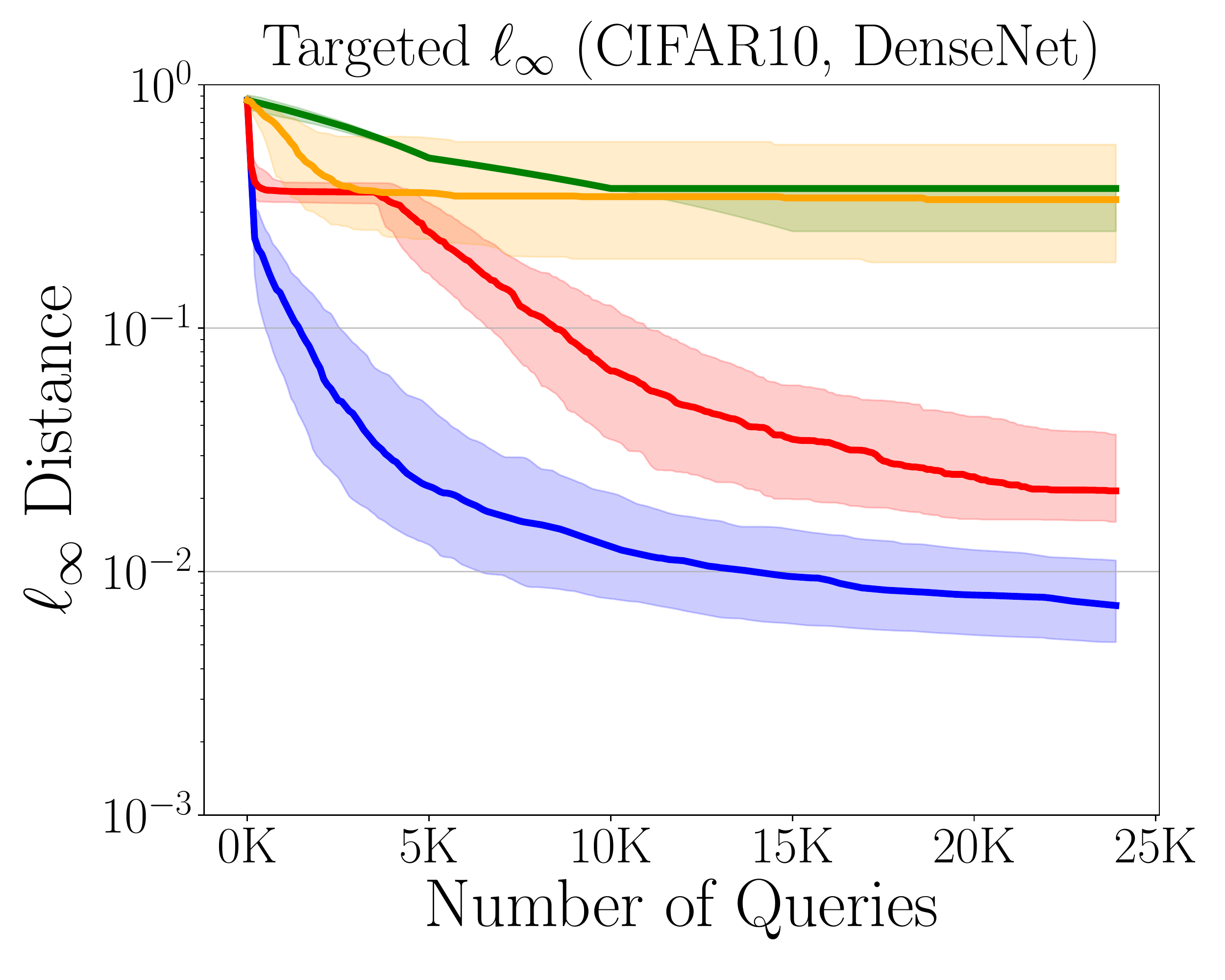}

\includegraphics[width=0.9\linewidth]{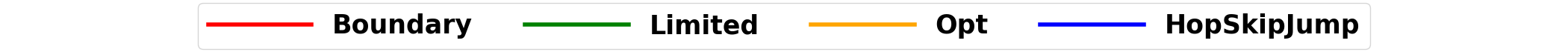}

\caption{Median distance versus number of model queries on MNIST with CNN, and CIFAR-10 with ResNet and DenseNet from top to
bottom rows.  1st column: untargeted $\ell_2$. 2nd col.: targeted
$\ell_2$. 3rd col.: untargeted $\ell_\infty$. 4th col.: targeted
$\ell_\infty$.}
\label{fig:queries}
\end{figure*}


\section{Experiments}\label{sec:exp}
In this section, we carry out experimental analysis of HopSkipJumpAttack.  
We compare the efficiency of HopSkipJumpAttack with {several previously proposed decision-based attacks} on image classification tasks. {In addition, we evaluate the robustness of three defense mechanisms under our attack method.}
All experiments were carried out on a Tesla K80 GPU, with code available online.\footnote{See \url{https://github.com/Jianbo-Lab/HSJA/}.} Our algorithm is also available on CleverHans~\cite{papernot2018cleverhans} and Foolbox~\cite{rauber2017foolbox}, which are two popular Python packages to craft adversarial examples for machine learning models.

\begin{figure*}[!bt]
\centering
\includegraphics[width=0.23\linewidth]{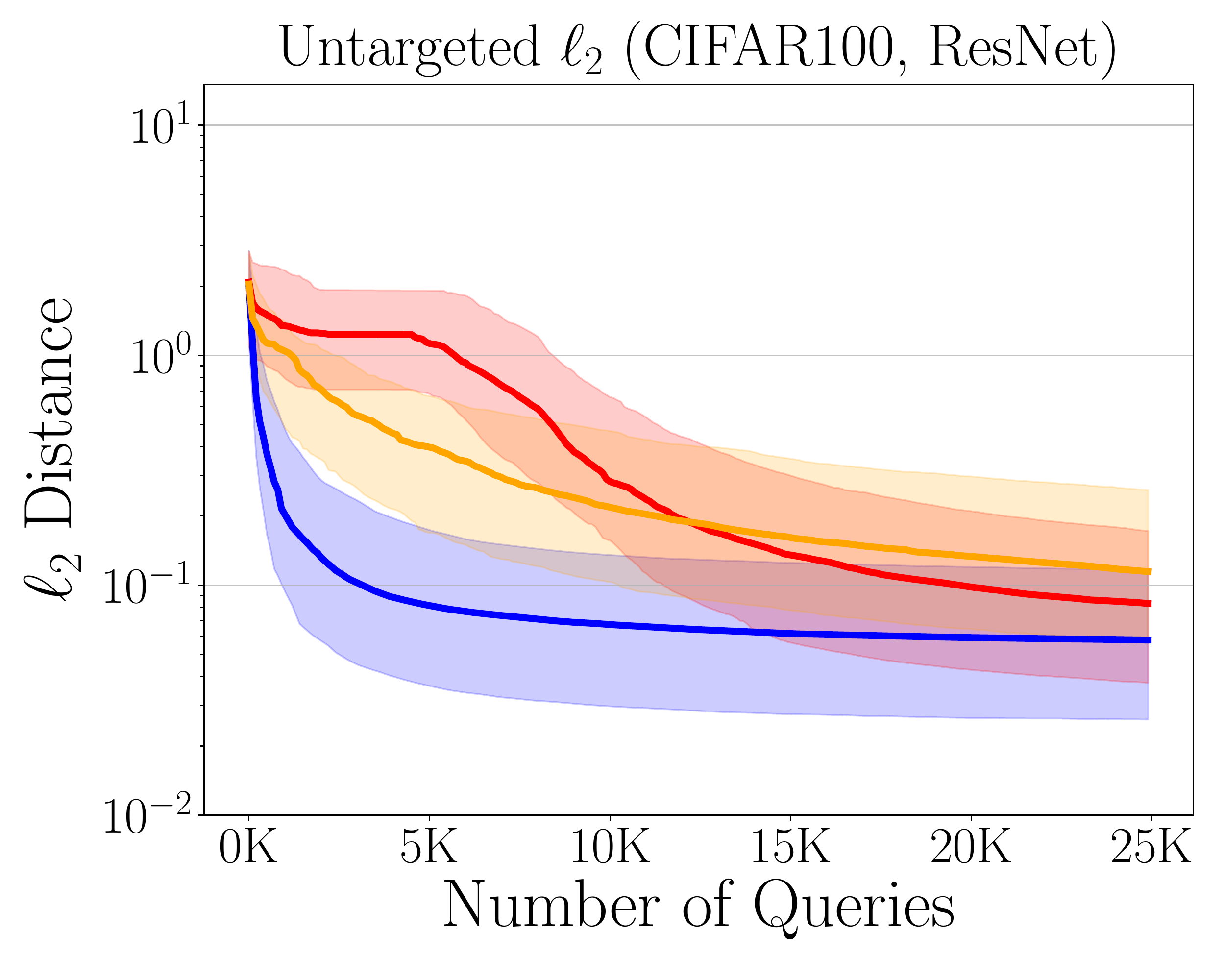}
\includegraphics[width=0.23\linewidth]{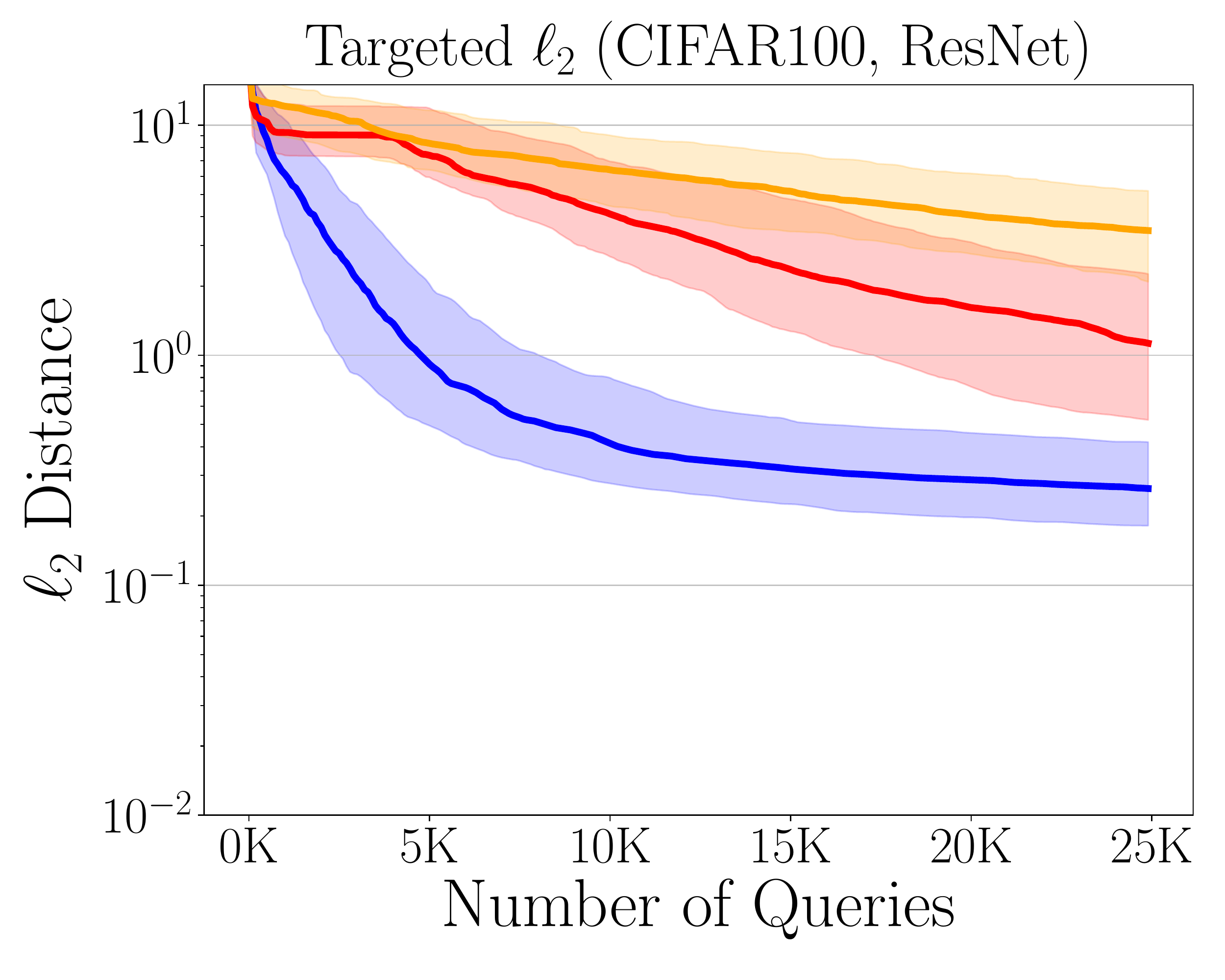}
\includegraphics[width=0.23\linewidth]{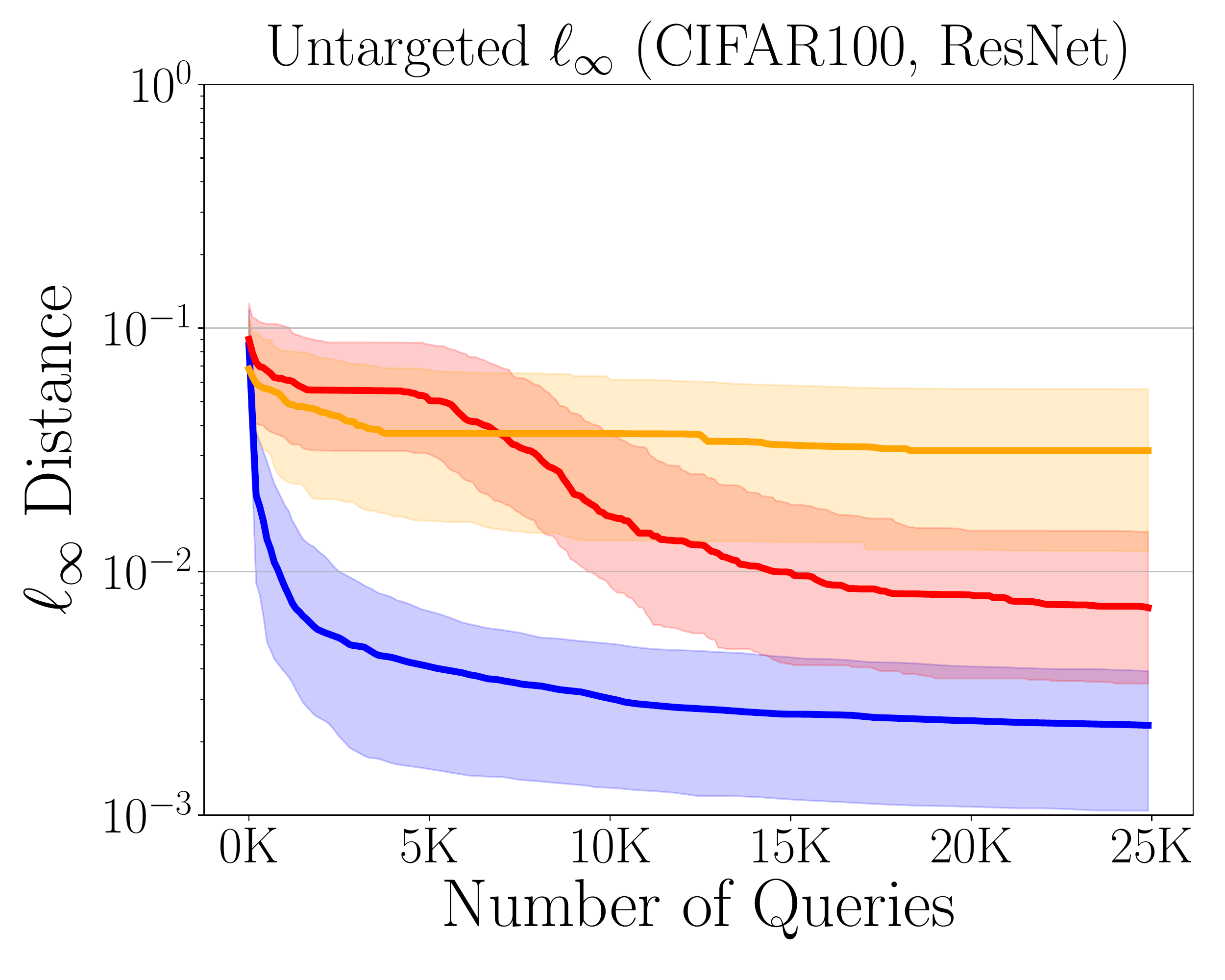} 
\includegraphics[width=0.23\linewidth]{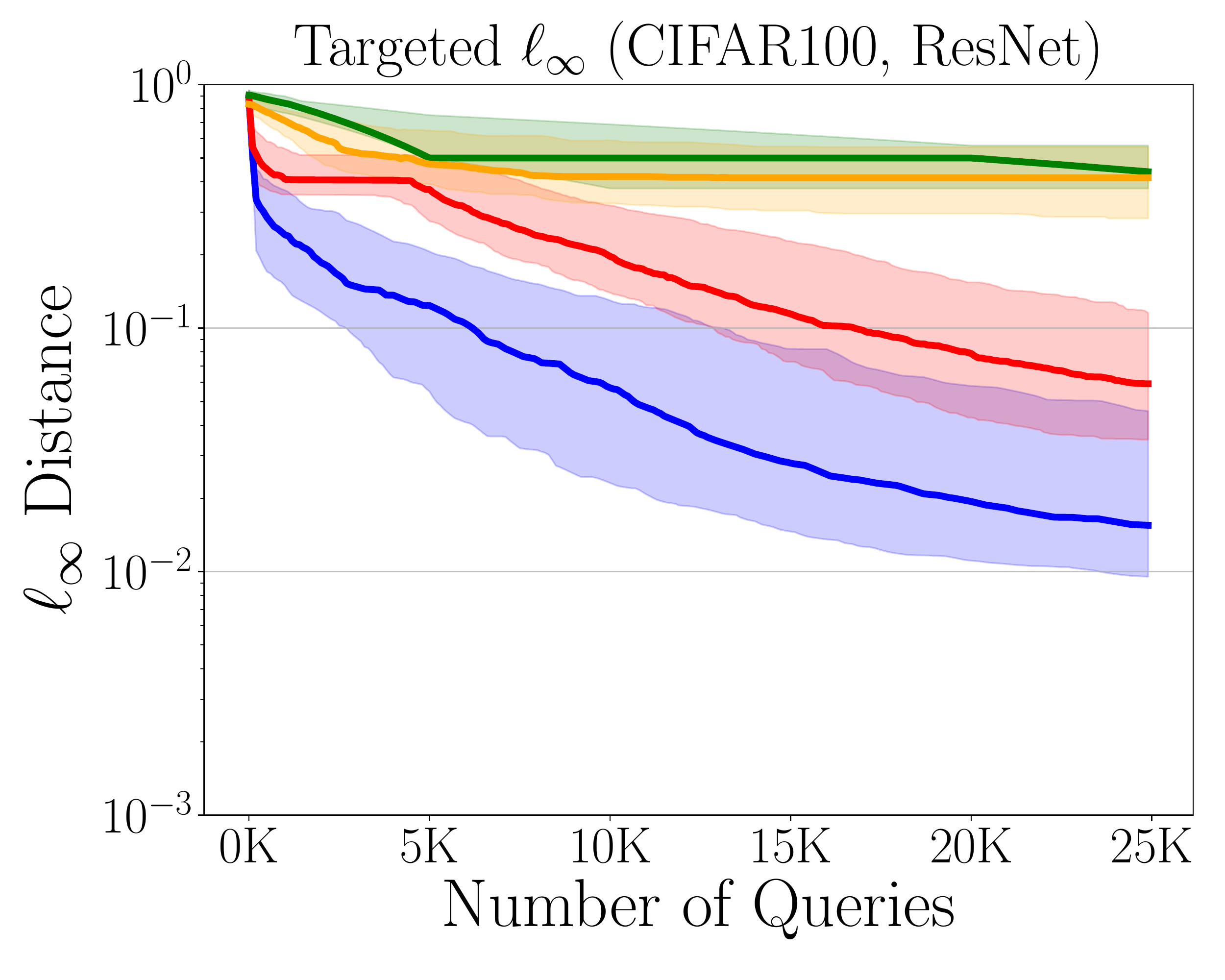}

\includegraphics[width=0.23\linewidth]{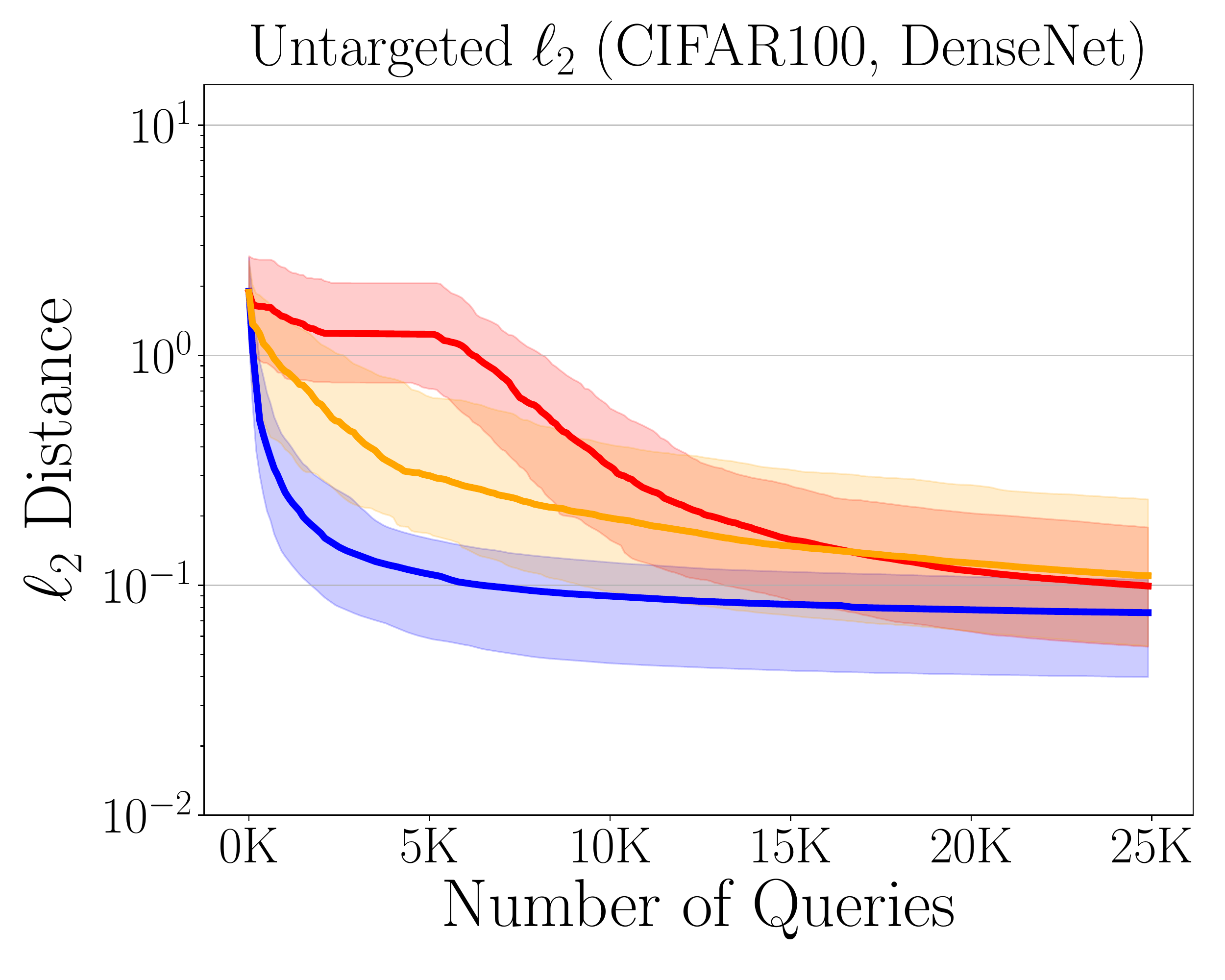} 
\includegraphics[width=0.23\linewidth]{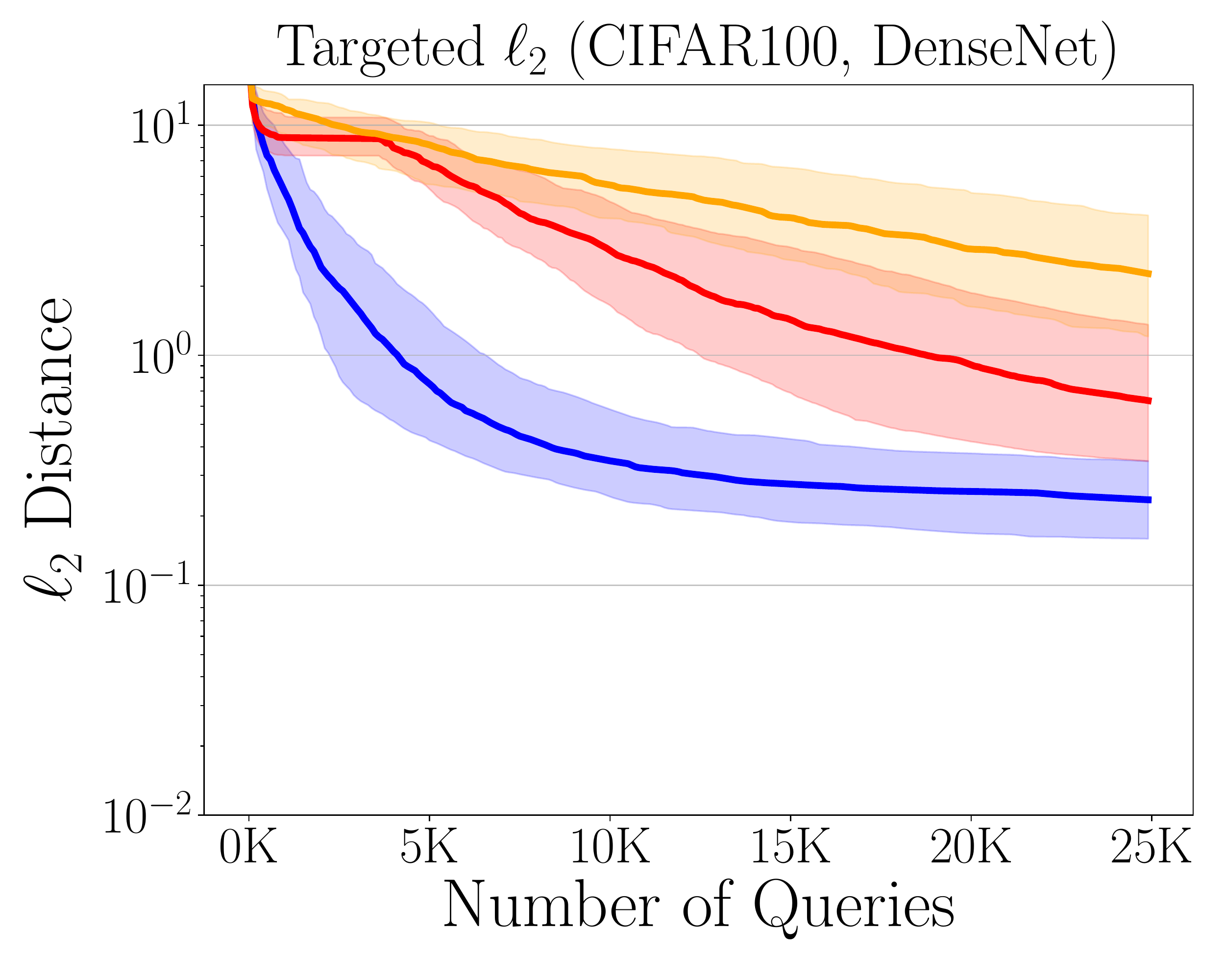} 
\includegraphics[width=0.23\linewidth]{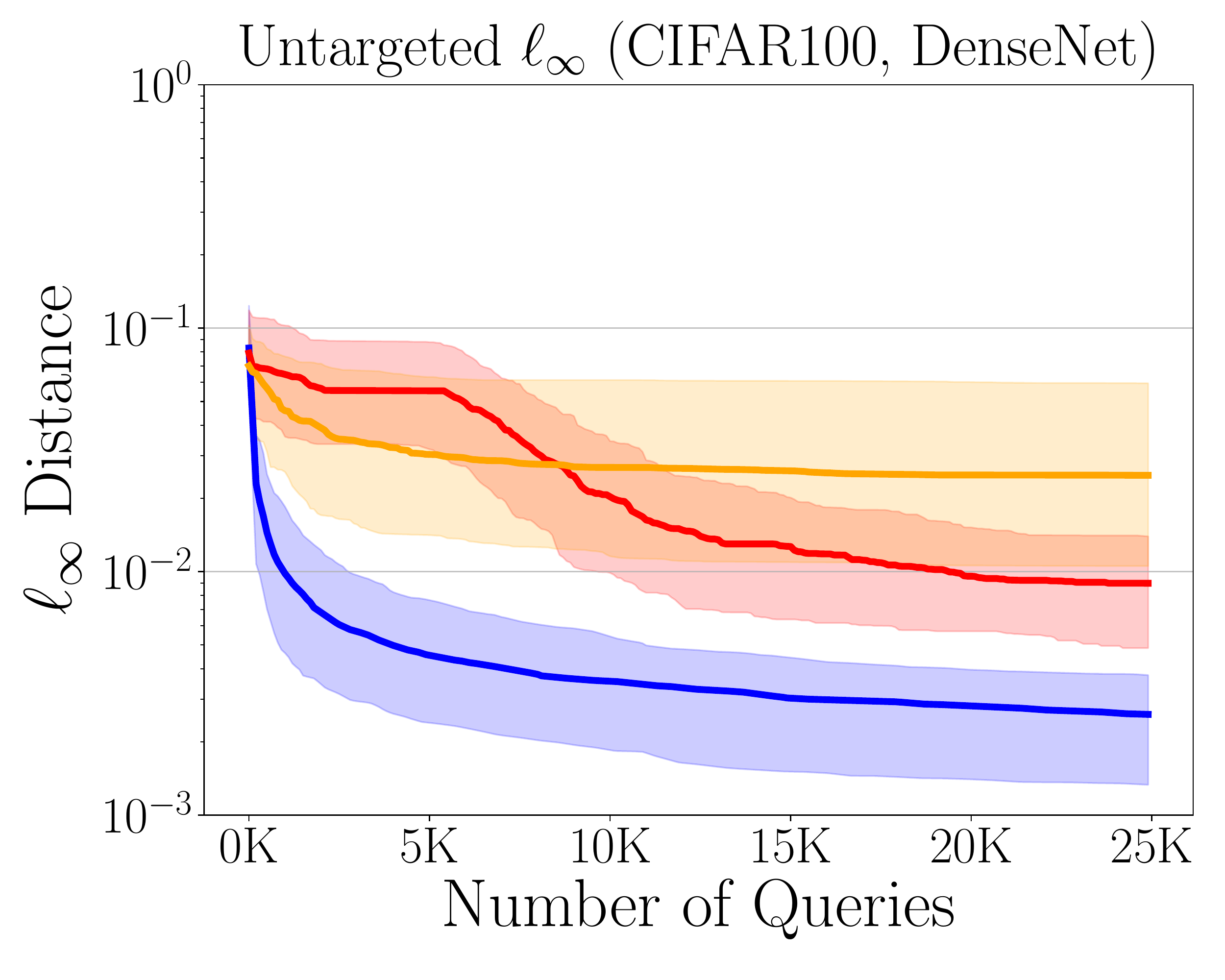} 
\includegraphics[width=0.23\linewidth]{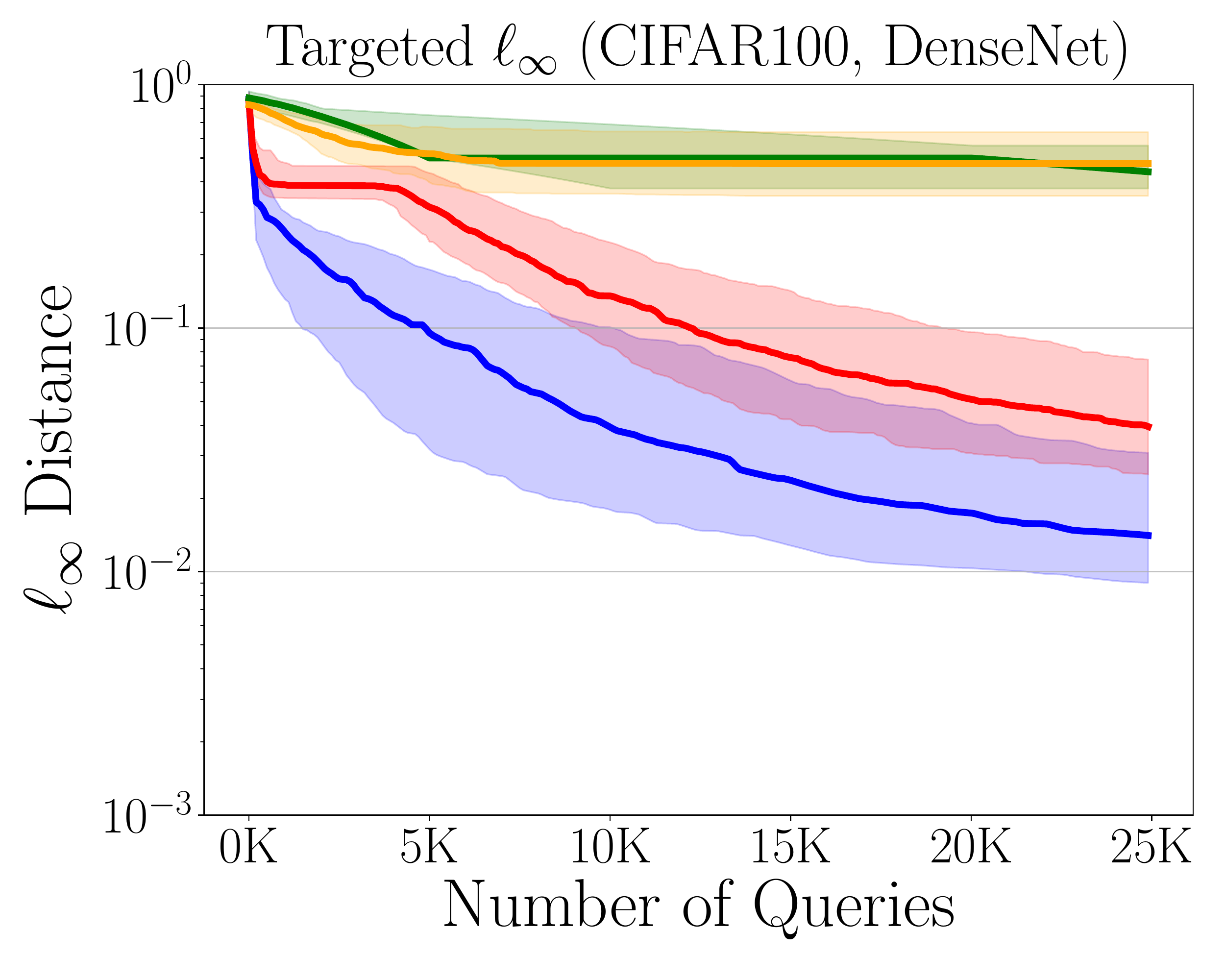} 

\includegraphics[width=0.23\linewidth]{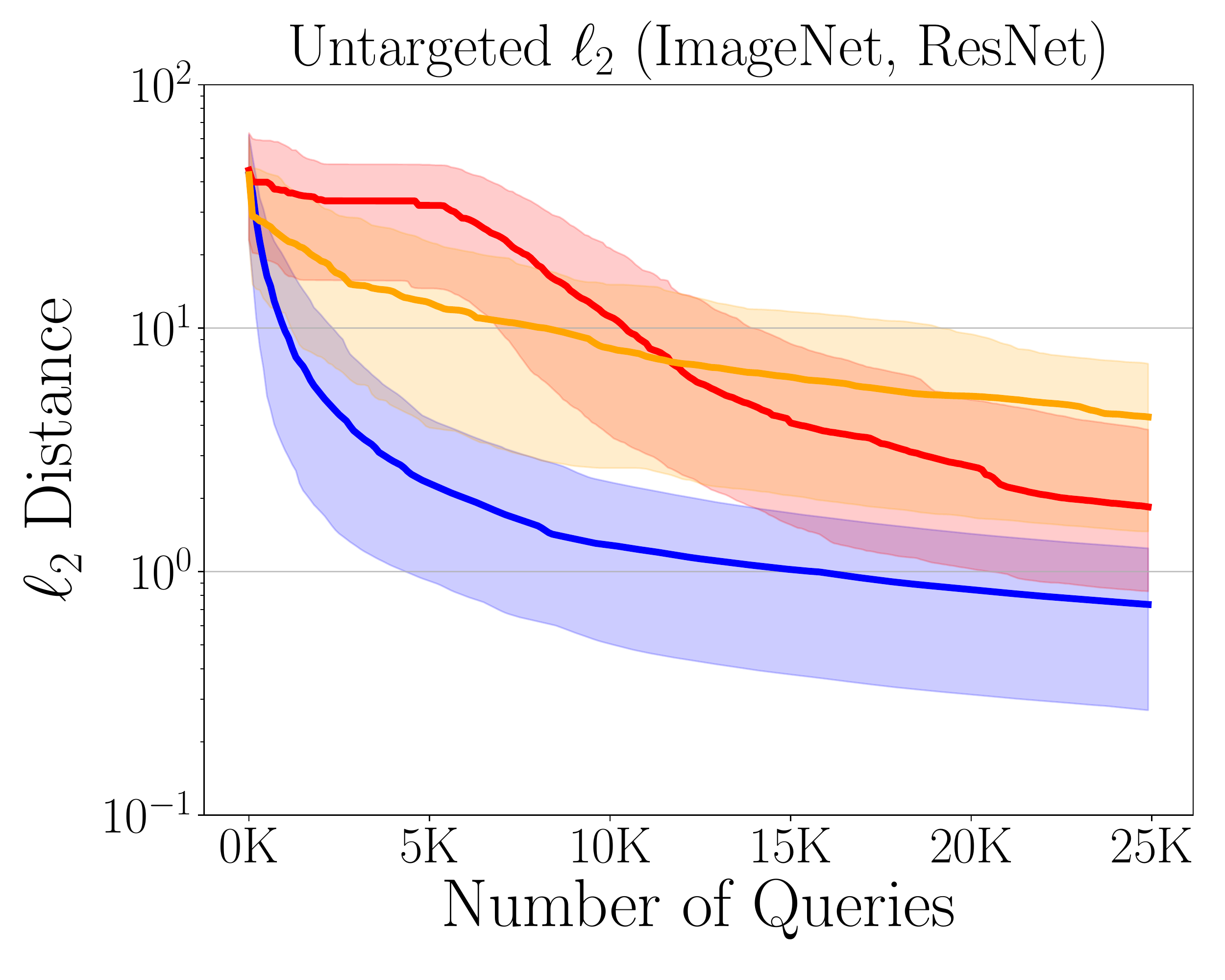} 
\includegraphics[width=0.23\linewidth]{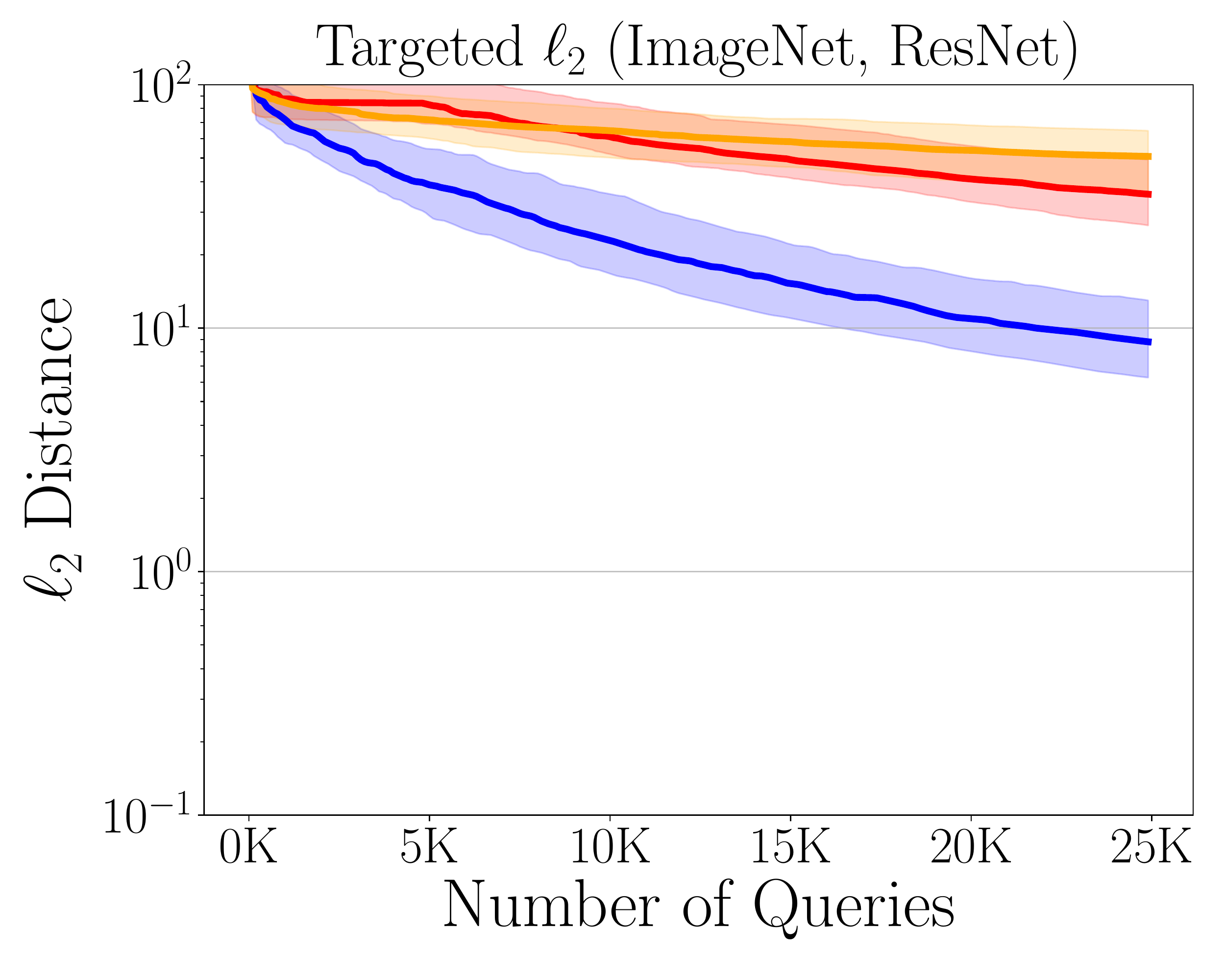} 
\includegraphics[width=0.23\linewidth]{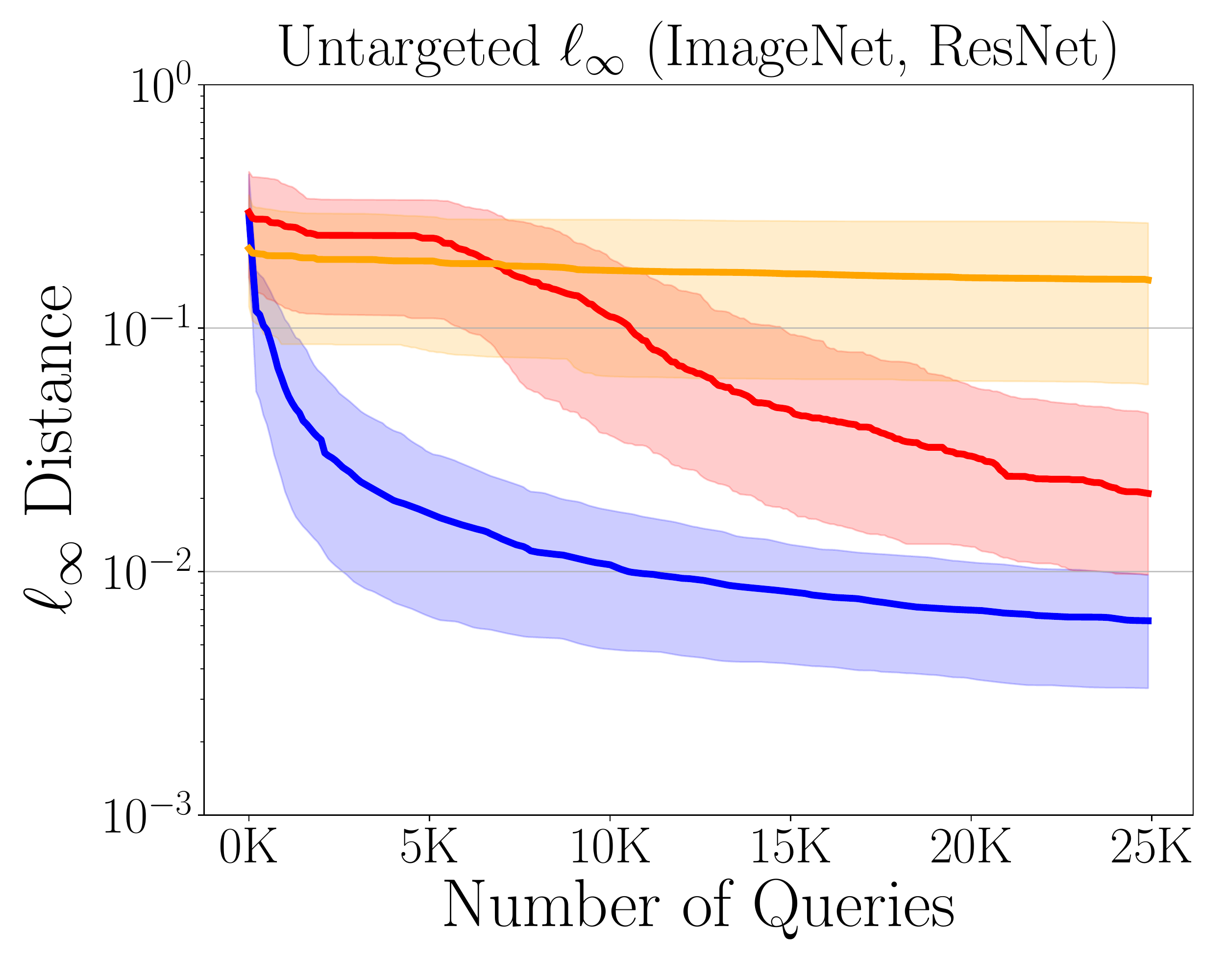} 
\includegraphics[width=0.23\linewidth]{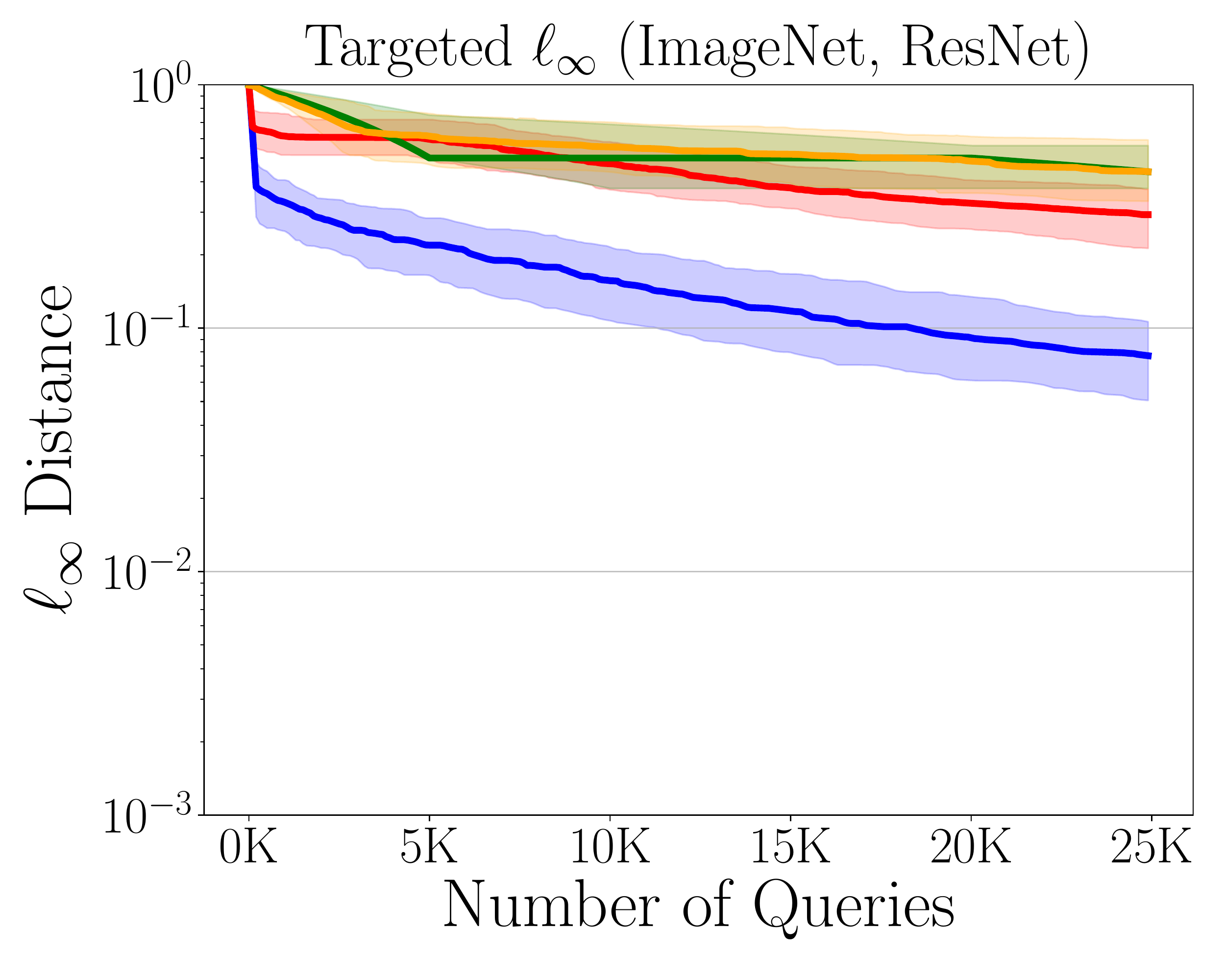} 

\includegraphics[width=0.9\linewidth]{figures_updated/legend-norm_vs_attack.png}

\caption{Median distance versus number of model queries on CIFAR-100 with ResNet, DenseNet, and ImageNet with ResNet from top to bottom rows. 1st column: untargeted $\ell_2$. 2nd col.: targeted
$\ell_2$. 3rd col.: untargeted $\ell_\infty$. 4th col.: targeted
$\ell_\infty$.}
\label{fig:queries2}
\end{figure*}
\subsection{Efficiency evaluation}\label{sec:efficiency}
\paragraph{Baselines} {We compare HopSkipJumpAttack with three state-of-the-art decision-based attacks: {Boundary Attack~\cite{brendel2018decisionbased}, Limited Attack~\cite{ilyas2018black} and Opt Attack~\cite{cheng2018queryefficient}. We use the implementation of the three algorithms with the suggested hyper-parameters from the publicly available source code online. Limited Attack is only included under the targeted $\ell_\infty$ setting, as in~\citet{ilyas2018black}.}
}
\begin{figure*}[!bt]
\centering
\includegraphics[width=0.23\linewidth]{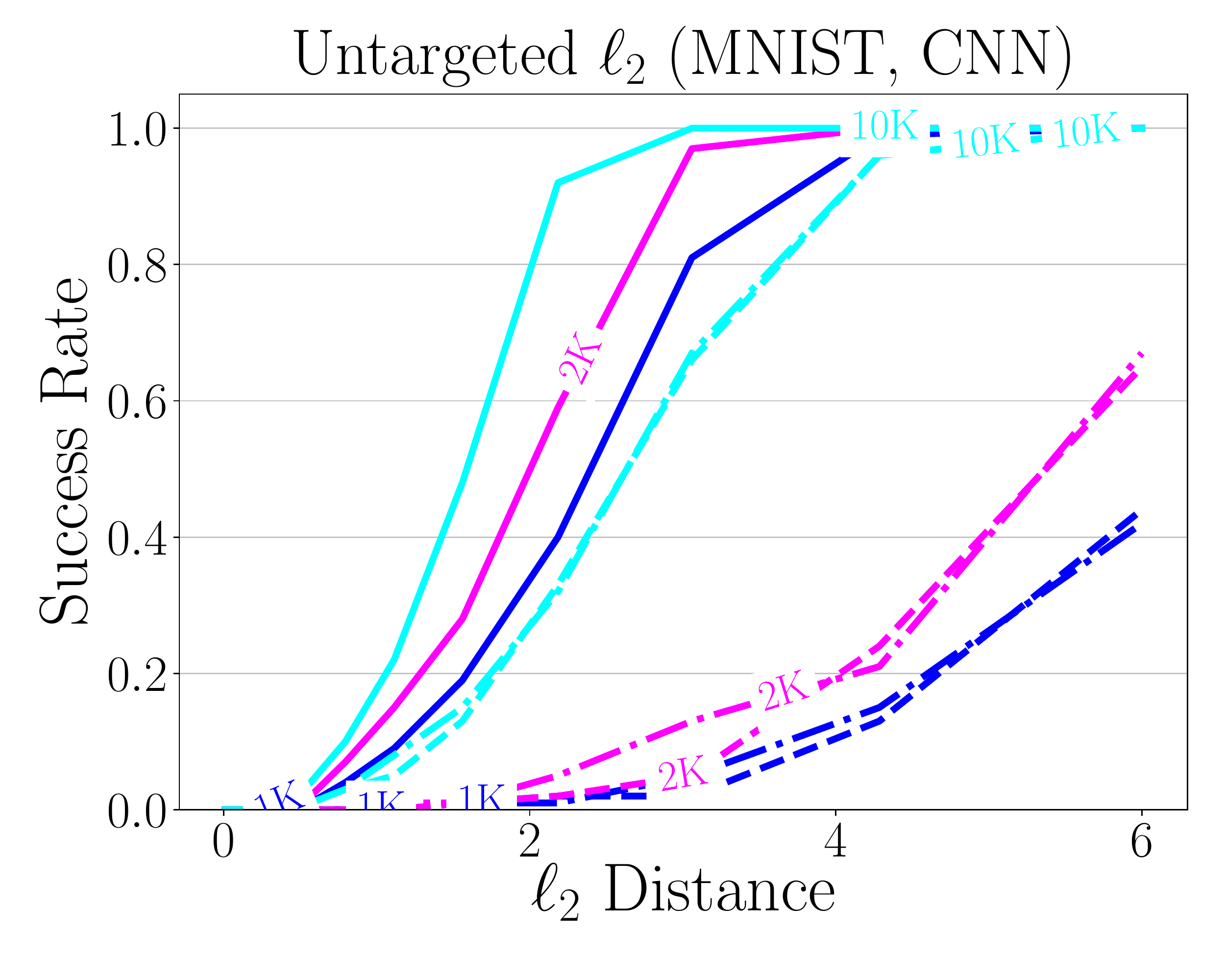}
\includegraphics[width=0.23\linewidth]{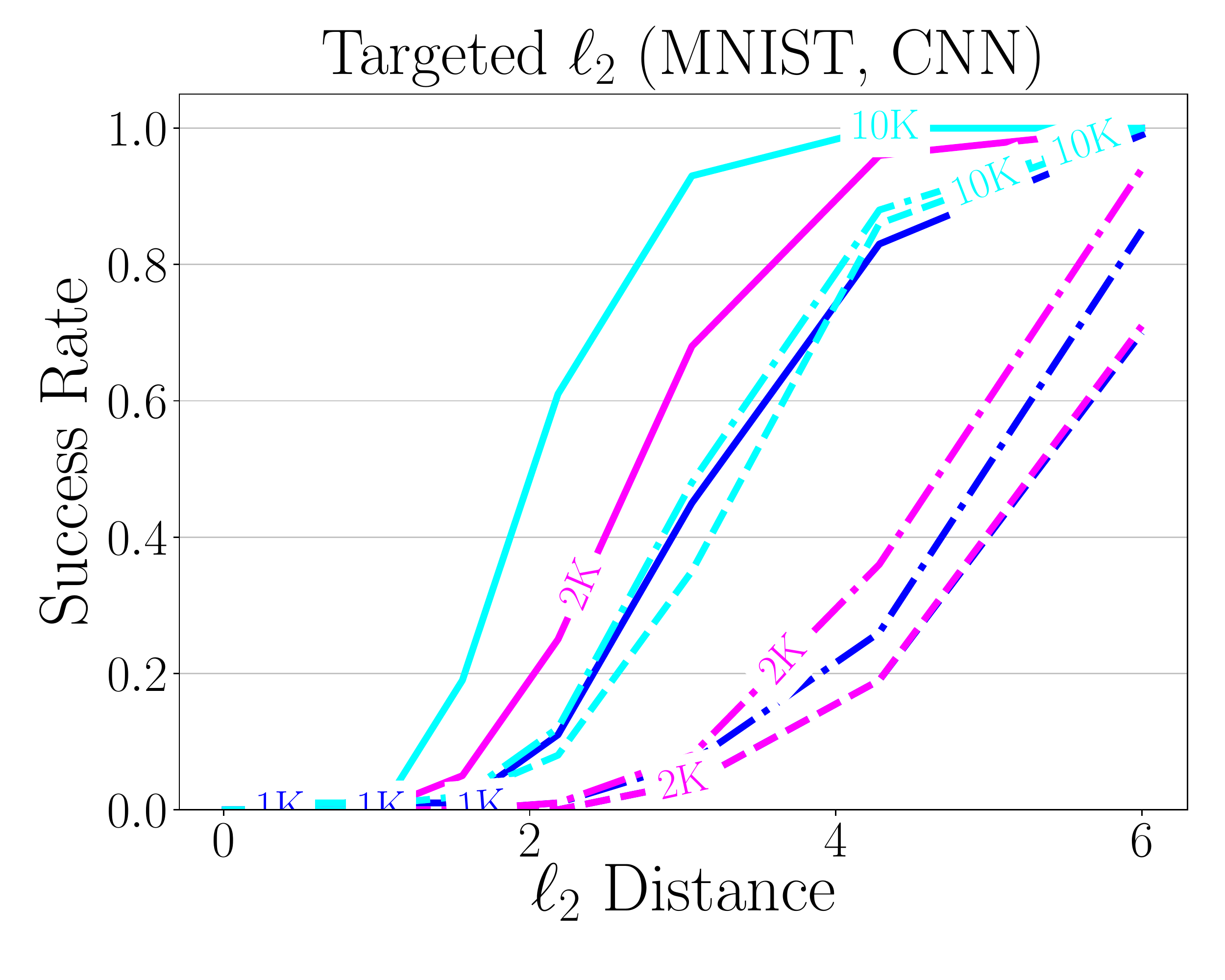} 
\includegraphics[width=0.23\linewidth]{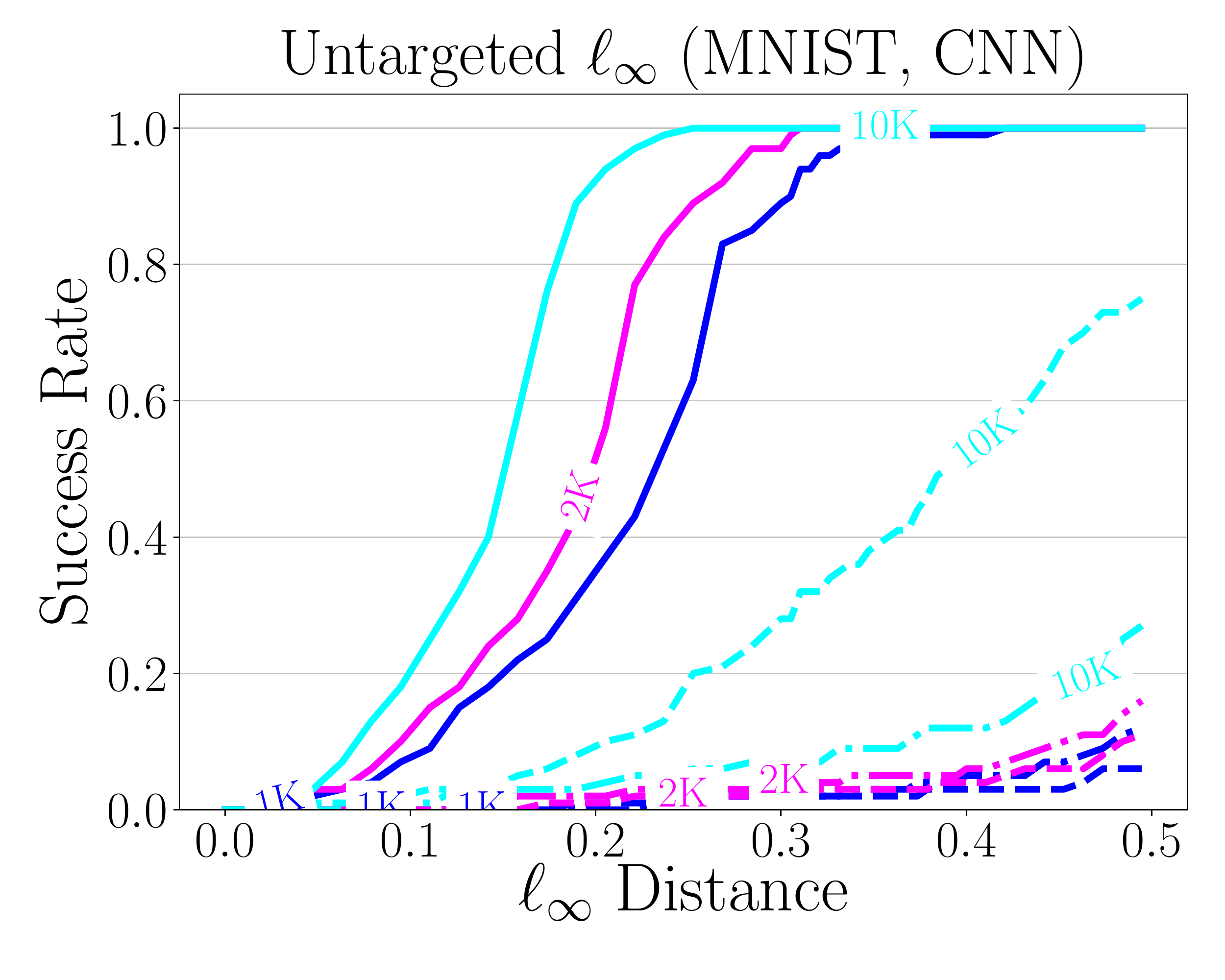}
\includegraphics[width=0.23\linewidth]{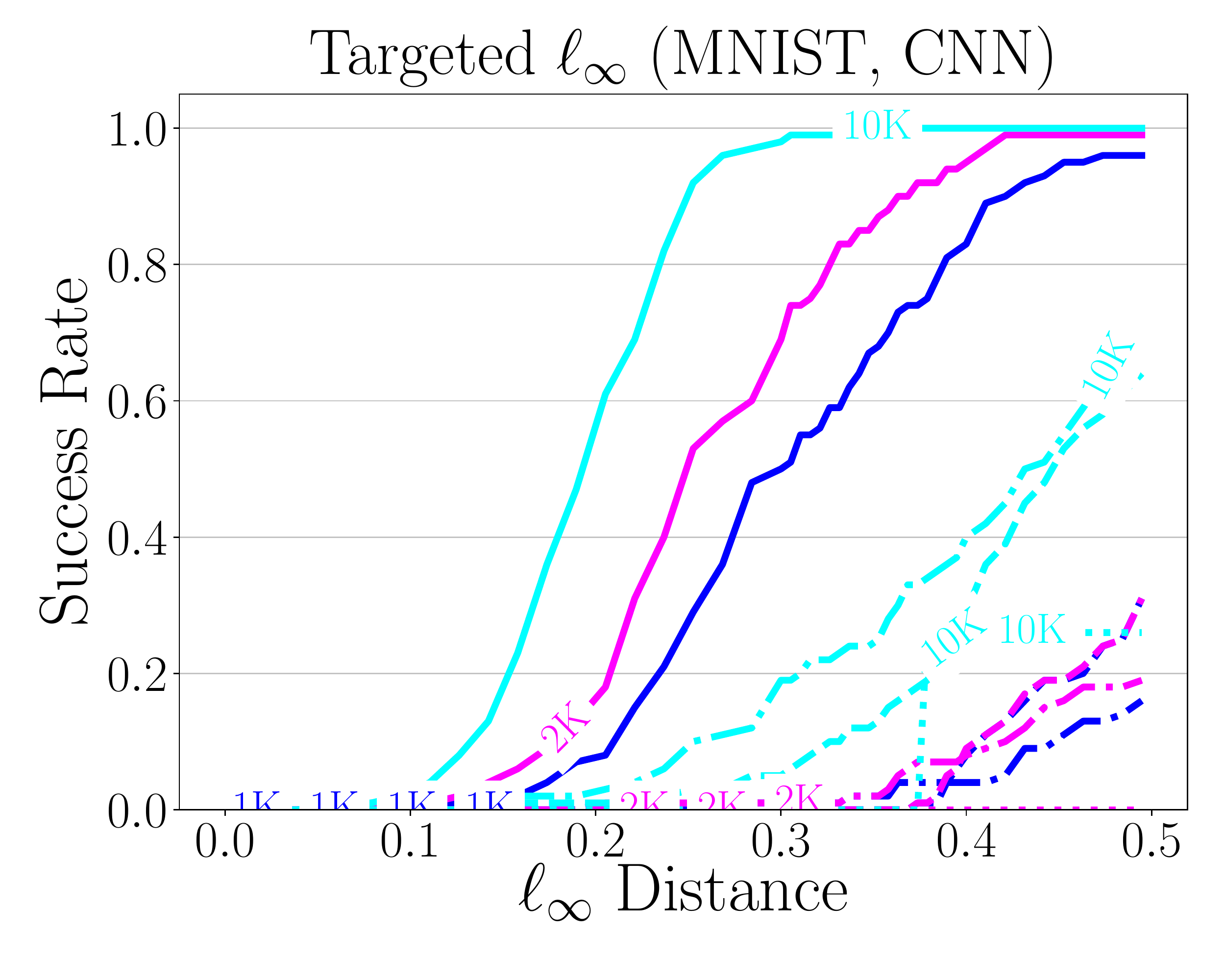} 

\includegraphics[width=0.23\linewidth]{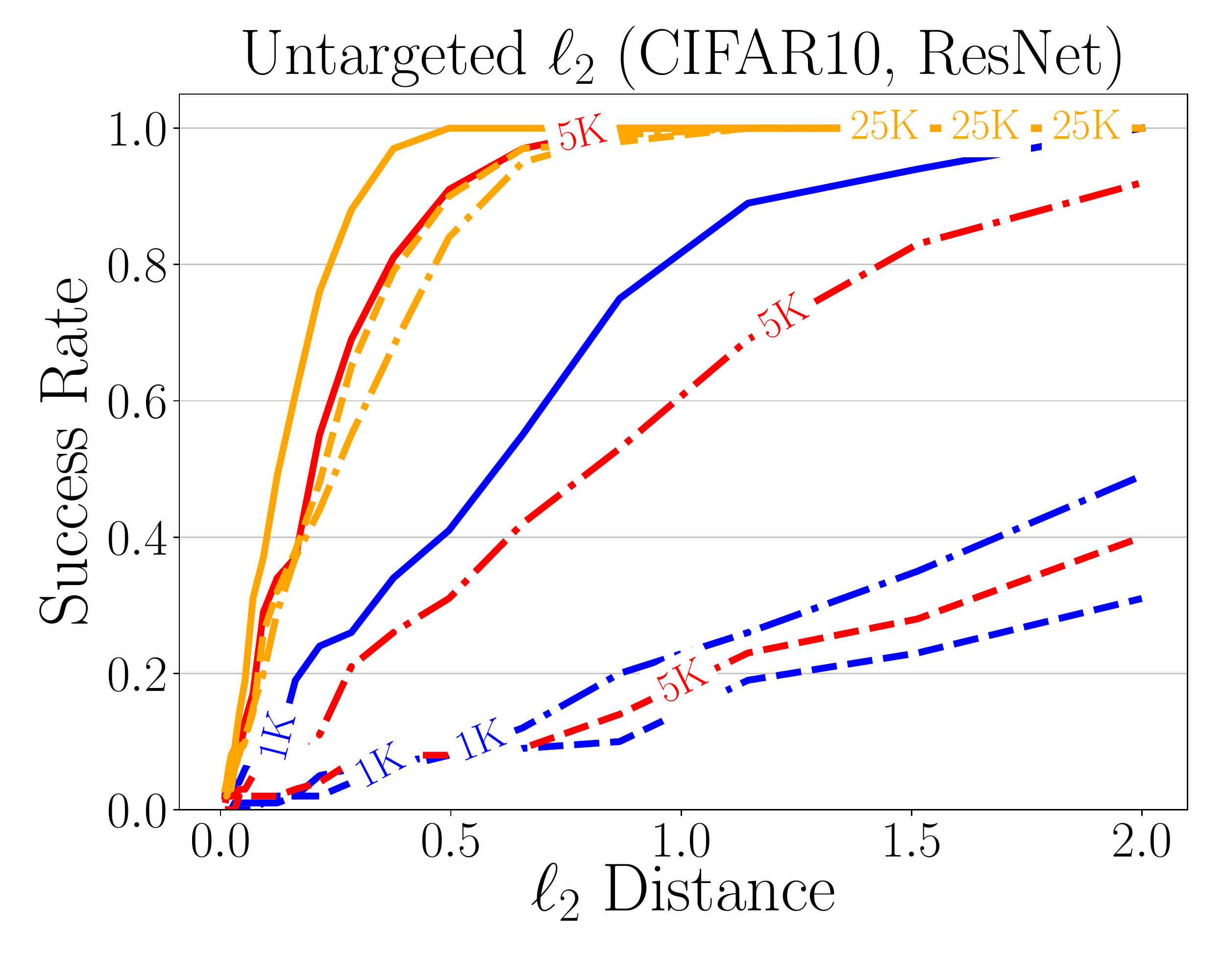} 
\includegraphics[width=0.23\linewidth]{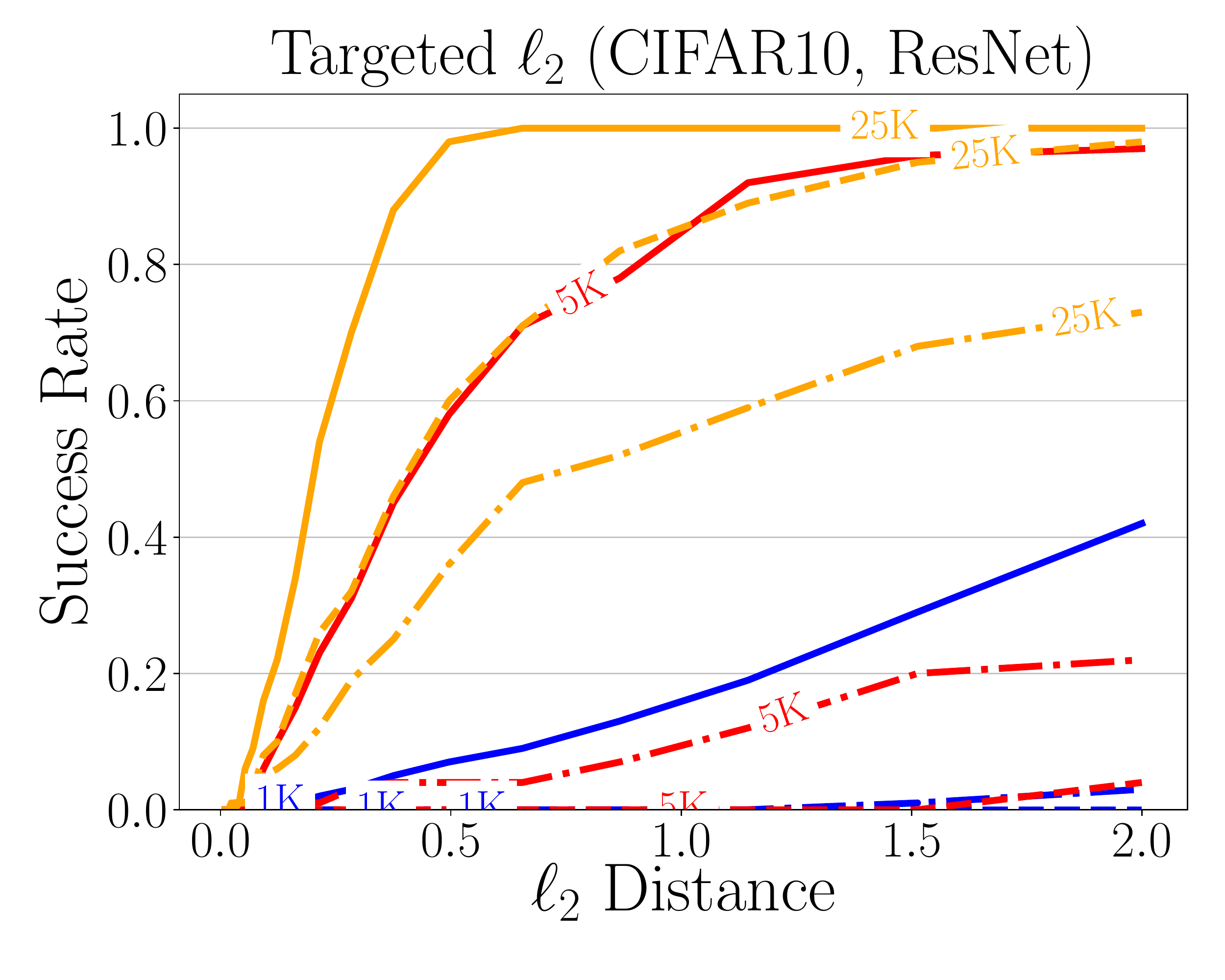}
\includegraphics[width=0.23\linewidth]{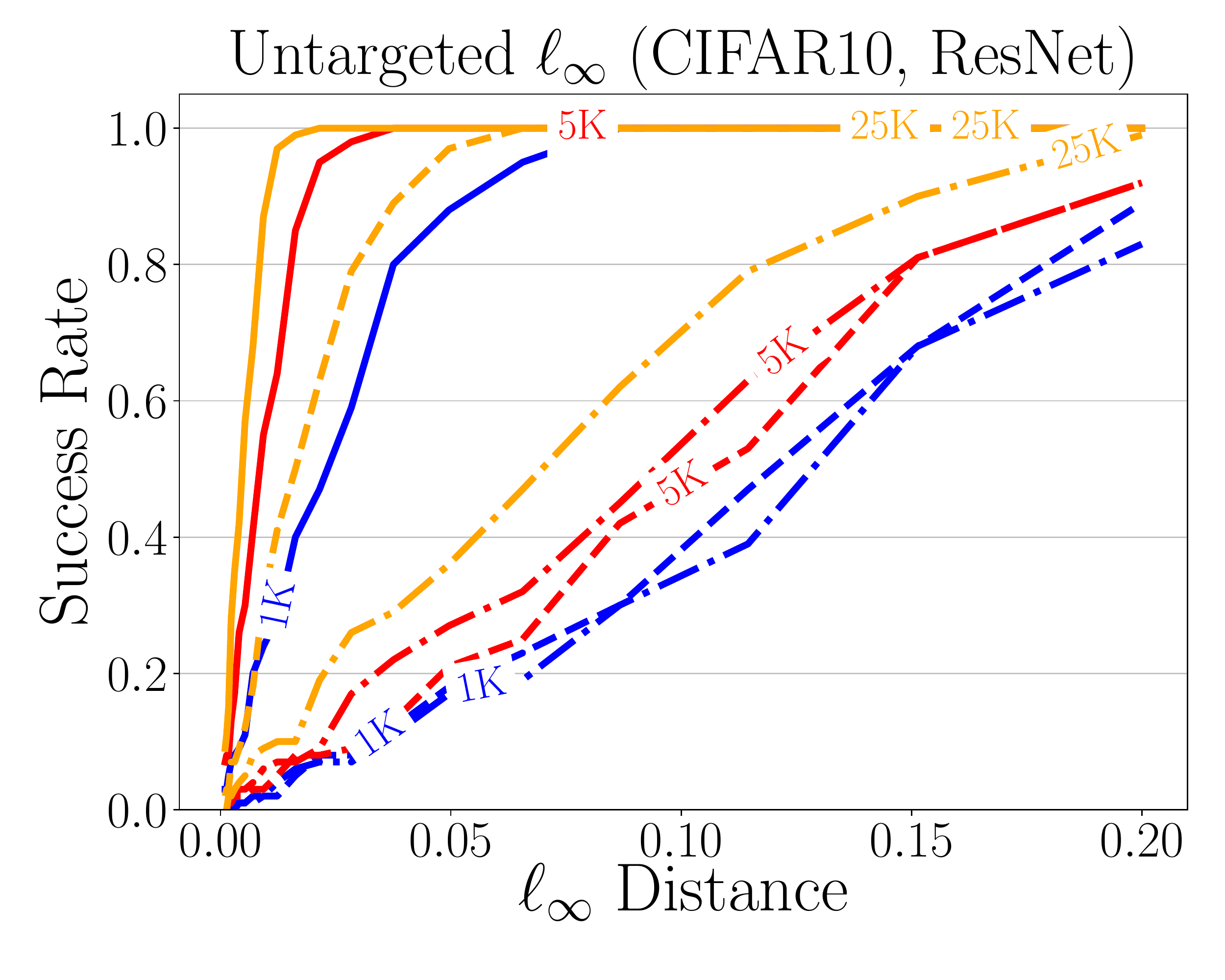} 
\includegraphics[width=0.23\linewidth]{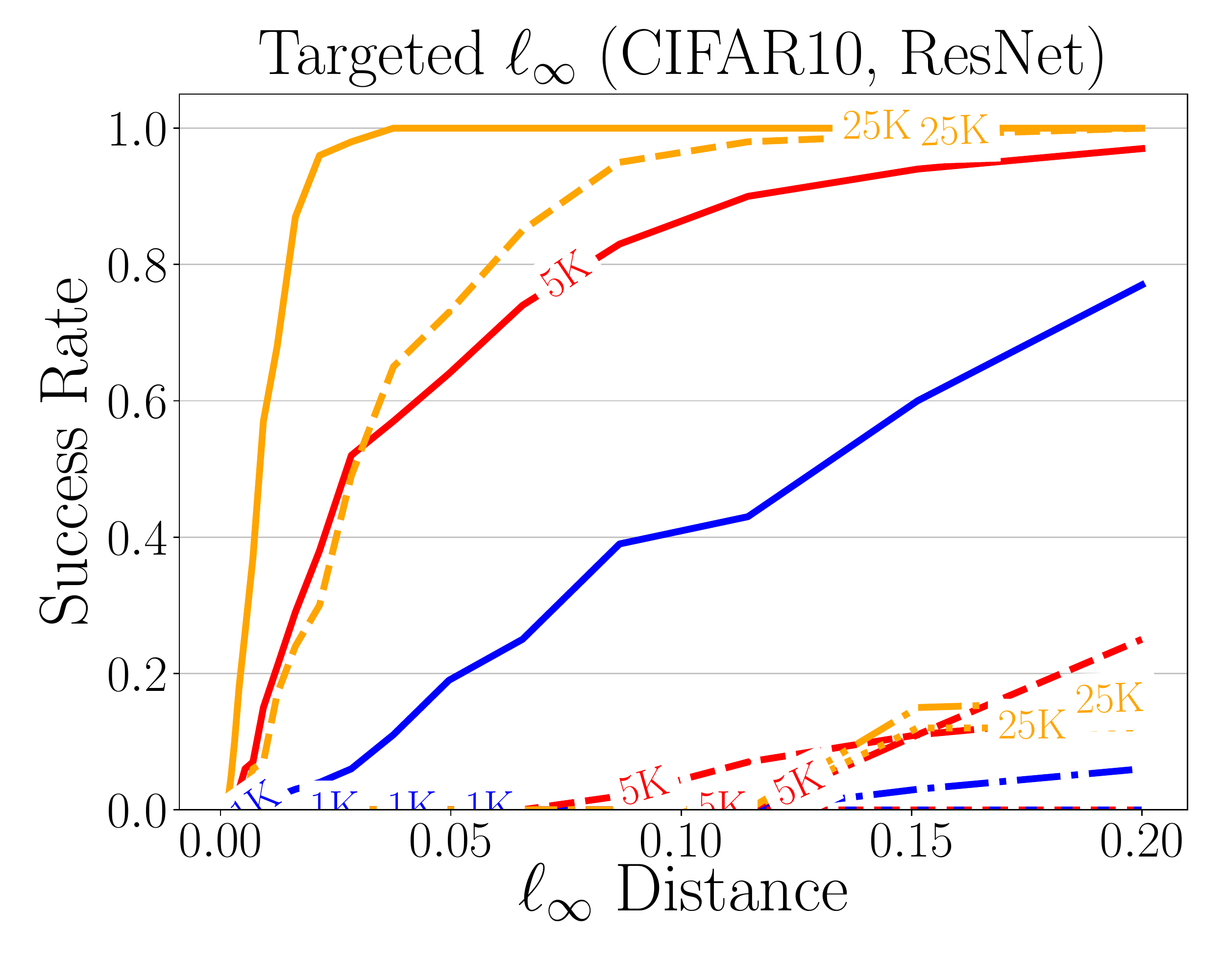}

\includegraphics[width=0.23\linewidth]{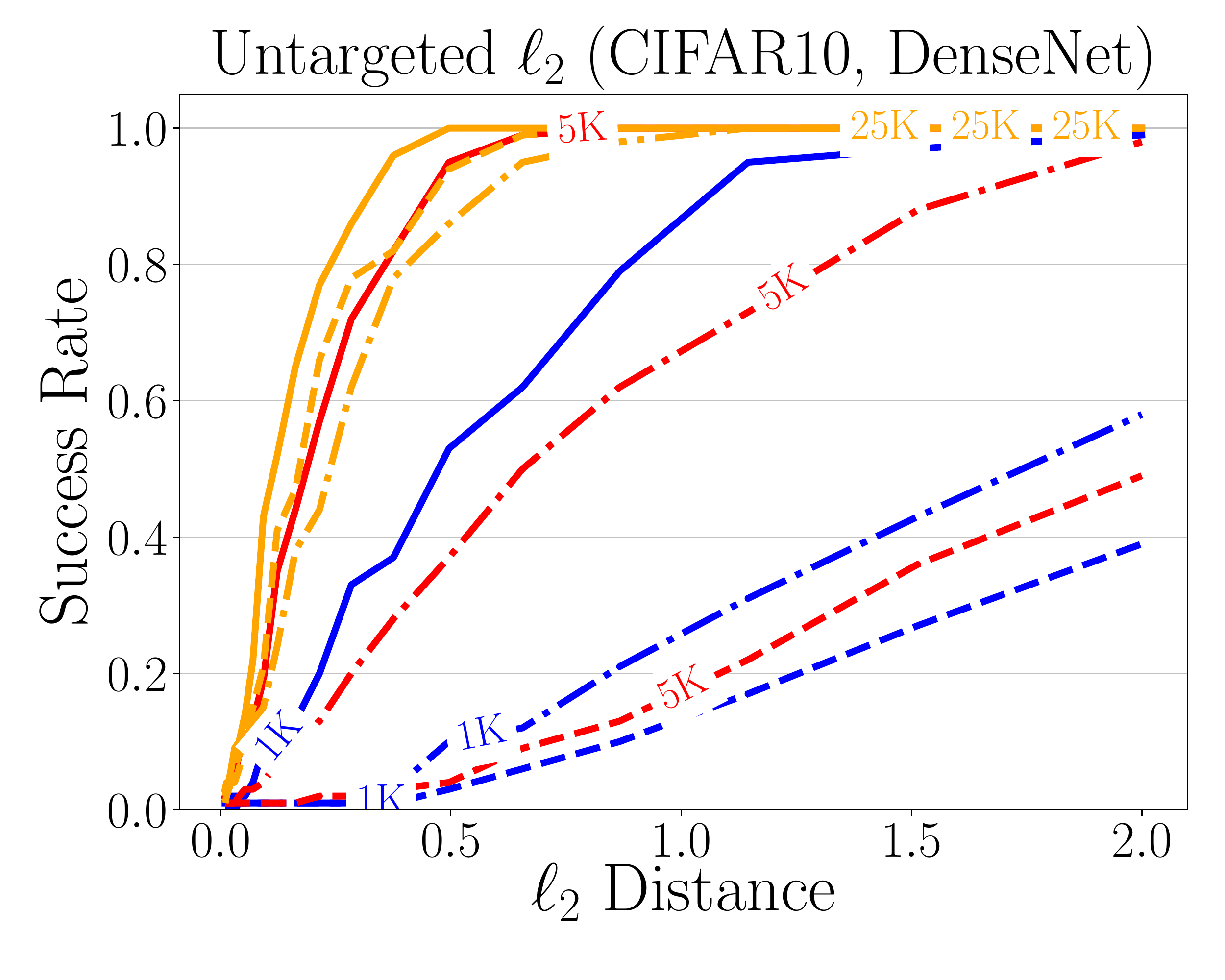} 
\includegraphics[width=0.23\linewidth]{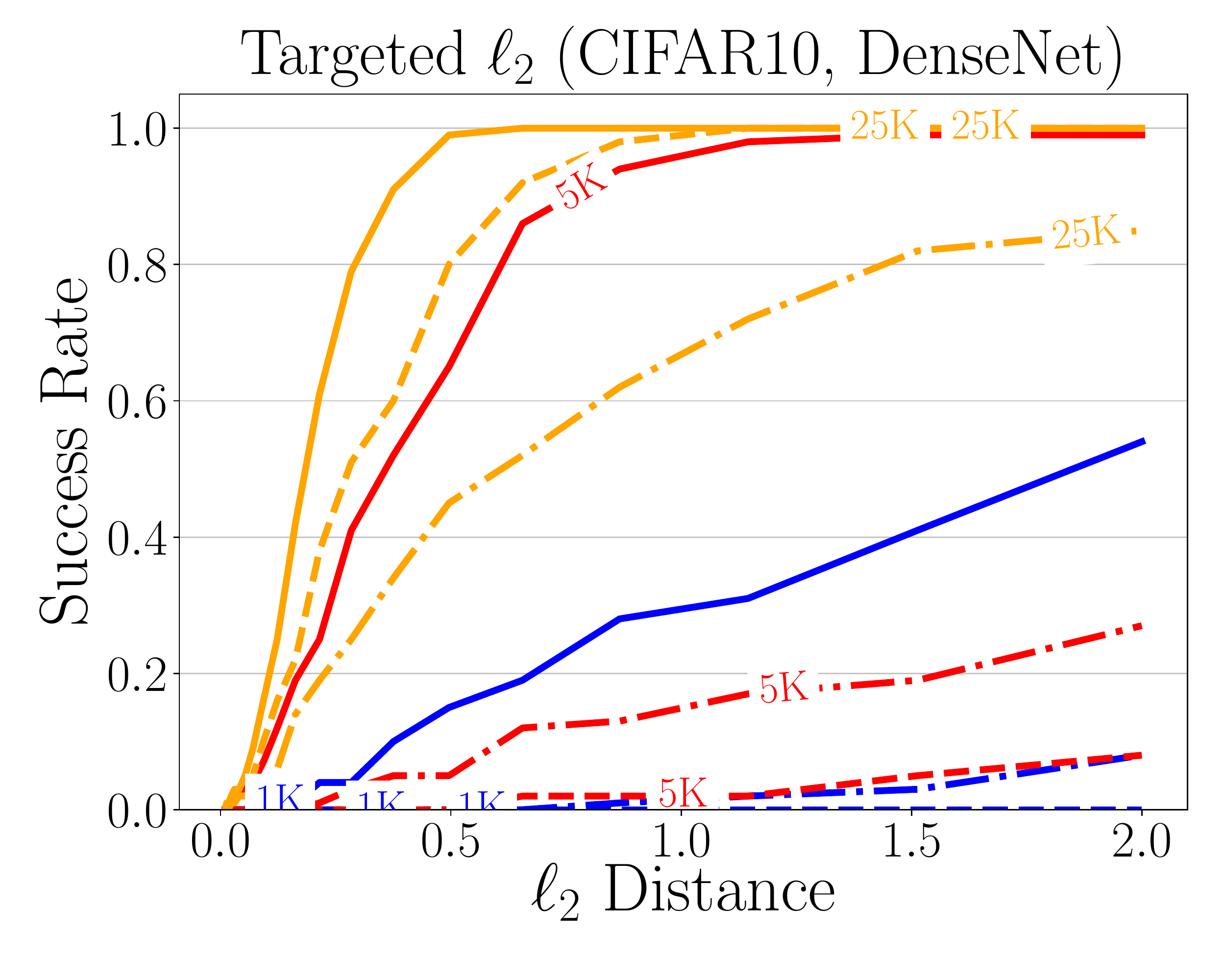} 
\includegraphics[width=0.23\linewidth]{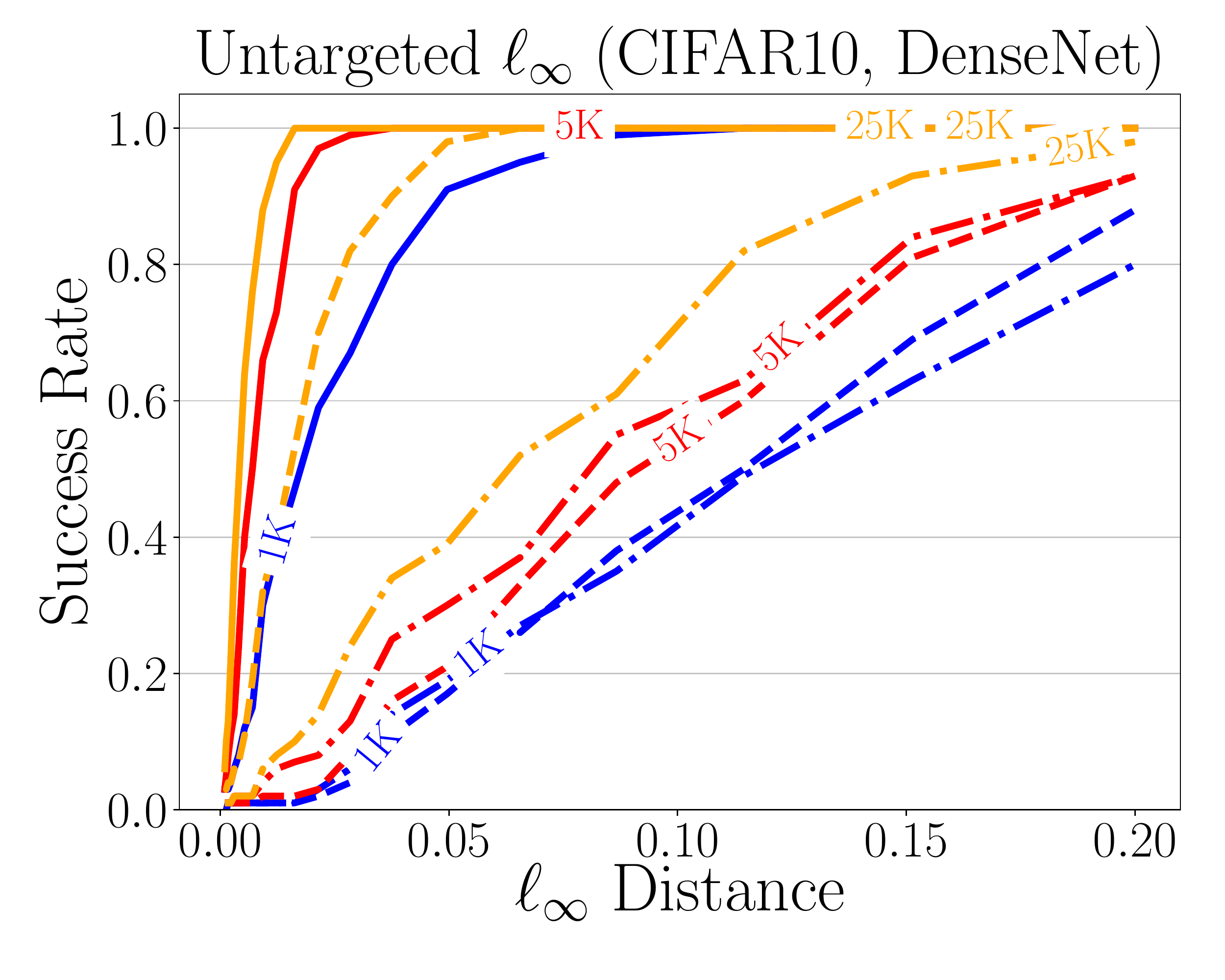} 
\includegraphics[width=0.23\linewidth]{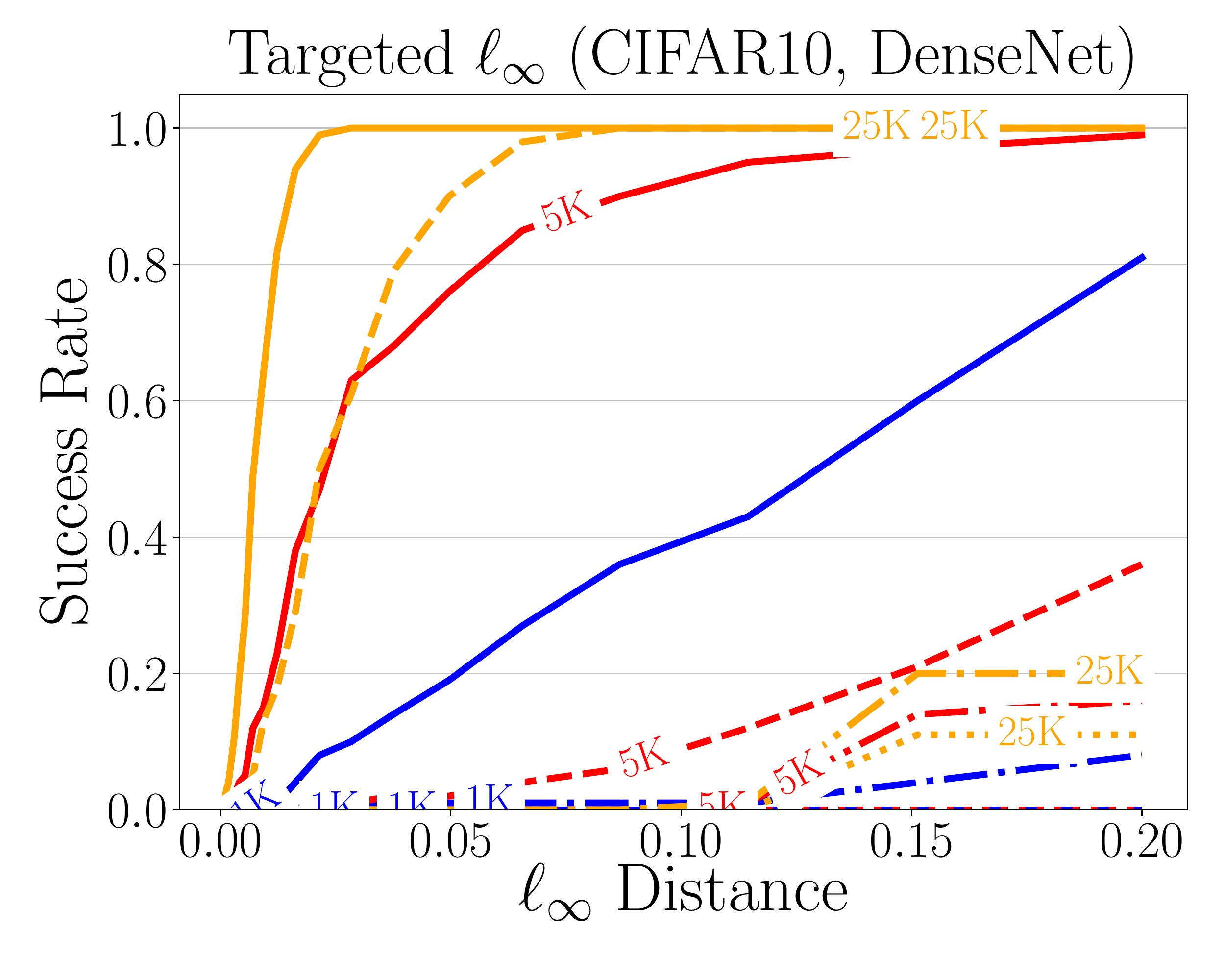} 

\includegraphics[width=0.9\linewidth]{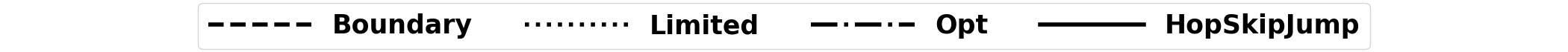}
\caption{Success rate versus
distance threshold for MNIST with CNN, and CIFAR-10 with ResNet, DenseNet from top to bottom rows. 1st column:
untargeted $\ell_2$. 2nd column: targeted $\ell_2$. 3rd column:
untargeted $\ell_\infty$. 4th column: targeted $\ell_\infty$.}
\label{fig:success}
\end{figure*} 
\paragraph{Data and models} For a comprehensive evaluation of HopSkipJumpAttack, we use a wide range of data and models, with varied image dimensions, data set sizes, complexity levels of task and model structures.

The experiments are carried out over four image data sets:
MNIST, CIFAR-10~\cite{krizhevsky2009learning}, CIFAR-100~\cite{krizhevsky2009learning}, and ImageNet~\cite{imagenet_cvpr09} with the standard train/test split~\cite{chollet2015keras}. The four data sets have varied image dimensions and class numbers.
MNIST contains 70K $28\times 28$ gray-scale images of handwritten digits in the range 0-9. CIFAR-10 and CIFAR-100 are both composed of $32\times 32\times 3$ images. CIFAR-10 has 10 classes, with 6K images per class, while CIFAR-100 has 100 classes, with 600 images per class. ImageNet has $1,000$ classes. Images in ImageNet are rescaled to $224\times 224\times 3$. For MNIST, CIFAR-10 and CIFAR-100, $1,000$ correctly classified test images are used, which are randomly drawn from the test data set, and evenly distributed across classes. For ImageNet, we use $100$ correctly classified test images, evenly distributed among $10$ randomly selected classes. The selection scheme follows \citet{metzen2017detecting} for reproducibility. 

We also use models of varied structure, from simple to complex.
For MNIST, we use a simple convolutional network composed of two convolutional layers
followed by a hidden dense layer with $1024$ units. Two convolutional
layers have $32, 64$ filters respectively, each of which is followed
by a max-pooling layer. For both CIFAR-10 and CIFAR-100, we train a
$20$-layer ResNet~\cite{he2016identity} and $121$-layer
DenseNet~\cite{huang2017densely} respectively, with the canonical network structure~\cite{chollet2015keras}.  For ImageNet, we use a pre-trained $50$-layer ResNet~\cite{he2016identity}. All models achieve close to state-of-the-art 
accuracy on the respective data set. All pixels are scaled
to be in the range $[0, 1]$. For all experiments, we clip the perturbed image into the input domain $[0,1]$ for all algorithms by default. 

\paragraph{Initialization} For untargeted attack, we initialize all attacks by blending an original image
with uniform random noise, and increasing the weight of uniform noise
gradually until it is misclassified, a procedure which is available on Foolbox~\cite{rauber2017foolbox}, as the default
initialization of Boundary Attack. For targeted attack, the target
class is sampled uniformly among the incorrect labels. {An image belonging to the target class is randomly sampled from the test set as the initialization.} The same target class and a common initialization image are used for all attacks.

\paragraph{Metrics} The first metric is the median $\ell_p$ distance
between perturbed and original samples over a subset of test
images, which was commonly used in previous work, such as~\citet{carlini2017towards}. A version normalized by image
dimension was employed by \citet{brendel2018decisionbased} for
evaluating Boundary Attack. The $\ell_2$ distance can be interpreted in the following way: Given a byte image of size $h\times w\times 3$, perturbation of size $d$ in $\ell_2$ distance on the rescaled input image amounts to perturbation on the original image of $\lceil d/\sqrt{h\times w \times 3} * 255\rceil$ bits per pixel on average, in the range $[0,255]$. The perturbation of size $d$ in $\ell_\infty$ distance amounts to a maximum perturbation of $\lceil255\cdot d\rceil$ bits across all pixels on the raw image. 

As an alternative metric, we also plot the
success rate at various distance thresholds for both algorithms given
a limited budget of model queries. An adversarial example is
defined a success if the size of perturbation does not exceed a
given distance threshold. The success rate can be directly related to the accuracy of a model on perturbed data under a given distance threshold:
\begin{align}
\text{perturbed acc.} = \text{original acc.} \times (1-\text{success rate}).
\end{align}
Throughout the experiments, we limit the maximum budget of queries per image to 25,000, the setting of practical interest, due to limited computational resources.

\begin{figure*}[!bt]
\centering
\includegraphics[width=0.23\linewidth]{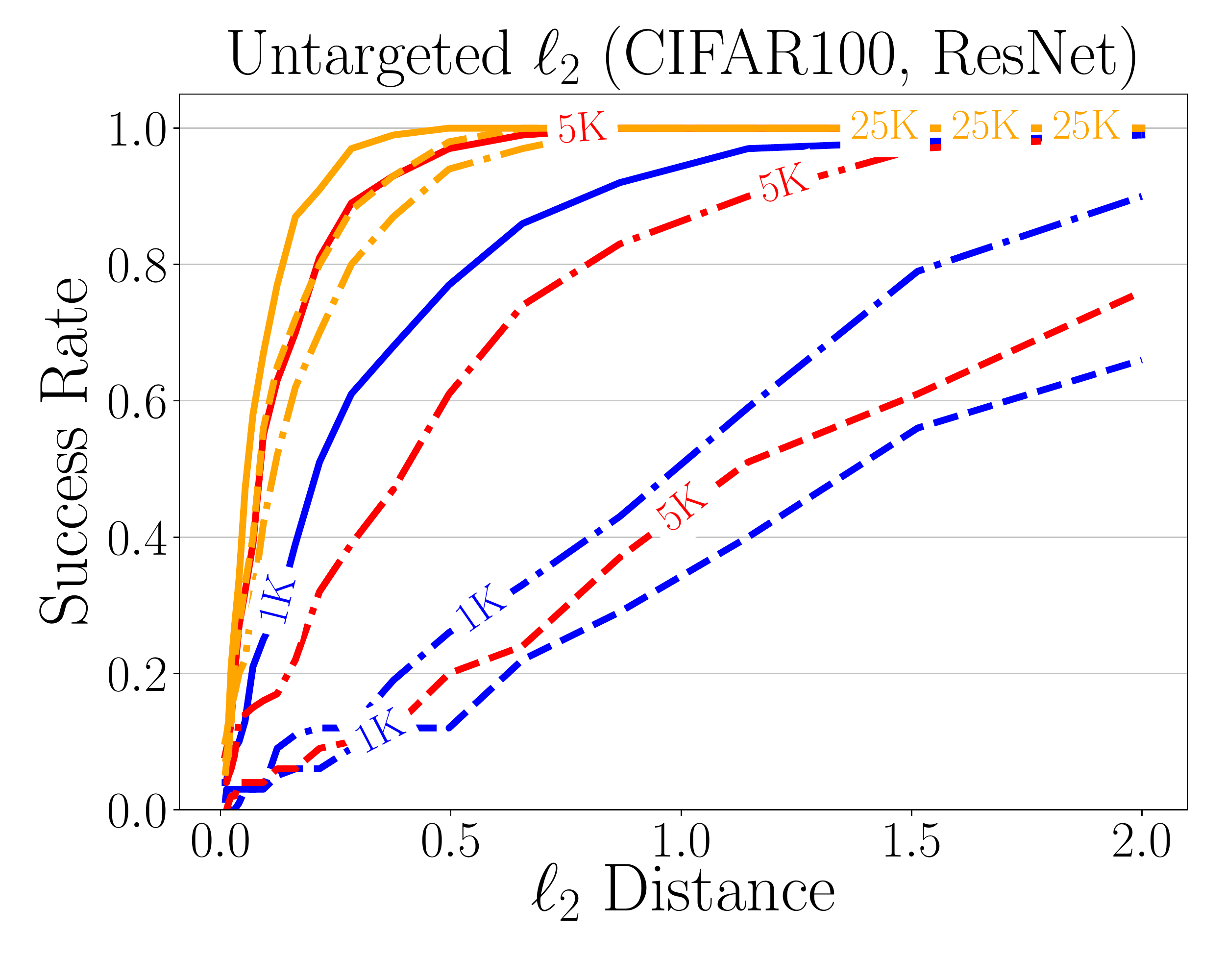} 
\includegraphics[width=0.23\linewidth]{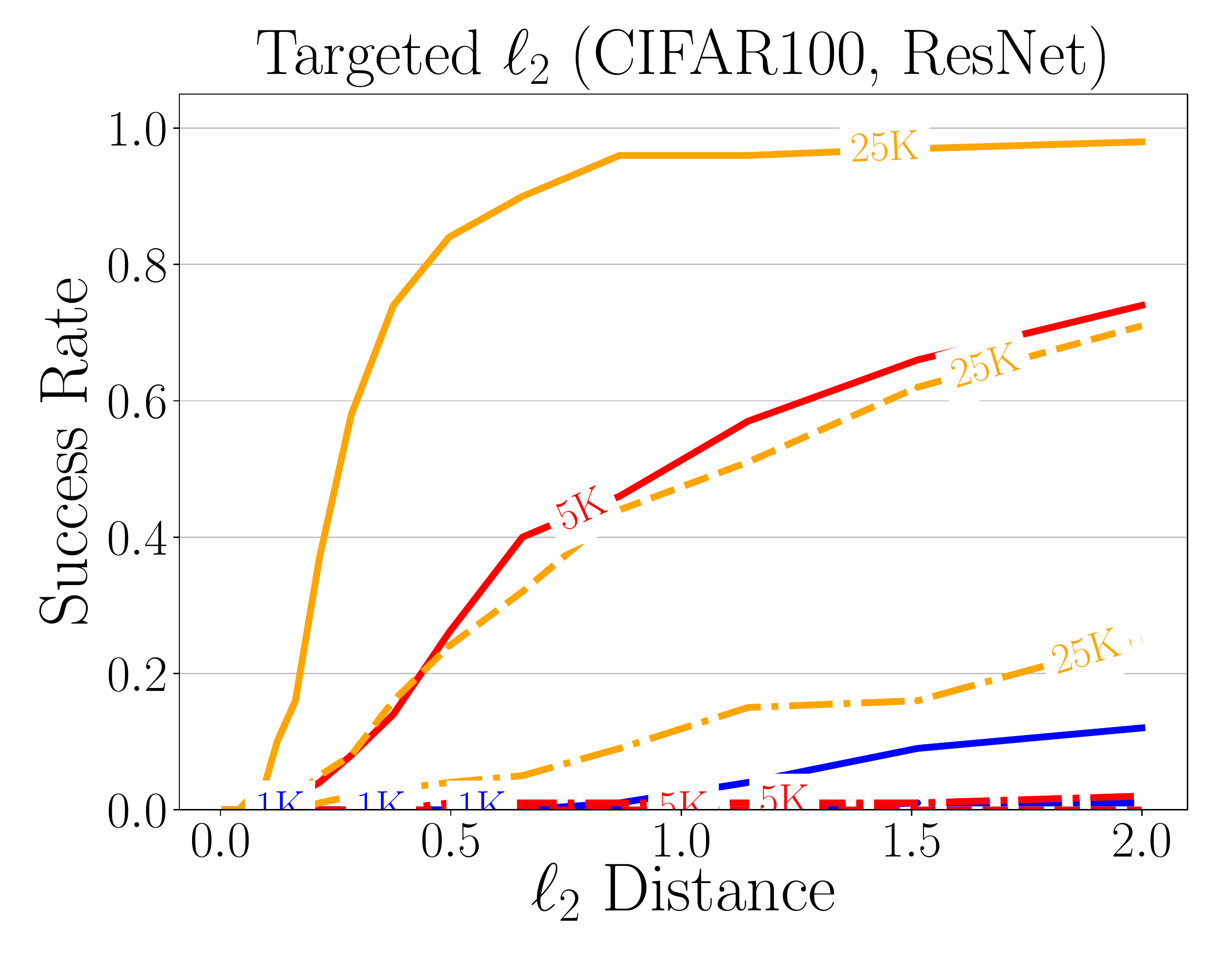} 
\includegraphics[width=0.23\linewidth]{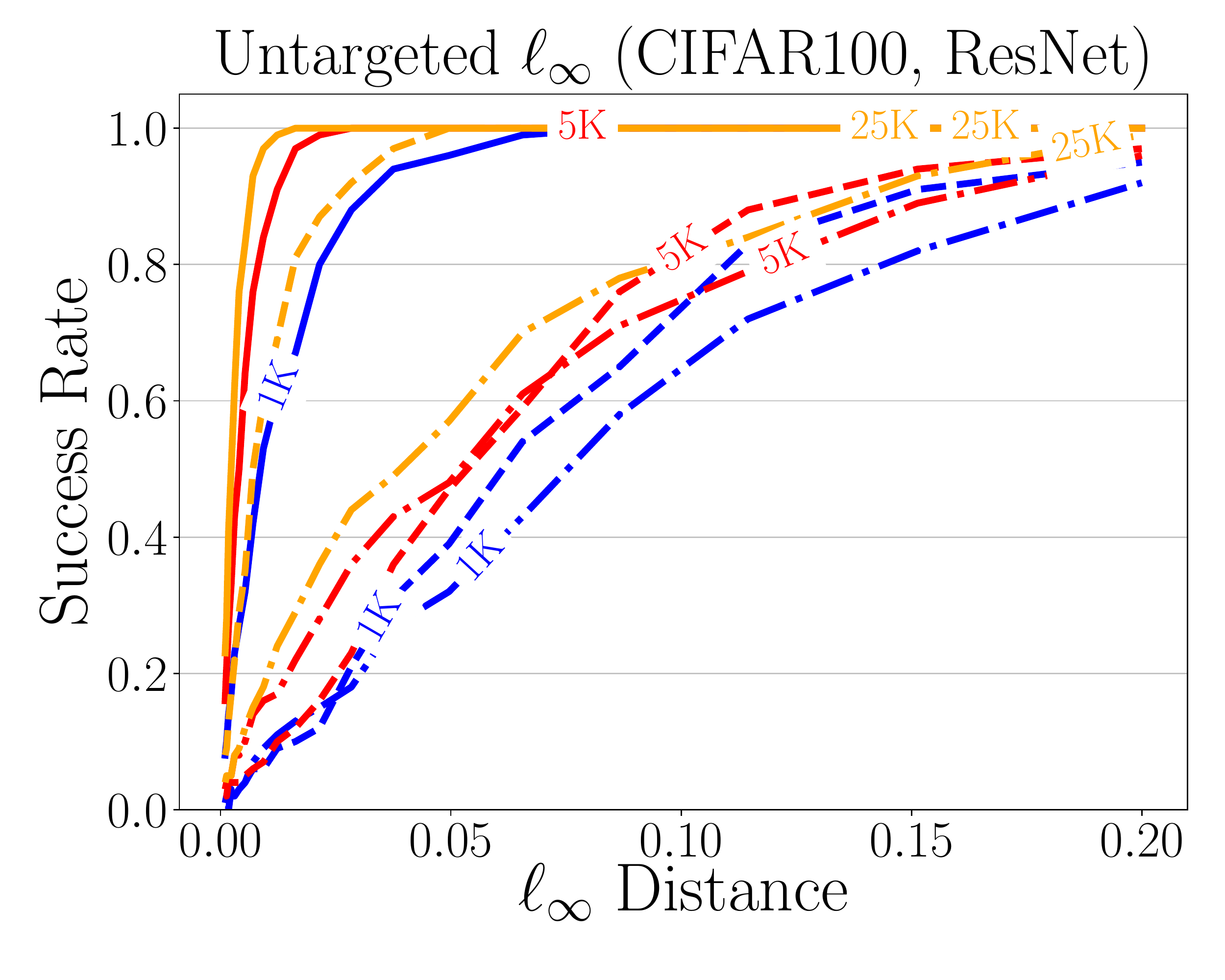} 
\includegraphics[width=0.23\linewidth]{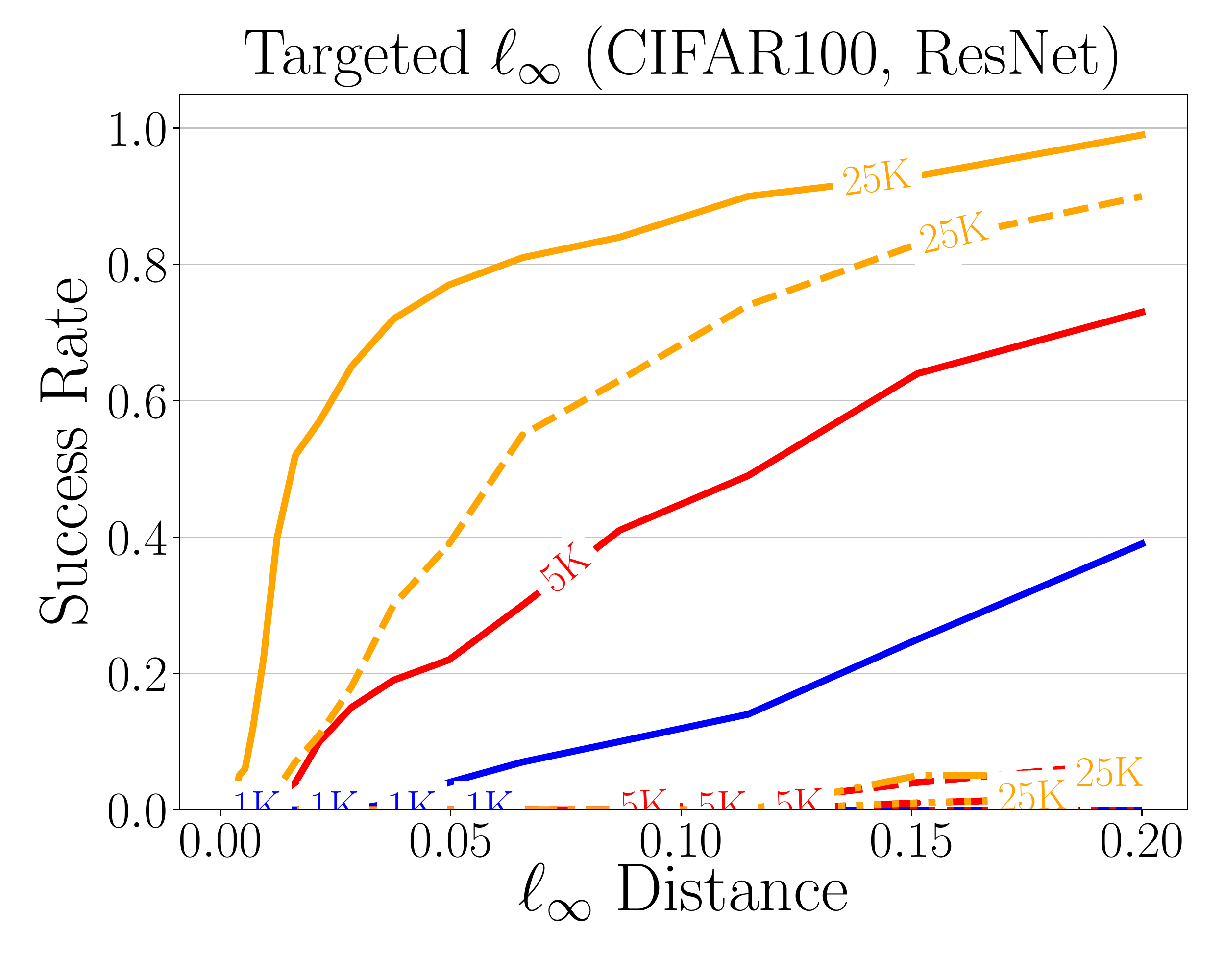} 

\includegraphics[width=0.23\linewidth]{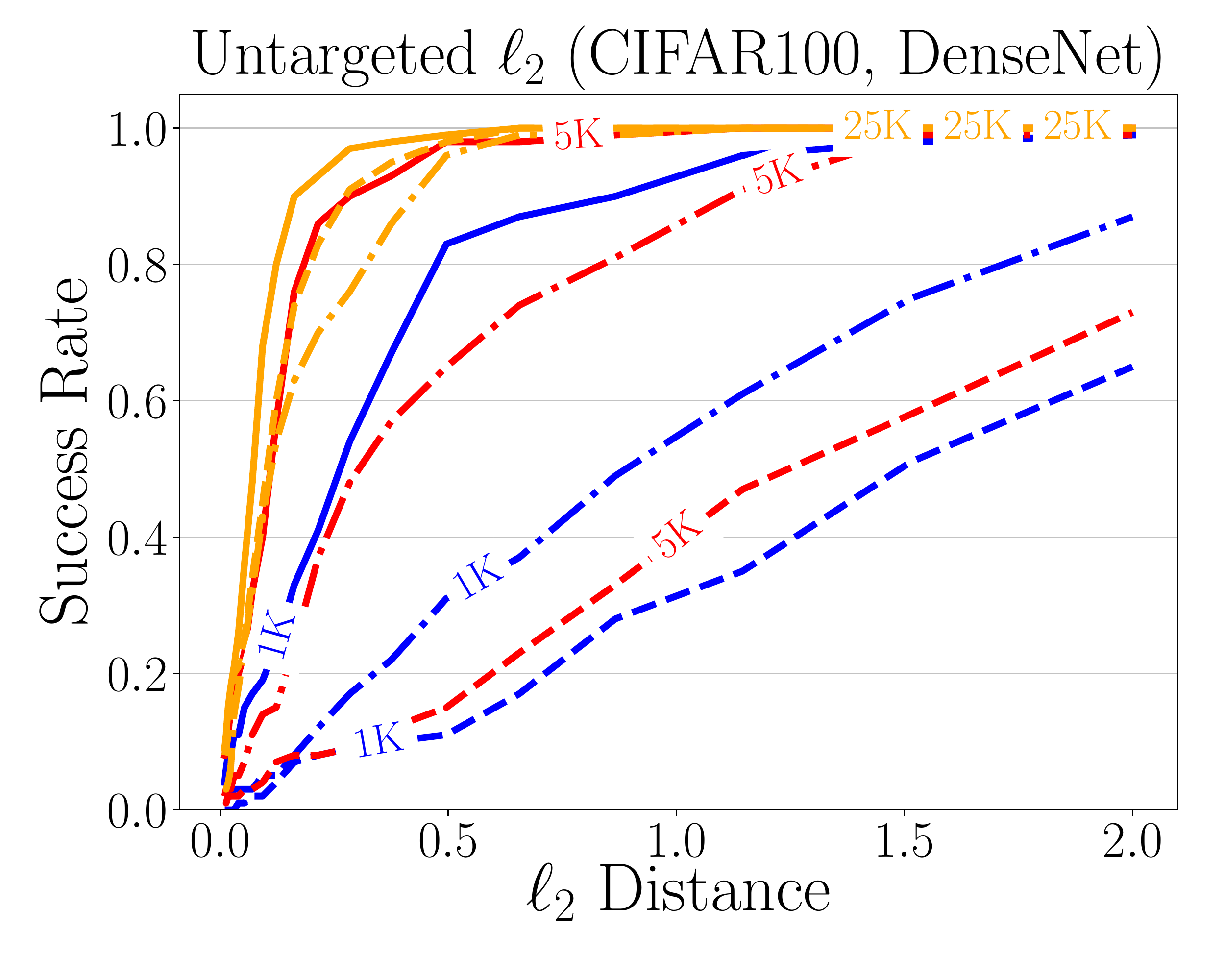} 
\includegraphics[width=0.23\linewidth]{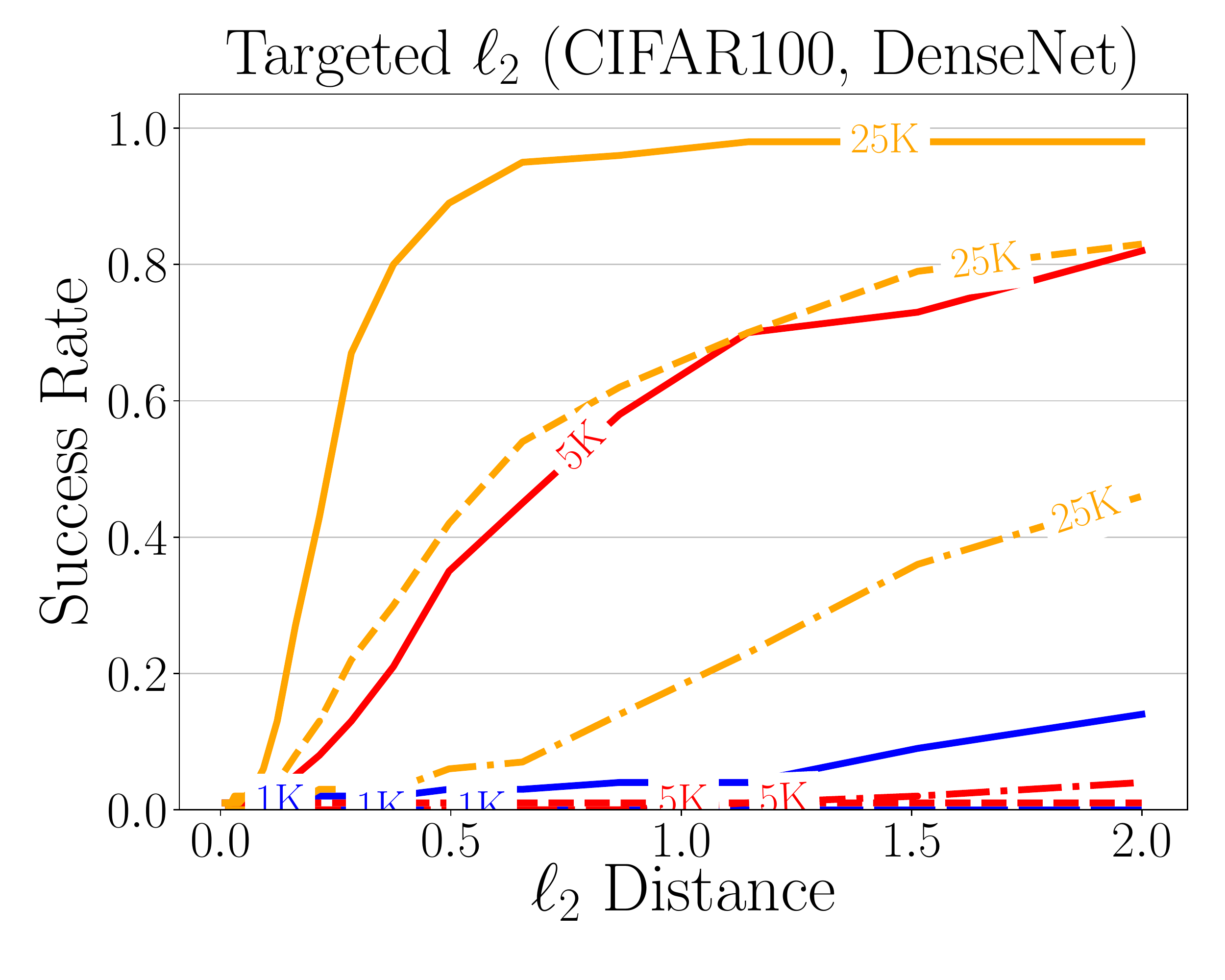}
\includegraphics[width=0.23\linewidth]{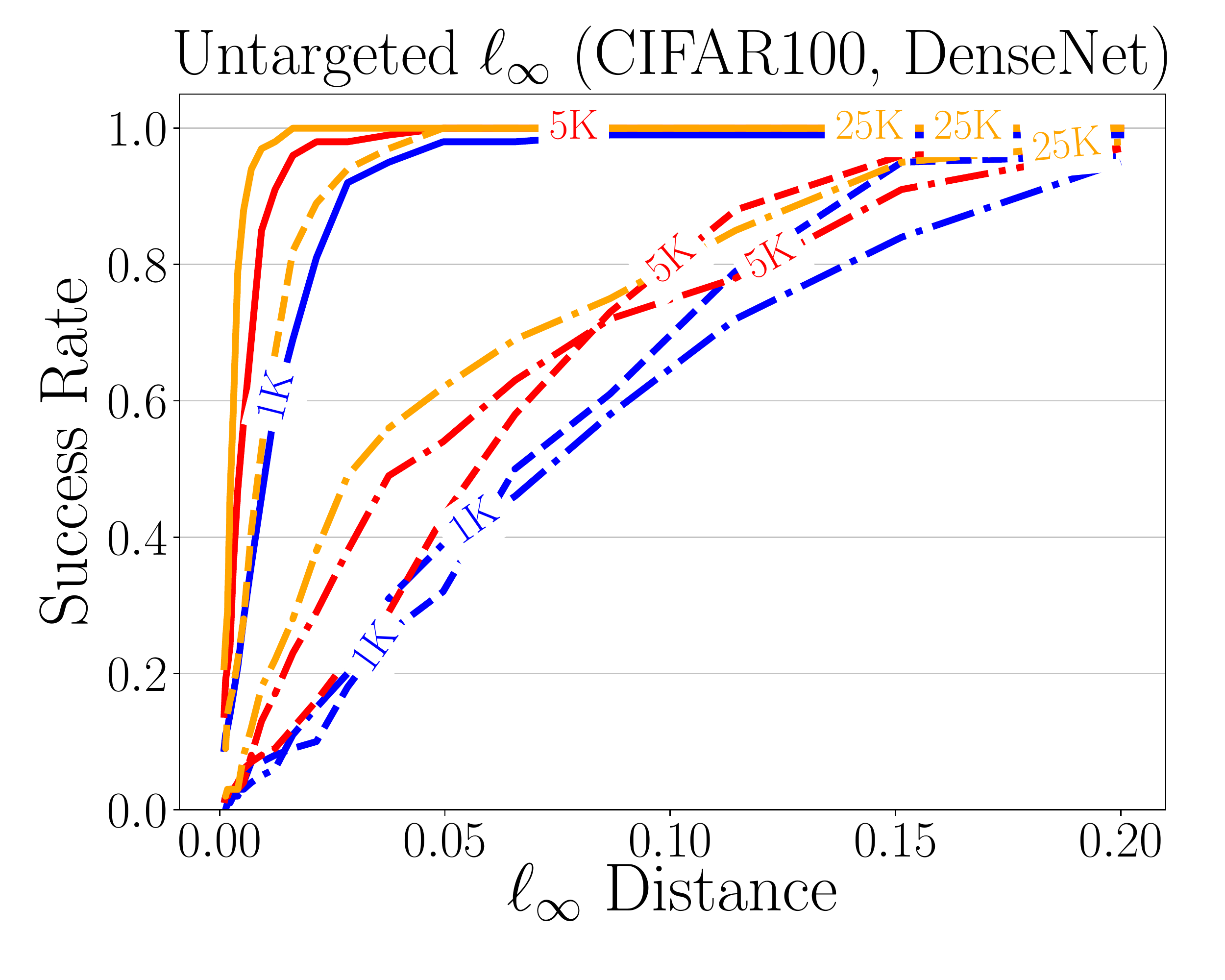} 
\includegraphics[width=0.23\linewidth]{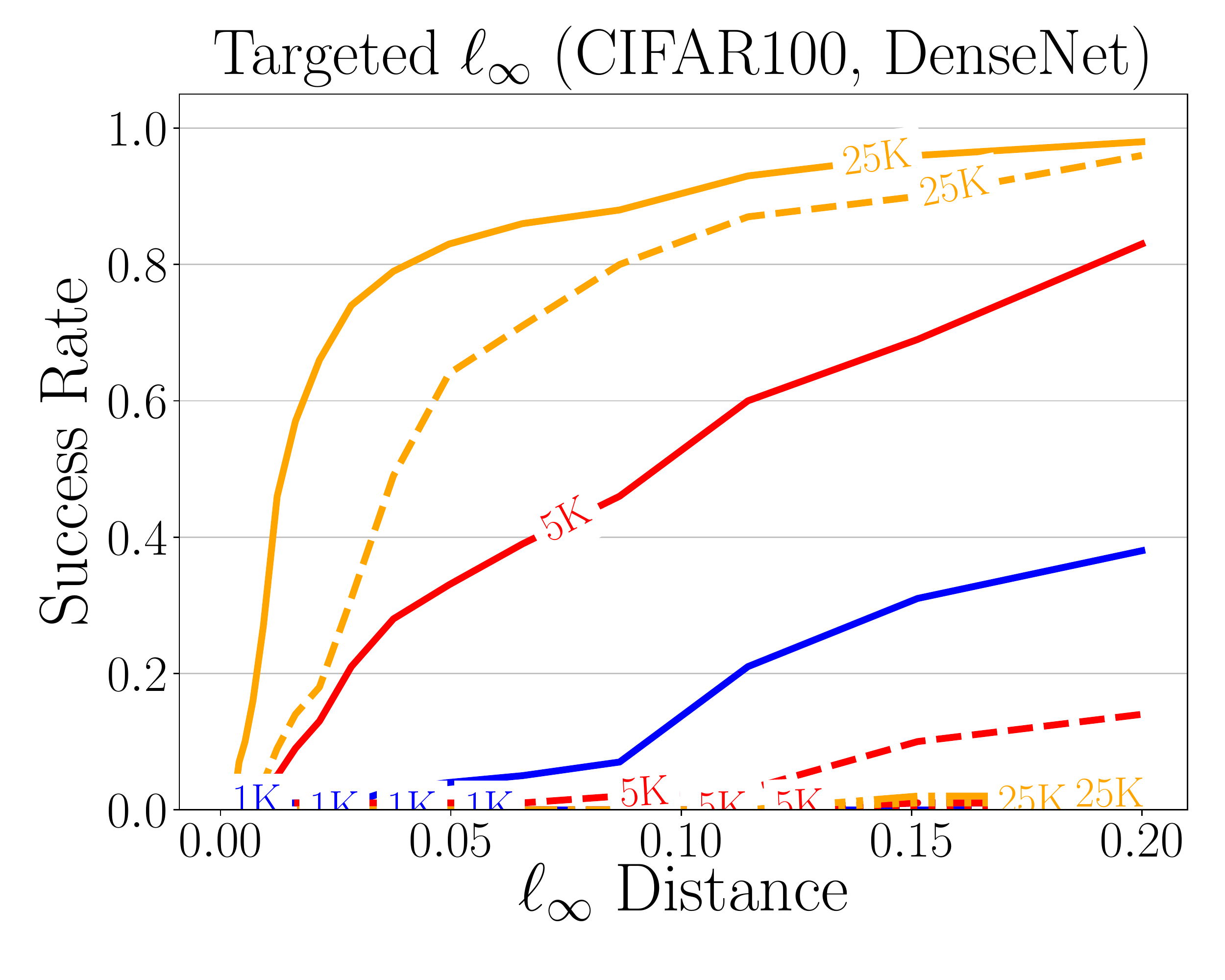} 

\includegraphics[width=0.23\linewidth]{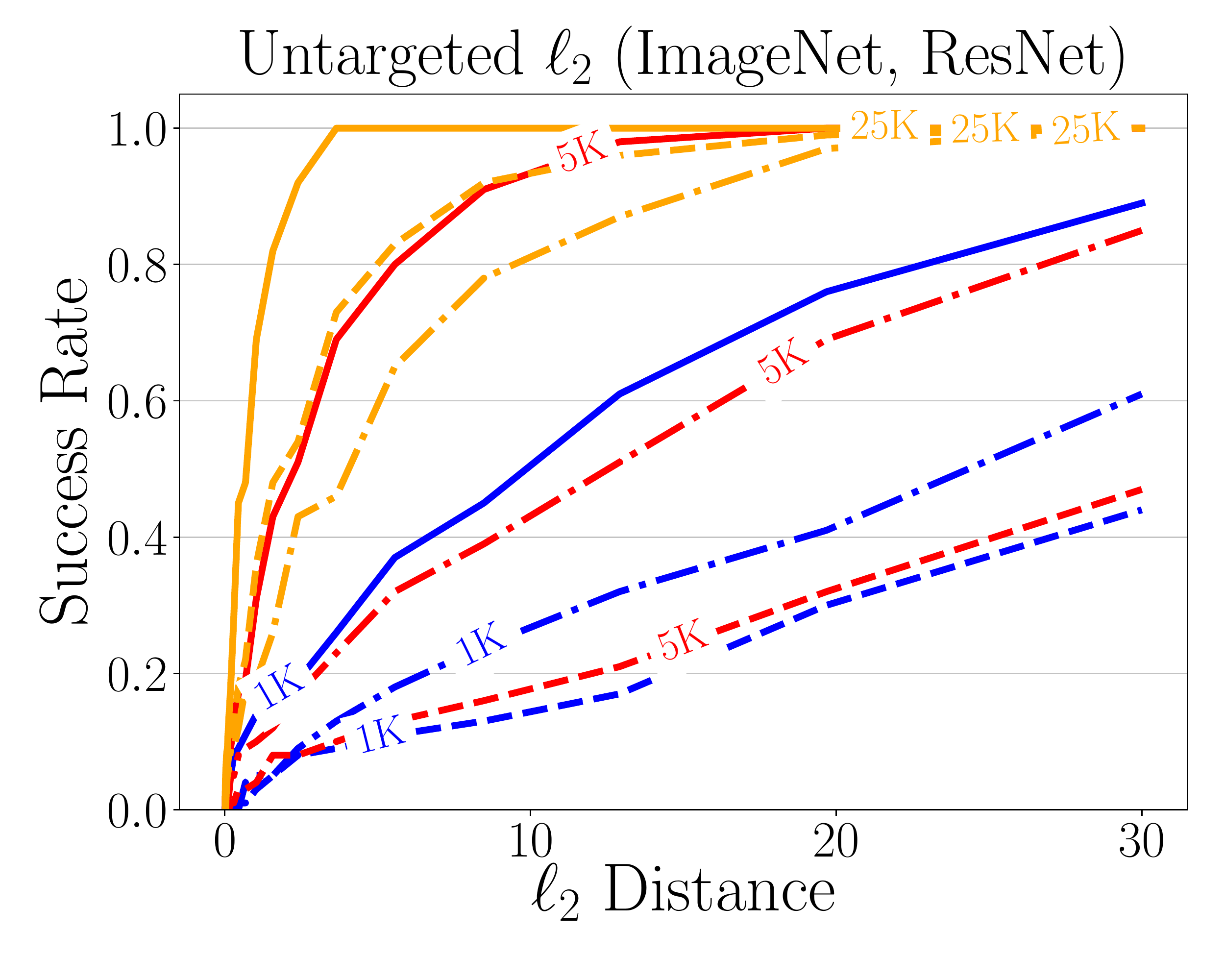} 
\includegraphics[width=0.23\linewidth]{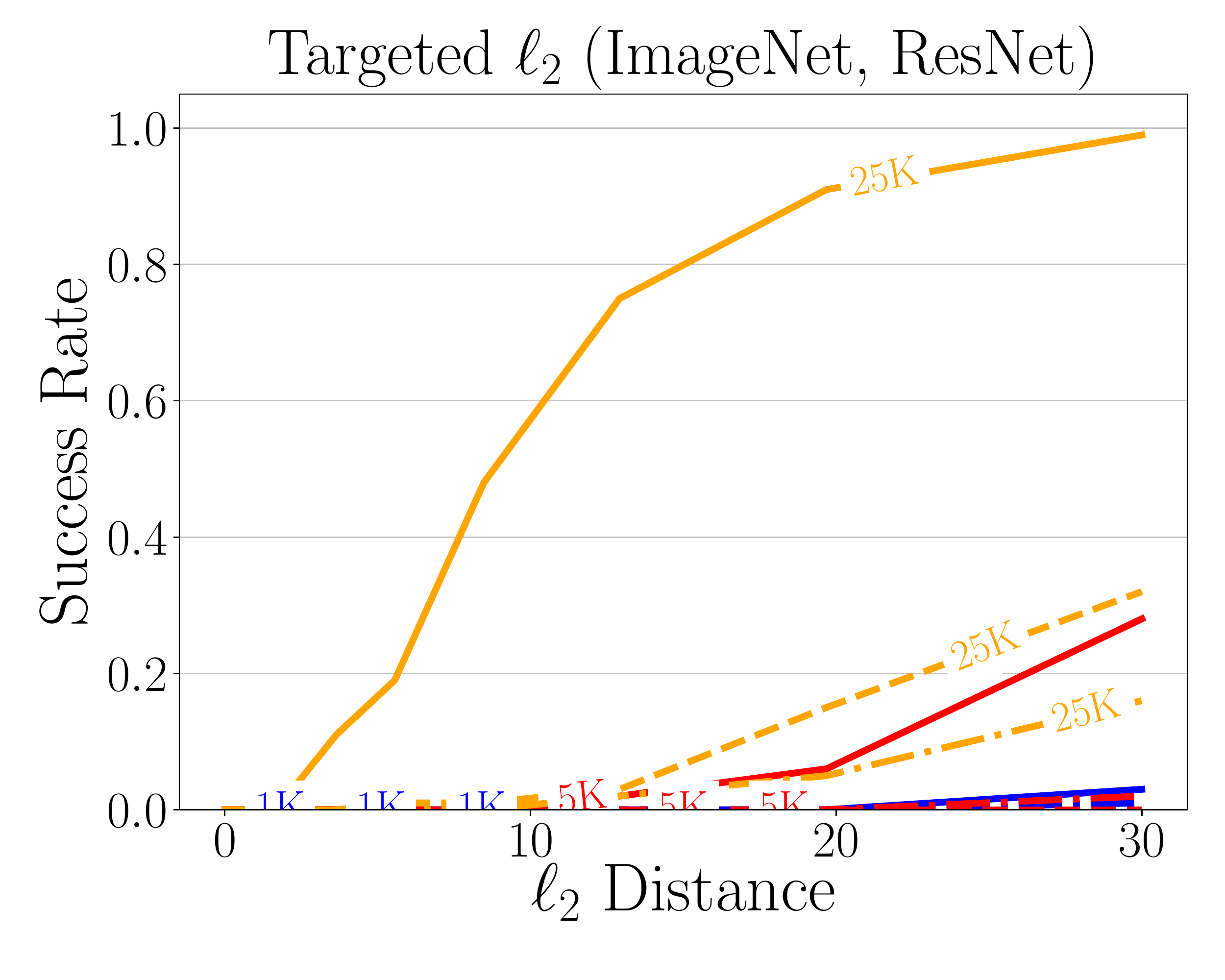}
\includegraphics[width=0.23\linewidth]{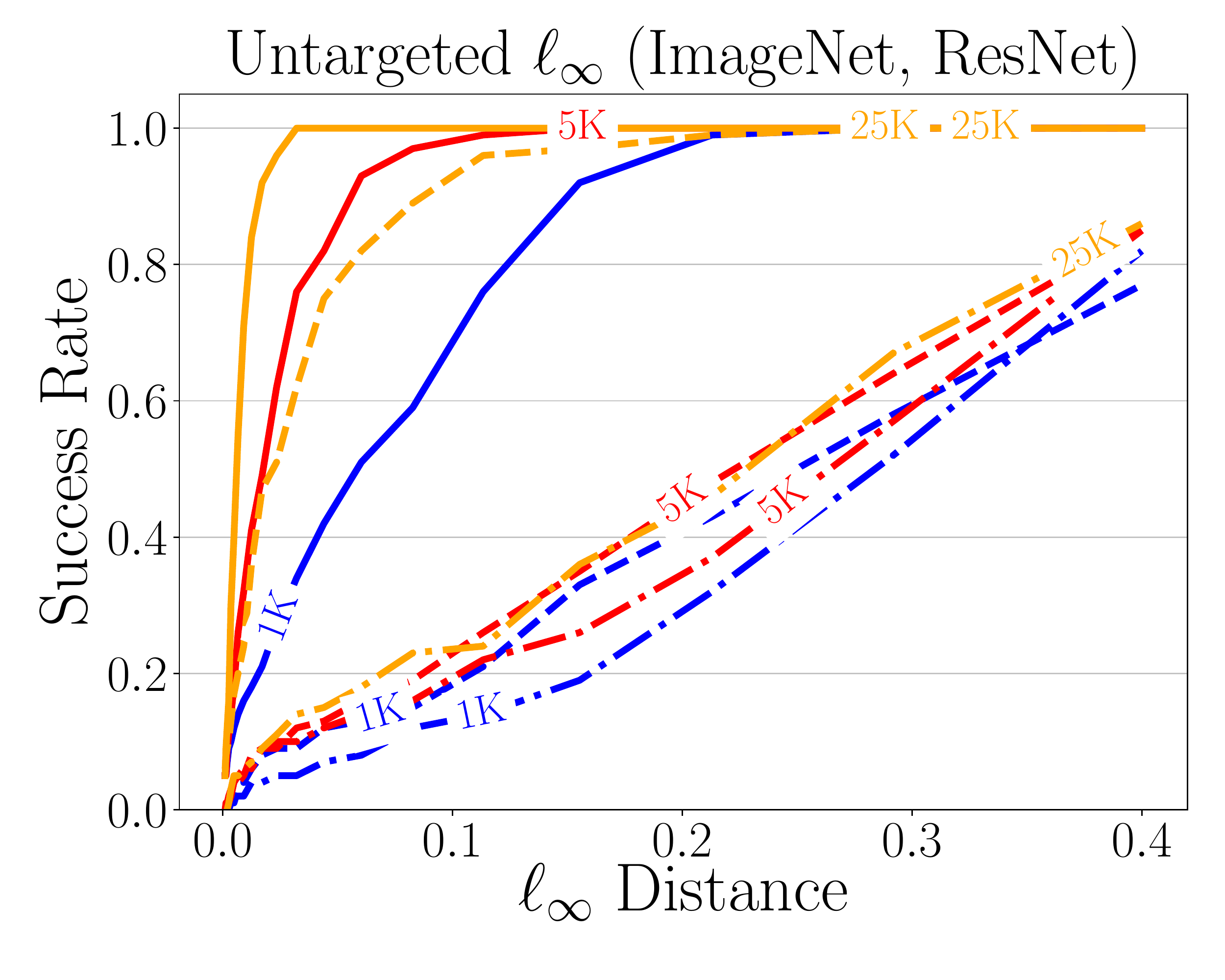} 
\includegraphics[width=0.23\linewidth]{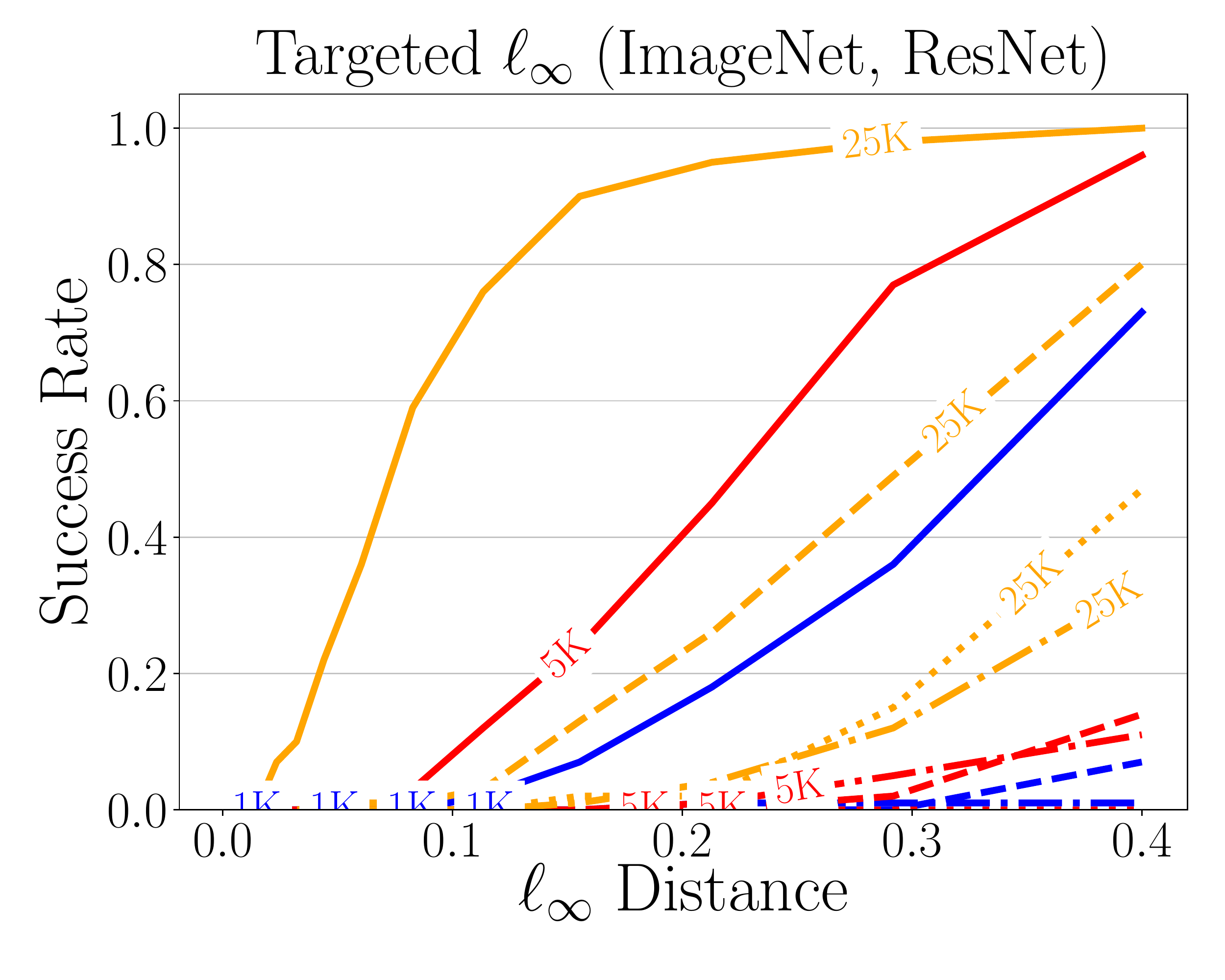} 

\includegraphics[width=0.9\linewidth]{figures_updated/legend-success_rate.png}
\caption{Success rate versus
distance threshold for CIFAR-100 with ResNet, DenseNet, and ImageNet with ResNet from top to bottom rows. 1st column:
untargeted $\ell_2$. 2nd column: targeted $\ell_2$. 3rd column:
untargeted $\ell_\infty$. 4th column: targeted $\ell_\infty$.}
\label{fig:success2}
\end{figure*}

\paragraph{Results} Figure~\ref{fig:queries} and \ref{fig:queries2} show the median distance (on a log scale) against the queries, with the first and third quartiles used as lower and upper error bars. For Boundary, Opt and HopSkipJumpAttack, Table~\ref{table:efficiency} summarizes the median distance when the number of queries is fixed at {1,000, 5,000, and 20,000} across all distance types, data, models and objectives. Figure~\ref{fig:success} and \ref{fig:success2} show the success rate against the distance threshold. Figure~\ref{fig:queries} and \ref{fig:success} contain results on MNIST with CNN, and CIFAR-10 with ResNet, Denset, subsequently from the top row to the bottom row. Figure~\ref{fig:queries2} and \ref{fig:success2} contain results on CIFAR-100 with ResNet and DenseNet, and ImageNet with ResNet, subsequently from the top row to the bottom row. The four columns are for untargeted $\ell_2$, targeted $\ell_2$, untargeted $\ell_\infty$ and targeted $\ell_\infty$ attacks respectively.

With a limited number of queries, HopSkipJumpAttack is able to
craft adversarial examples of a significantly smaller distance with
the corresponding original examples across all data sets, followed by Boundary Attack and Opt Attack. As a concrete example, Table~\ref{table:efficiency} shows that untargeted $\ell_2$-optimized
HopSkipJumpAttack achieves a median distance of $0.559$ on CIFAR-10
with a ResNet model at $1,000$ queries, which amounts to below
$3/255$ per pixel on average. At the same budget of queries, {Boundary Attack and Opt Attack only achieve median $\ell_2$-distances of $2.78$ and $2.07$ respectively}. The difference in efficiency becomes more significant for $\ell_\infty$ attacks. As shown in Figure~\ref{fig:success}, under an untargeted $\ell_\infty$-optimized HopSkipJumpAttack with 1,000 queries, all pixels are within an $8/255$-neighborhood of the original image for around 70\% of adversarial examples, a success rate achieved by Boundary Attack only after 20,000 queries. 

By comparing the odd and even columns of Figure~\ref{fig:queries}-\ref{fig:success2}, we can find that targeted HopSkipJumpAttack takes more queries than the untargeted one to achieve a comparable distance. This phenomenon becomes more explicit on CIFAR-100 and ImageNet, which have more classes. With the same number of queries, there is an order-of-magnitude difference in median distance between untargeted and targeted attacks (Figure~\ref{fig:queries}~and~\ref{fig:queries2}). For $\ell_2$-optimized HopSkipJumpAttack, while the untargeted version is able to craft adversarial images by perturbing $4$ bits per pixel on average {within 1,000 queries for $70\%-90\%$ of images in CIFAR-10 and CIFAR-100}, the targeted counterpart takes 2,000-5,000 queries. The other attacks fail to achieve a comparable performance even with 25,000 queries. On ImageNet, untargeted $\ell_2$-optimized HopSkipJumpAttack is able to fool the model with a perturbation of size $6$ bits per pixel on average for close to $50\%$ of images with $1,000$ queries; untargeted $\ell_\infty$-optimized HopSkipJumpAttack controls the maximum perturbation across all pixels within $16$ bits for $50\%$ images within $1,000$ queries. The targeted Boundary Attack is not able to control the perturbation size to such a small scale until after around $25,000$ queries. On the one hand, the larger query budget requirement results from a strictly more powerful formulation of targeted attack than untargeted attack. On the other hand, this is also because we initialize targeted HopSkipJumpAttack from an arbitrary image in the target class. The algorithm may be trapped in a bad local minimum with such an initialization. Future work can address systematic approaches to better initialization.

As a comparison between data sets and models, we see that adversarial images often have a larger distance to their corresponding original images on MNIST than on CIFAR-10 and CIFAR-100, which has also been observed in previous work (e.g., \cite{carlini2017towards}). This might be because it is more difficult to fool a model on simpler tasks. On the other hand, HopSkipJumpAttack also converges in a fewer number of queries on MNIST, as is shown in Figure~\ref{fig:queries}. It does not converge even after $25,000$ queries on ImageNet. We conjecture the query budget is related to the input dimension, and the smoothness of decision boundary.
We also observe the difference in model structure does not have a large influence on decision-based algorithms, if the training algorithm and the data set keep the same.
For ResNet and DenseNet trained on a common data set, a decision-based algorithm achieves comparable performance in crafting adversarial examples, although DenseNet has a more complex structure than ResNet. 

As a comparison with state-of-the-art white-box targeted attacks, C\&W attack~\cite{carlini2017towards} achieves an average $\ell_2$-distance of $0.33$ on CIFAR-10, and BIM~\cite{kurakin2016adversarial} achieves an average $\ell_\infty$-distance of $0.014$ on CIFAR-10. Targeted HopSkipJumpAttack achieves a comparable distance with 5K-10K model queries on CIFAR-10, without access to model details. On ImageNet, targeted C\&W attack and BIM achieve an $\ell_2$-distance of $0.96$ and an $\ell_\infty$-distance of $0.01$ respectively. Untargeted HopSkipJumpAttack achieves a comparable performance with $10,000-15,000$ queries. The targeted version is not able to perform comparably as targeted white-box attacks when the budget of queries is limited within $25,000$.

\begin{figure*}[!t]
\centering
\includegraphics[width=0.75\linewidth]{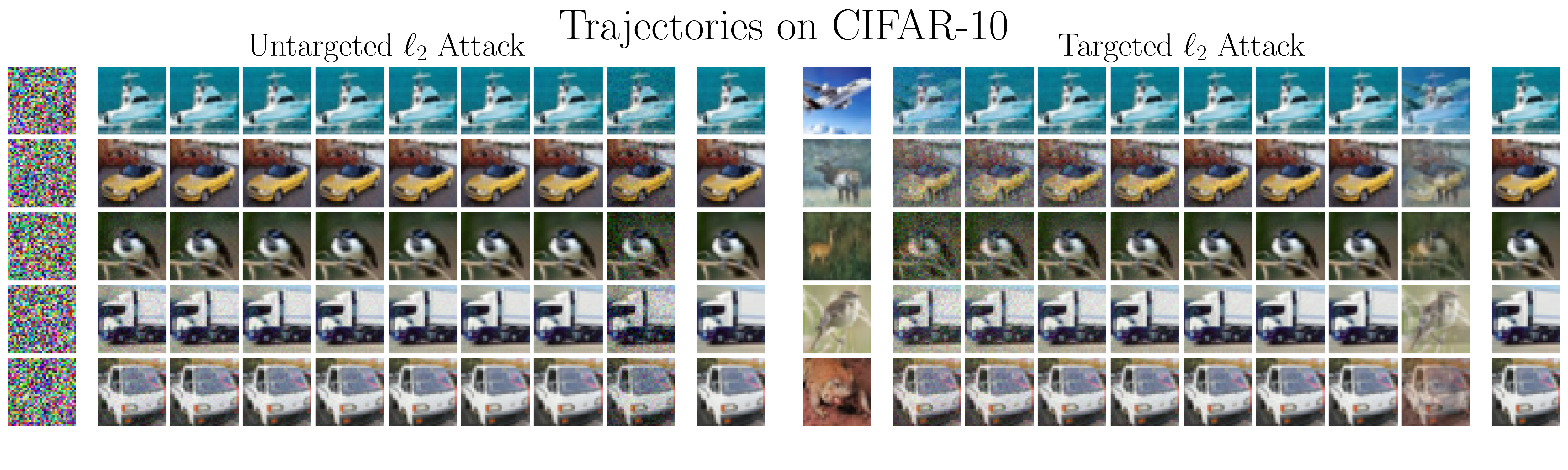}
\includegraphics[width=0.75\linewidth]{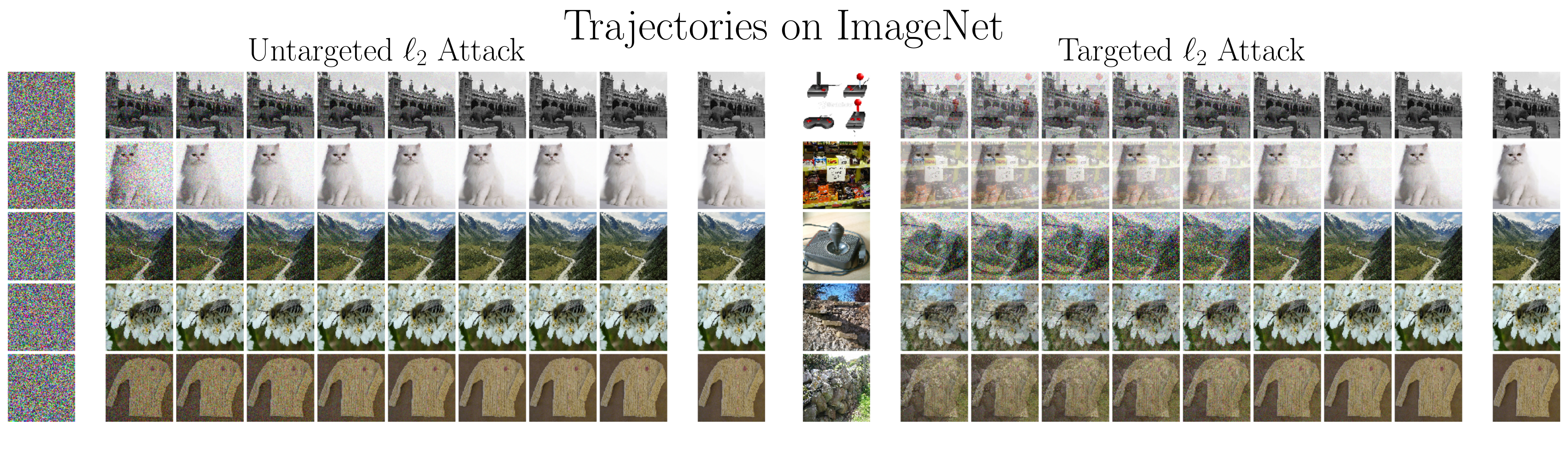} 
\caption{
Visualized trajectories of HopSkipJumpAttack {for optimizing $\ell_2$ distance} on randomly selected images in CIFAR-10 and ImageNet. 
1st column: initialization (after blended with original
images). 2nd-9th columns: images at 100, 200, 500, 1K, 2K, 5K, 10K,
25K model queries. 10th column: original images.}
\label{fig:traj1}
\end{figure*}

Visualized trajectories of HopSkipJumpAttack optimized for $\ell_2$ distances along varied queries on CIFAR10 and ImageNet can be found in
Figure~\ref{fig:traj1}. 
On CIFAR-10, we observe untargeted adversarial examples can be crafted within around $500$ queries; targeted HopSkipJumpAttack is capable of crafting human indistinguishable targeted adversarial examples within around $1,000-2,000$ queries. 
On ImageNet, untargeted HopSkipJumpAttack is able to craft good adversarial examples with $1,000$ queries, while targeted HopSkipJumpAttack takes $10,000-20,000$ queries. 

\subsection{Defense mechanisms under decision-based attacks}\label{sec:defense}
{We investigate the robustness of various defense mechanisms under decision-based attacks.}
\paragraph{Defense mechanisms} {Three defense mechanisms are evaluated: defensive distillation, region-based classification, and adversarial training. Defensive distillation~\cite{papernot2016distillation}, a form of gradient masking~\cite{papernot2017practical}, trains a second model to predict the output probabilities of an existing model of the same structure. We use the implementaion provided by~\citet{carlini2017towards} for defensive distillation. The second defense, region-based classification, belongs to a wide family of mechanisms which add test-time randomness to the inputs or the model, causing the gradients to be randomized~\cite{athalye2018obfuscated}. Multiple variants have been proposed to randomize the gradients~\cite{cao2017mitigating, liu2018towards, guneet2018stochastic, cohen2019certified, xie2018mitigating}. We adopt the implementation in \citet{cao2017mitigating} with suggested noise levels. Given a trained base model, region-based classification samples points from the hypercube centered at the input image, predicts the label for each sampled point with the base model, and then takes a majority vote to output the label. Adversarial training~\cite{goodfellow2014explaining, kurakin2016adversarial, madry2018towards, tramer2018ensemble} is known to be one of the most effective defense mechanisms against adversarial perturbation~\cite{carlini2017adversarial,
  athalye2018obfuscated}. 
We evaluate a publicly available model trained through a robust optimization method proposed by~\citet{madry2018towards}. We further evaluate our attack method by constructing a non-differentiable model via input binarization followed by a random forest in Appendix~\ref{app:nondiff}. The evaluation is carried out on MNIST, where defense mechanisms such as adversarial training work most effectively.}

\paragraph{Baselines} We compare our algorithm with state-of-the-art attack algorithms that require access to gradients, including C\&W Attack~\cite{carlini2017towards}, DeepFool~\cite{moosavi2016deepfool} for minimizing $\ell_2$-distance, and FGSM~\cite{goodfellow2014explaining}, and BIM~\cite{kurakin2018adversarial,madry2018towards} for minimizing $\ell_\infty$-distance. {For region-based classification, the gradient of the base classifier is taken with respect to the original input.}

{
We further include methods designed specifically for the defense mechanisms under threat. For defensive distillation, we include the $\ell_\infty$-optimized C\&W Attack~\cite{carlini2017towards}. For region-based classification, we include backward pass differentiable approximation (BPDA)~\cite{athalye2018obfuscated}, which calculates the gradient of the model at a randomized input to replace the gradient at the original input in C\&W Attack and BIM. All of these methods assume access to model details or even defense mechanisms, which is a stronger threat model than the one required for decision-based attacks. We also include Boundary Attack as a decision-based baseline.
}

{
  For HopSkipJumpAttack and Boundary Attack, we include the success rate at three different scales of query budget: 2K, 10K and 50K, so as to evaluate our method both with limited queries and a sufficient number of queries. We find the convergence of HopSkipJumpAttack becomes unstable on region-based classification, resulting from the difficulty of locating the boundary in the binary search step when uncertainty is increased near the boundary. Thus, we increase the binary search threshold to 0.01 to resolve this issue. 
}

\begin{figure*}[!bt]
\centering
\includegraphics[width=0.30\linewidth]{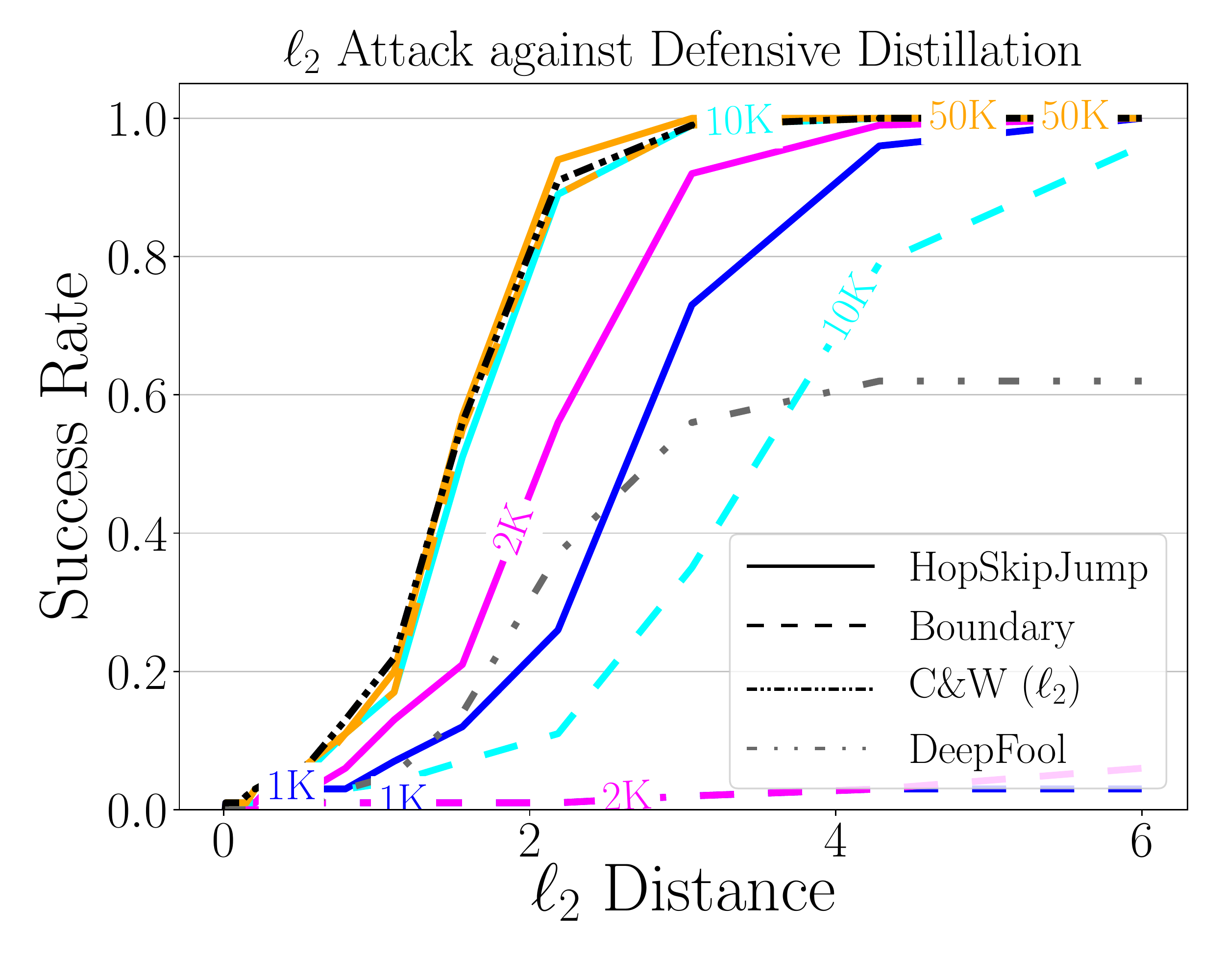}
\includegraphics[width=0.30\linewidth]{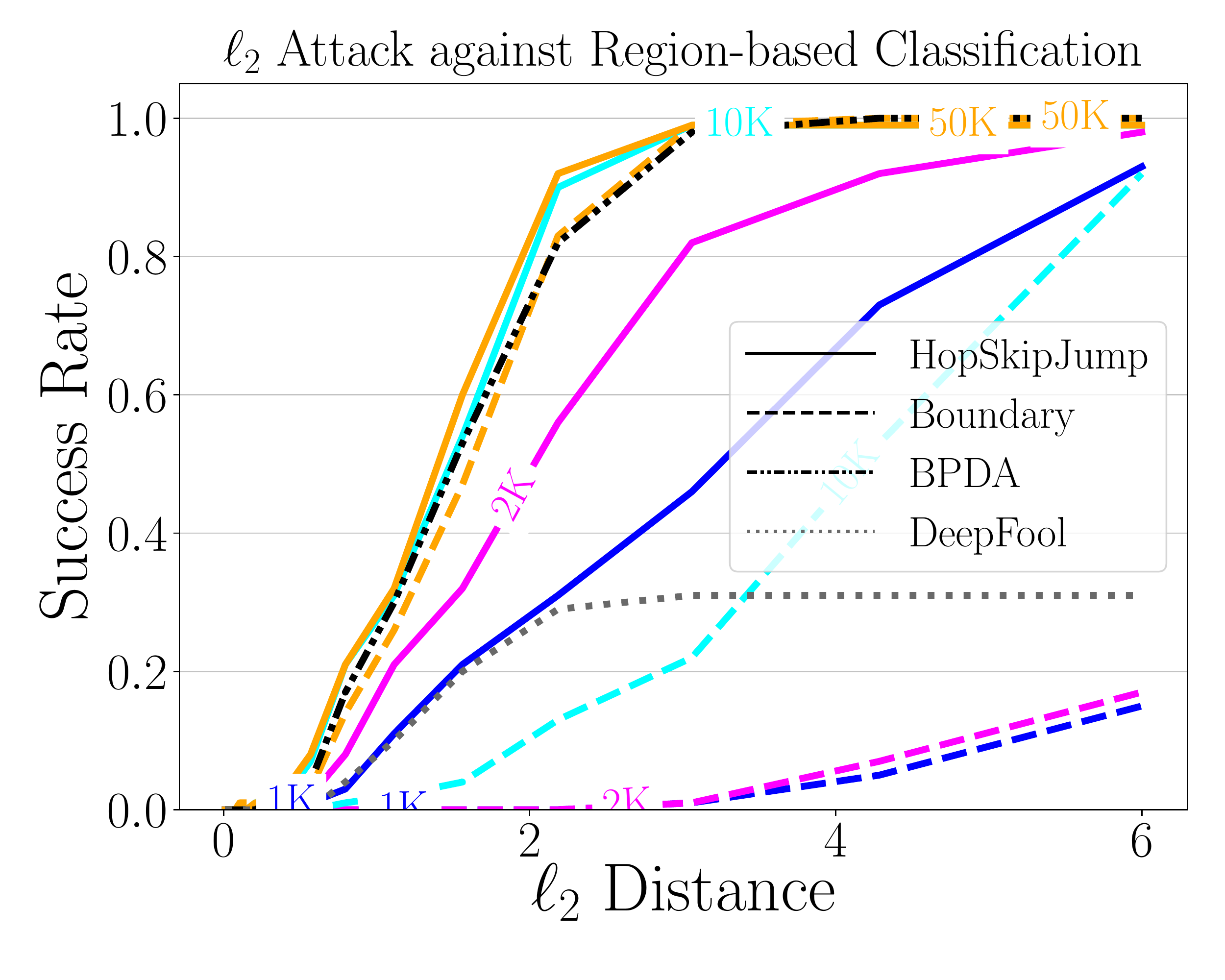}
\includegraphics[width=0.30\linewidth]{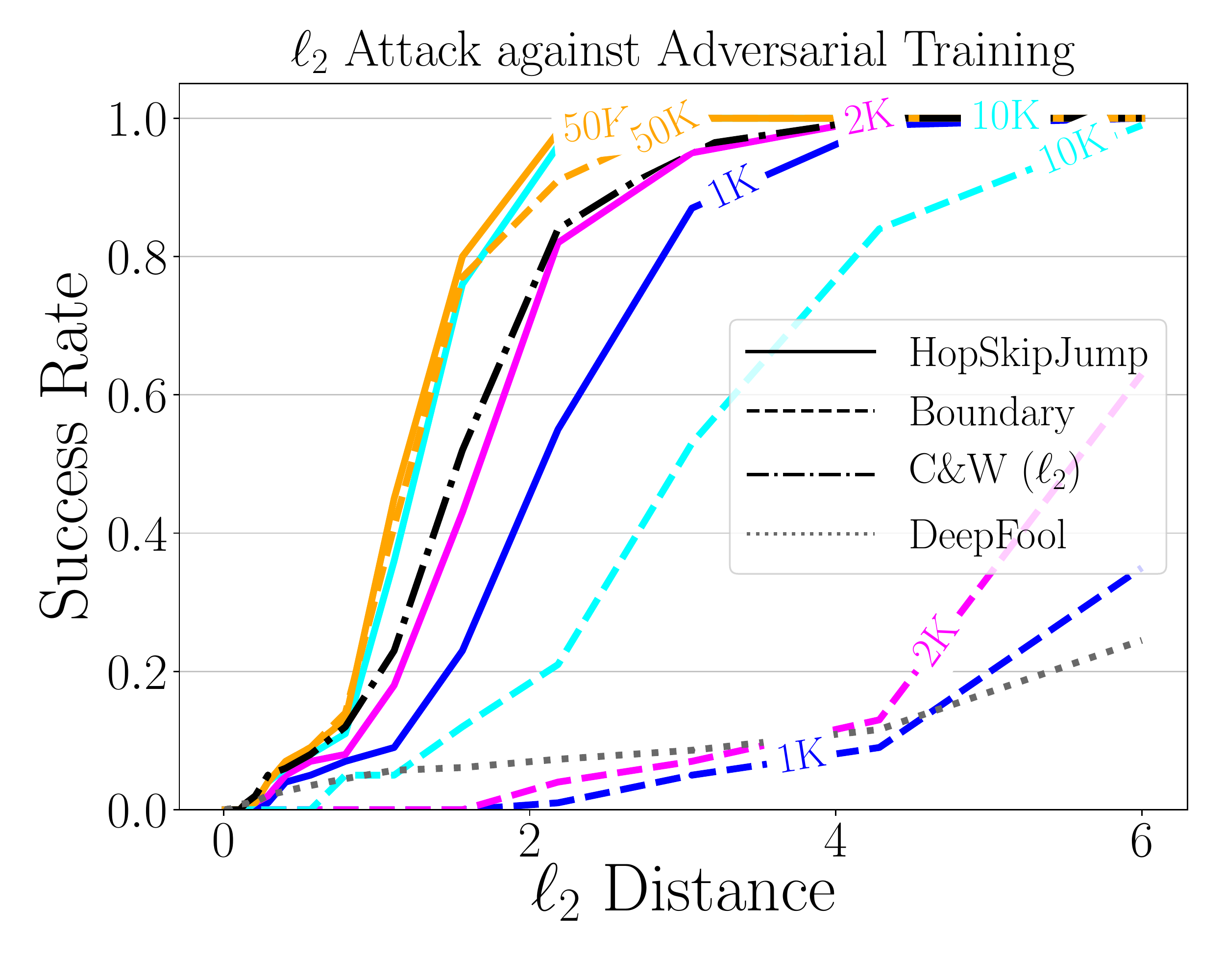}

\includegraphics[width=0.30\linewidth]{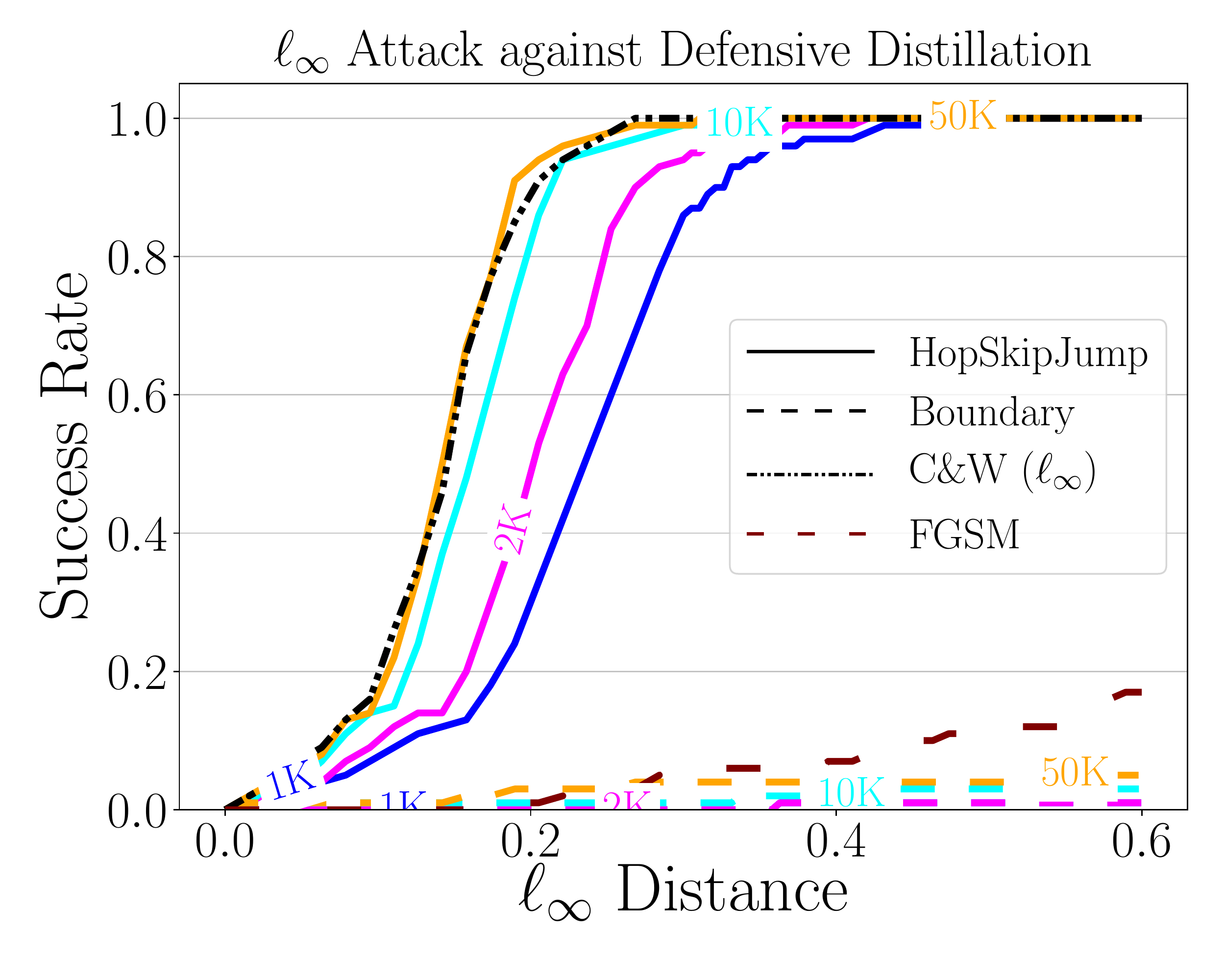}
\includegraphics[width=0.30\linewidth]{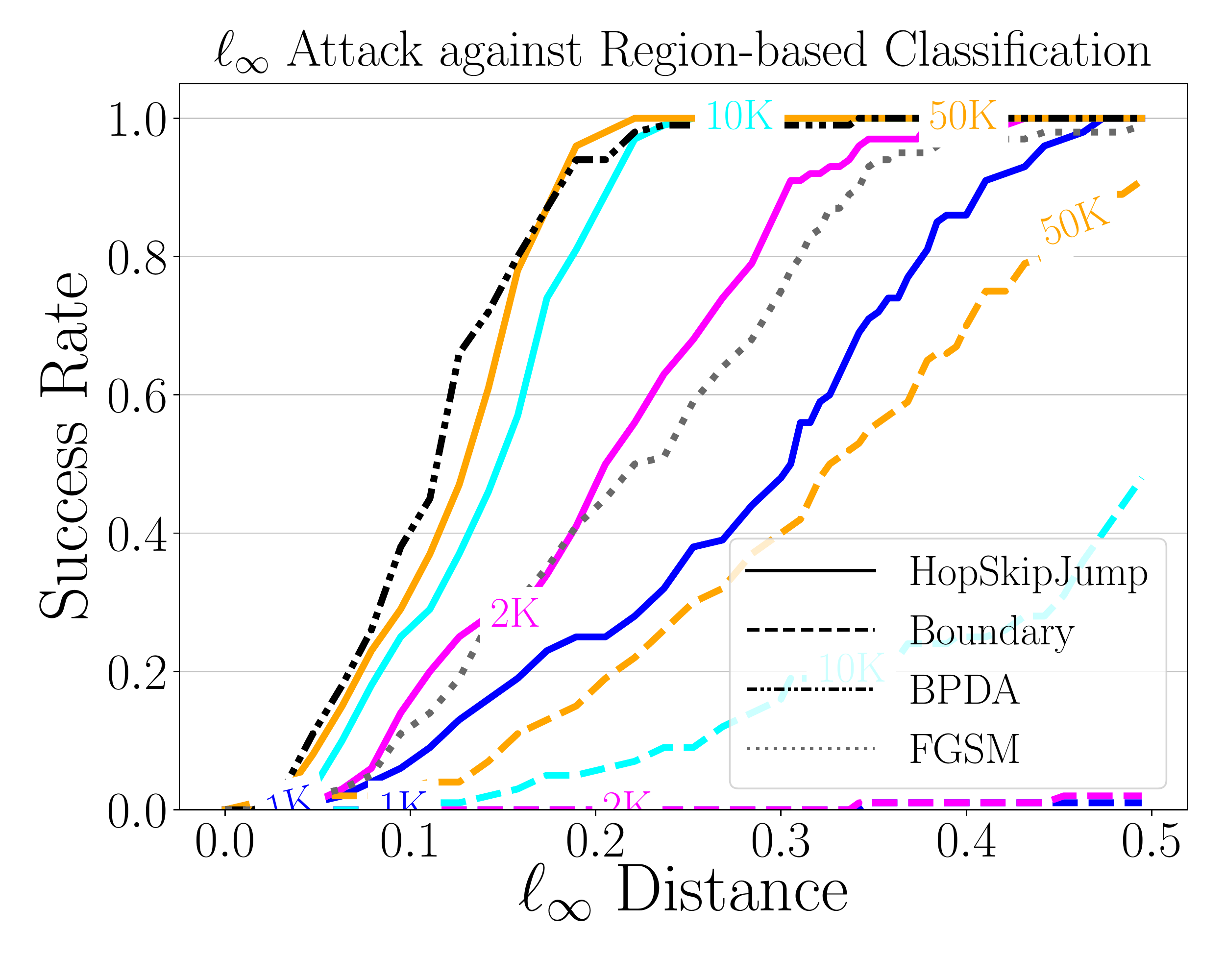}
\includegraphics[width=0.30\linewidth]{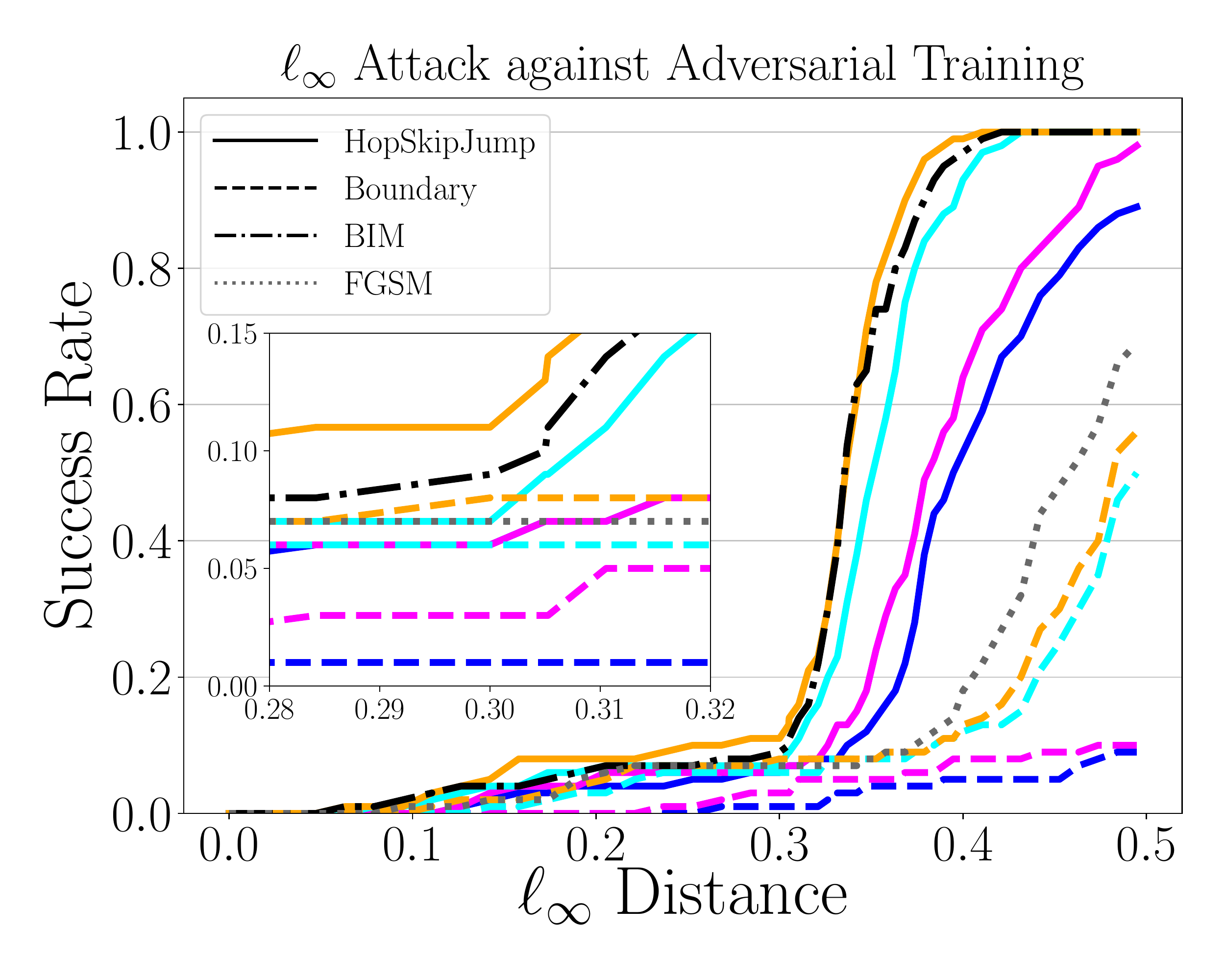}
\caption{{Success rate versus distance threshold for a distilled model, a region-based classifier and an adversarially trained model on MNIST. Blue, magenta, cyan and orange lines are used for HopSkipJumpAttack and Boundary Attack at the budget of 1K, 2K, 10K and 50K respectively. Different attacks are plotted with different line styles. An amplified figure is included near the critical $\ell_\infty$-distance of $0.3$ for adversarial training.}}
\label{fig:success_vs_dist_defense}
\end{figure*} 
\paragraph{Results} {Figure~\ref{fig:success_vs_dist_defense} shows the success rate of various attacks at different distance thresholds for the three defense mechanisms. On all of the three defenses, HopSkipJumpAttack demonstrates similar or superior performance compared to state-of-the-art white-box attacks with sufficient model queries. Even with only 1K-2K model queries, it also achieves acceptable performance, although worse than the best white-box attacks. With sufficient queries, Boundary Attack achieves a comparable performance under the $\ell_2$-distance metric. But it is not able to generate any adversarial examples when the number of queries is limited to $1,000$. We think this is because the strength of our batch gradient direction estimate over the random walk step in Boundary Attack becomes more explicit when there is uncertainty or non-smoothness near the decision boundary. We also observe that Boundary Attack does not work in optimizing the $\ell_\infty$-distance metric for adversarial examples, making it difficult to evaluate defenses designed for $\ell_\infty$ distance, such as adversarial training proposed by \citet{madry2018towards}.}

{On a distilled model, when the $\ell_\infty$-distance is thresholded at $0.3$, a perturbation size proposed by~\citet{madry2018towards} to measure adversarial robustness, HopSkipJumpAttack achieves success
rates of $86\%$ and $99\%$ with 1K and 50K queries respectively. At an $\ell_2$-distance of 3.0, the success rate is $91\%$ with 2K queries. HopSkipJumpAttack achieves a comparable performance with C\&W attack under both distance metrics with 10K-50K queries. Also, gradient masking~\cite{papernot2017practical} by defensive distillation does not have a large influence on the query efficiency of HopSkipJumpAttack, indicating that the gradient direction estimate is robust under the setting where the model does not have useful gradients for certain white-box attacks. }

{On region-based classification, with 2K queries, HopSkipJumpAttack achieves success rates of $82\%$ and $93\%$ at the same $\ell_\infty$- and $\ell_2$-distance thresholds respectively. With 10K-50K queries, it is able to achieve a comparable performance to BPDA, a white-box attack tailored to such defense mechanisms. On the other hand, we observe that HopSkipJumpAttack converges slightly slower on region-based classification than itself on ordinary models, which is because stochasticity near the boundary may prevent binary search in HopSkipJumpAttack from locating the boundary accurately. } 

{On an adversarially trained model, HopSkipJumpAttack achieves a success
rate of {$11.0\%$} with 50K queries when the $\ell_\infty$-distance is thresholded at $0.3$. As a comparison, BIM has a success rate of $7.4\%$ at the given distance threshold. 
The success rate of $\ell_\infty$-HopSkipJumpAttack transfers to 
an accuracy of {$87.58\%$} on adversarially perturbed data, close to the state-of-the-art performance achieved by white-box attacks.\footnote{See \url{https://github.com/MadryLab/mnist_challenge}.} {With 1K queries, HopSkipJumpAttack also achieves comparable performance to BIM and C\&W attack.} }

\section{Discussion}
\label{SecDiscussion}

We have proposed a family of query-efficient algorithms based on a
novel gradient-direction estimate, HopSkipJumpAttack, for
decision-based generation of adversarial examples, which is capable of
optimizing $\ell_2$ and $\ell_\infty$-distances for both targeted and
untargeted attacks. Convergence analysis has been carried out given
access to the gradient. We have also provided analysis for the error
of our Monte Carlo estimate of gradient direction, which comes from
three sources: bias at the boundary for a nonzero perturbation size,
bias of deviation from the boundary, and variance. Theoretical
analysis has provided insights for selecting the step size and the
perturbation size, which leads to a hyperparameter-free algorithm. We
have also carried out extensive experiments, showing HopSkipJumpAttack
compares favorably to Boundary Attack in query efficiency, and achieves competitive performance on several defense mechanisms.

Given the fact that HopSkipJumpAttack is able to craft a human-indistinguishable adversarial example within a realistic budget of queries, it becomes important for the
community to consider the real-world impact of decision-based threat
models. We have also demonstrated that HopSkipJumpAttack is able to achieve comparable 
or even superior performance to state-of-the-art white-box attacks on 
several defense mechanisms, under a much weaker threat model. In particular, masked gradients, stochastic gradients, and non-differentiability are not barriers to our algorithm. 
Because of its effectiveness, efficiency, and applicability to non-differentiable models, we suggest future research on adversarial defenses may evaluate the designed mechanism against HopSkipJumpAttack as a first step.

One limitation of all existing decision-based algorithms, including HopSkipJumpAttack, is that they require evaluation of the target model near the boundary. They may not work effectively by limiting the queries near the boundary, or by widening the decision boundary through insertion of an additional ``unknown'' class for inputs with low confidence. We have also observed that it still takes tens of thousands of model queries for HopSkipJumpAttack to craft imperceptible adversarial examples with a target class on ImageNet, which has a relatively large image size. Future work may seek the combination of HopSkipJumpAttack with transfer-based attack to resolve these issues.

\newpage
\section{Acknowledgement}
We would like to thank Nicolas Papernot and anonymous reviewers for providing their helpful feedback. 
\begin{small}
\bibliography{zero}

\begin{thebibliography}{42}
\providecommand{\natexlab}[1]{#1}
\providecommand{\url}[1]{\texttt{#1}}
\expandafter\ifx\csname urlstyle\endcsname\relax
  \providecommand{\doi}[1]{doi: #1}\else
  \providecommand{\doi}{doi: \begingroup \urlstyle{rm}\Url}\fi

\bibitem[Szegedy et~al.(2014)Szegedy, Zaremba, Sutskever, Bruna, Erhan,
  Goodfellow, and Fergus]{szegedy2013intriguing}
Christian Szegedy, Wojciech Zaremba, Ilya Sutskever, Joan Bruna, Dumitru Erhan,
  Ian Goodfellow, and Rob Fergus.
\newblock Intriguing properties of neural networks.
\newblock In \emph{International Conference on Learning Representations}, 2014.

\bibitem[Goodfellow et~al.(2015)Goodfellow, Shlens, and
  Szegedy]{goodfellow2014explaining}
Ian~J Goodfellow, Jonathon Shlens, and Christian Szegedy.
\newblock Explaining and harnessing adversarial examples.
\newblock In \emph{Proceedings of the International Conference on Learning
  Representations}, 2015.

\bibitem[Kurakin et~al.(2017)Kurakin, Goodfellow, and
  Bengio]{kurakin2016adversarial}
Alexey Kurakin, Ian Goodfellow, and Samy Bengio.
\newblock Adversarial machine learning at scale.
\newblock In \emph{International Conference on Learning Representations}, 2017.

\bibitem[Moosavi-Dezfooli et~al.(2016)Moosavi-Dezfooli, Fawzi, and
  Frossard]{moosavi2016deepfool}
Seyed-Mohsen Moosavi-Dezfooli, Alhussein Fawzi, and Pascal Frossard.
\newblock Deepfool: a simple and accurate method to fool deep neural networks.
\newblock In \emph{Proceedings of the IEEE Conference on Computer Vision and
  Pattern Recognition}, pages 2574--2582, 2016.

\bibitem[Papernot et~al.(2016{\natexlab{a}})Papernot, McDaniel, Jha,
  Fredrikson, Celik, and Swami]{papernot2016limitations}
Nicolas Papernot, Patrick McDaniel, Somesh Jha, Matt Fredrikson, Z~Berkay
  Celik, and Ananthram Swami.
\newblock The limitations of deep learning in adversarial settings.
\newblock In \emph{2016 IEEE European Symposium on Security and Privacy}, pages
  372--387. IEEE, 2016{\natexlab{a}}.

\bibitem[Carlini and Wagner(2017{\natexlab{a}})]{carlini2017towards}
Nicholas Carlini and David Wagner.
\newblock Towards evaluating the robustness of neural networks.
\newblock In \emph{2017 IEEE Symposium on Security and Privacy}, pages 39--57.
  IEEE, 2017{\natexlab{a}}.

\bibitem[Madry et~al.(2018)Madry, Makelov, Schmidt, Tsipras, and
  Vladu]{madry2018towards}
Aleksander Madry, Aleksandar Makelov, Ludwig Schmidt, Dimitris Tsipras, and
  Adrian Vladu.
\newblock Towards deep learning models resistant to adversarial attacks.
\newblock In \emph{International Conference on Learning Representations}, 2018.

\bibitem[Chen et~al.(2017)Chen, Zhang, Sharma, Yi, and Hsieh]{chen2017zoo}
Pin-Yu Chen, Huan Zhang, Yash Sharma, Jinfeng Yi, and Cho-Jui Hsieh.
\newblock Zoo: Zeroth order optimization based black-box attacks to deep neural
  networks without training substitute models.
\newblock In \emph{Proceedings of the 10th ACM Workshop on Artificial
  Intelligence and Security}, pages 15--26. ACM, 2017.

\bibitem[Ilyas et~al.(2018)Ilyas, Engstrom, Athalye, and Lin]{ilyas2018black}
Andrew Ilyas, Logan Engstrom, Anish Athalye, and Jessy Lin.
\newblock Black-box adversarial attacks with limited queries and information.
\newblock In \emph{International Conference on Machine Learning}, pages
  2142--2151, 2018.

\bibitem[Ilyas et~al.(2019)Ilyas, Engstrom, and Madry]{ilyas2018prior}
Andrew Ilyas, Logan Engstrom, and Aleksander Madry.
\newblock Prior convictions: Black-box adversarial attacks with bandits and
  priors.
\newblock In \emph{International Conference on Learning Representations}, 2019.

\bibitem[Liu et~al.(2017)Liu, Chen, Liu, and Song]{liu2016delving}
Yanpei Liu, Xinyun Chen, Chang Liu, and Dawn Song.
\newblock Delving into transferable adversarial examples and black-box attacks.
\newblock In \emph{Proceedings of the International Conference on Learning
  Representations}, 2017.

\bibitem[Papernot et~al.(2016{\natexlab{b}})Papernot, McDaniel, and
  Goodfellow]{papernot2016transferability}
Nicolas Papernot, Patrick McDaniel, and Ian Goodfellow.
\newblock Transferability in machine learning: from phenomena to black-box
  attacks using adversarial samples.
\newblock \emph{arXiv preprint arXiv:1605.07277}, 2016{\natexlab{b}}.

\bibitem[Papernot et~al.(2017)Papernot, McDaniel, Goodfellow, Jha, Celik, and
  Swami]{papernot2017practical}
Nicolas Papernot, Patrick McDaniel, Ian Goodfellow, Somesh Jha, Z~Berkay Celik,
  and Ananthram Swami.
\newblock Practical black-box attacks against machine learning.
\newblock In \emph{Proceedings of the 2017 ACM on Asia Conference on Computer
  and Communications Security}, pages 506--519. ACM, 2017.

\bibitem[Brendel et~al.(2018)Brendel, Rauber, and
  Bethge]{brendel2018decisionbased}
Wieland Brendel, Jonas Rauber, and Matthias Bethge.
\newblock Decision-based adversarial attacks: Reliable attacks against
  black-box machine learning models.
\newblock In \emph{International Conference on Learning Representations}, 2018.

\bibitem[Brunner et~al.(2018)Brunner, Diehl, Le, and
  Knoll]{brunner2018guessing}
Thomas Brunner, Frederik Diehl, Michael~Truong Le, and Alois Knoll.
\newblock Guessing smart: Biased sampling for efficient black-box adversarial
  attacks.
\newblock \emph{arXiv preprint arXiv:1812.09803}, 2018.

\bibitem[Cheng et~al.(2019)Cheng, Le, Chen, Zhang, Yi, and
  Hsieh]{cheng2018queryefficient}
Minhao Cheng, Thong Le, Pin-Yu Chen, Huan Zhang, JinFeng Yi, and Cho-Jui Hsieh.
\newblock Query-efficient hard-label black-box attack: An optimization-based
  approach.
\newblock In \emph{International Conference on Learning Representations}, 2019.

\bibitem[Tramèr et~al.(2018)Tramèr, Kurakin, Papernot, Goodfellow, Boneh, and
  McDaniel]{tramer2018ensemble}
Florian Tramèr, Alexey Kurakin, Nicolas Papernot, Ian Goodfellow, Dan Boneh,
  and Patrick McDaniel.
\newblock Ensemble adversarial training: Attacks and defenses.
\newblock In \emph{International Conference on Learning Representations}, 2018.

\bibitem[Flaxman et~al.(2005)Flaxman, Kalai, and McMahan]{flaxman2005online}
Abraham~D Flaxman, Adam~Tauman Kalai, and H~Brendan McMahan.
\newblock Online convex optimization in the bandit setting: gradient descent
  without a gradient.
\newblock In \emph{Proceedings of the Sixteenth Annual ACM-SIAM Symposium on
  Discrete Algorithms}, pages 385--394. SIAM, 2005.

\bibitem[Agarwal et~al.(2011)Agarwal, Foster, Hsu, Kakade, and
  Rakhlin]{agarwal2011stochastic}
Alekh Agarwal, Dean~P Foster, Daniel~J Hsu, Sham~M Kakade, and Alexander
  Rakhlin.
\newblock Stochastic convex optimization with bandit feedback.
\newblock In \emph{Advances in Neural Information Processing Systems}, pages
  1035--1043, 2011.

\bibitem[Nesterov and Spokoiny(2017)]{nesterov2017random}
Yurii Nesterov and Vladimir Spokoiny.
\newblock Random gradient-free minimization of convex functions.
\newblock \emph{Foundations of Computational Mathematics}, 17\penalty0
  (2):\penalty0 527--566, 2017.

\bibitem[Duchi et~al.(2015)Duchi, Jordan, Wainwright, and
  Wibisono]{duchi2015optimal}
John~C Duchi, Michael~I Jordan, Martin~J Wainwright, and Andre Wibisono.
\newblock Optimal rates for zero-order convex optimization: The power of two
  function evaluations.
\newblock \emph{IEEE Transactions on Information Theory}, 61\penalty0
  (5):\penalty0 2788--2806, 2015.

\bibitem[Liu et~al.(2018{\natexlab{a}})Liu, Kailkhura, Chen, Ting, Chang, and
  Amini]{liu2018zeroth}
Sijia Liu, Bhavya Kailkhura, Pin-Yu Chen, Paishun Ting, Shiyu Chang, and Lisa
  Amini.
\newblock Zeroth-order stochastic variance reduction for nonconvex
  optimization.
\newblock In \emph{Advances in Neural Information Processing Systems}, pages
  3731--3741, 2018{\natexlab{a}}.

\bibitem[Kincaid et~al.(2009)Kincaid, Kincaid, and
  Cheney]{kincaid2009numerical}
David Kincaid, David~Ronald Kincaid, and Elliott~Ward Cheney.
\newblock \emph{Numerical {A}nalysis: {M}athematics of {S}cientific
  {C}omputing}, volume~2.
\newblock American Mathematical Soc., 2009.

\bibitem[Ledoux(2001)]{ledoux2001concentration}
Michel Ledoux.
\newblock \emph{The {C}oncentration of {M}easure {P}henomenon}.
\newblock Number~89. American Mathematical Soc., 2001.

\bibitem[Papernot et~al.(2018)Papernot, Faghri, Carlini, Goodfellow, Feinman,
  Kurakin, Xie, Sharma, Brown, Roy, Matyasko, Behzadan, Hambardzumyan, Zhang,
  Juang, Li, Sheatsley, Garg, Uesato, Gierke, Dong, Berthelot, Hendricks,
  Rauber, and Long]{papernot2018cleverhans}
Nicolas Papernot, Fartash Faghri, Nicholas Carlini, Ian Goodfellow, Reuben
  Feinman, Alexey Kurakin, Cihang Xie, Yash Sharma, Tom Brown, Aurko Roy,
  Alexander Matyasko, Vahid Behzadan, Karen Hambardzumyan, Zhishuai Zhang,
  Yi-Lin Juang, Zhi Li, Ryan Sheatsley, Abhibhav Garg, Jonathan Uesato, Willi
  Gierke, Yinpeng Dong, David Berthelot, Paul Hendricks, Jonas Rauber, and
  Rujun Long.
\newblock Technical report on the cleverhans v2.1.0 adversarial examples
  library.
\newblock \emph{arXiv preprint arXiv:1610.00768}, 2018.

\bibitem[Rauber et~al.(2017)Rauber, Brendel, and Bethge]{rauber2017foolbox}
Jonas Rauber, Wieland Brendel, and Matthias Bethge.
\newblock Foolbox: A python toolbox to benchmark the robustness of machine
  learning models.
\newblock \emph{arXiv preprint arXiv:1707.04131}, 2017.

\bibitem[Krizhevsky(2009)]{krizhevsky2009learning}
Alex Krizhevsky.
\newblock Learning multiple layers of features from tiny images.
\newblock Technical report, Citeseer, 2009.

\bibitem[Deng et~al.(2009)Deng, Dong, Socher, Li, Li, and
  Fei-Fei]{imagenet_cvpr09}
J.~Deng, W.~Dong, R.~Socher, L.-J. Li, K.~Li, and L.~Fei-Fei.
\newblock {ImageNet: A Large-Scale Hierarchical Image Database}.
\newblock In \emph{Proceedings of the IEEE Conference on Computer Vision and
  Pattern Recognition}, 2009.

\bibitem[Chollet et~al.(2015)]{chollet2015keras}
Fran\c{c}ois Chollet et~al.
\newblock Keras.
\newblock \url{https://keras.io}, 2015.

\bibitem[Metzen et~al.(2017)Metzen, Genewein, Fischer, and
  Bischoff]{metzen2017detecting}
Jan~Hendrik Metzen, Tim Genewein, Volker Fischer, and Bastian Bischoff.
\newblock On detecting adversarial perturbations.
\newblock In \emph{International Conference on Learning Representations}, 2017.

\bibitem[He et~al.(2016)He, Zhang, Ren, and Sun]{he2016identity}
Kaiming He, Xiangyu Zhang, Shaoqing Ren, and Jian Sun.
\newblock Identity mappings in deep residual networks.
\newblock In \emph{European Conference on Computer Vision}, pages 630--645.
  Springer, 2016.

\bibitem[Huang et~al.(2017)Huang, Liu, Van Der~Maaten, and
  Weinberger]{huang2017densely}
Gao Huang, Zhuang Liu, Laurens Van Der~Maaten, and Kilian~Q Weinberger.
\newblock Densely connected convolutional networks.
\newblock In \emph{Proceedings of the IEEE Conference on Computer Vision and
  Pattern Recognition}, pages 4700--4708, 2017.

\bibitem[Papernot et~al.(2016{\natexlab{c}})Papernot, McDaniel, Wu, Jha, and
  Swami]{papernot2016distillation}
Nicolas Papernot, Patrick McDaniel, Xi~Wu, Somesh Jha, and Ananthram Swami.
\newblock Distillation as a defense to adversarial perturbations against deep
  neural networks.
\newblock In \emph{2016 IEEE Symposium on Security and Privacy}, pages
  582--597. IEEE, 2016{\natexlab{c}}.

\bibitem[Athalye et~al.(2018)Athalye, Carlini, and
  Wagner]{athalye2018obfuscated}
Anish Athalye, Nicholas Carlini, and David Wagner.
\newblock Obfuscated gradients give a false sense of security: Circumventing
  defenses to adversarial examples.
\newblock In \emph{International Conference on Machine Learning}, pages
  274--283, 2018.

\bibitem[Cao and Gong(2017)]{cao2017mitigating}
Xiaoyu Cao and Neil~Zhenqiang Gong.
\newblock Mitigating evasion attacks to deep neural networks via region-based
  classification.
\newblock In \emph{Proceedings of the 33rd Annual Computer Security
  Applications Conference}, pages 278--287. ACM, 2017.

\bibitem[Liu et~al.(2018{\natexlab{b}})Liu, Cheng, Zhang, and
  Hsieh]{liu2018towards}
Xuanqing Liu, Minhao Cheng, Huan Zhang, and Cho-Jui Hsieh.
\newblock Towards robust neural networks via random self-ensemble.
\newblock In \emph{Proceedings of the European Conference on Computer Vision
  (ECCV)}, pages 369--385, 2018{\natexlab{b}}.

\bibitem[Dhillon et~al.(2018)Dhillon, Azizzadenesheli, Bernstein, Kossaifi,
  Khanna, Lipton, and Anandkumar]{guneet2018stochastic}
Guneet~S. Dhillon, Kamyar Azizzadenesheli, Jeremy~D. Bernstein, Jean Kossaifi,
  Aran Khanna, Zachary~C. Lipton, and Animashree Anandkumar.
\newblock Stochastic activation pruning for robust adversarial defense.
\newblock In \emph{International Conference on Learning Representations}, 2018.

\bibitem[Cohen et~al.(2019)Cohen, Rosenfeld, and Kolter]{cohen2019certified}
Jeremy Cohen, Elan Rosenfeld, and Zico Kolter.
\newblock Certified adversarial robustness via randomized smoothing.
\newblock In \emph{International Conference on Machine Learning}, pages
  1310--1320, 2019.

\bibitem[Xie et~al.(2018)Xie, Wang, Zhang, Ren, and Yuille]{xie2018mitigating}
Cihang Xie, Jianyu Wang, Zhishuai Zhang, Zhou Ren, and Alan Yuille.
\newblock Mitigating adversarial effects through randomization.
\newblock In \emph{International Conference on Learning Representations}, 2018.

\bibitem[Carlini and Wagner(2017{\natexlab{b}})]{carlini2017adversarial}
Nicholas Carlini and David Wagner.
\newblock Adversarial examples are not easily detected: Bypassing ten detection
  methods.
\newblock In \emph{Proceedings of the 10th ACM Workshop on Artificial
  Intelligence and Security}, pages 3--14. ACM, 2017{\natexlab{b}}.

\bibitem[Kurakin et~al.(2018)Kurakin, Goodfellow, and
  Bengio]{kurakin2018adversarial}
Alexey Kurakin, Ian~J Goodfellow, and Samy Bengio.
\newblock Adversarial examples in the physical world.
\newblock In \emph{Artificial Intelligence Safety and Security}, pages 99--112.
  Chapman and Hall/CRC, 2018.

\bibitem[Pedregosa et~al.(2011)Pedregosa, Varoquaux, Gramfort, Michel, Thirion,
  Grisel, Blondel, Prettenhofer, Weiss, Dubourg, Vanderplas, Passos,
  Cournapeau, Brucher, Perrot, and Duchesnay]{scikit-learn}
F.~Pedregosa, G.~Varoquaux, A.~Gramfort, V.~Michel, B.~Thirion, O.~Grisel,
  M.~Blondel, P.~Prettenhofer, R.~Weiss, V.~Dubourg, J.~Vanderplas, A.~Passos,
  D.~Cournapeau, M.~Brucher, M.~Perrot, and E.~Duchesnay.
\newblock Scikit-learn: Machine learning in {P}ython.
\newblock \emph{Journal of Machine Learning Research}, 12:\penalty0 2825--2830,
  2011.

\end{thebibliography}
\bibliographystyle{unsrtnat}
\end{small}


\appendices

\section{Proofs}

For notational simplicity, we use the shorthand $S \equiv S_\xorig$
throughout the proofs.


\subsection{Proof of Theorem~\ref{thm:convergence}}
\label{app:thm_convergence}

We denote $\tau_t := \interstep_t
/ \|\grad\|_2$, so that the update~\eqref{eq:update_l2} at iterate $t$
can be rewritten as
\begin{align}
\label{eq:update_l2_new}
x_{t+1} = \alpha_t \xorig + (1-\alpha_t) (x_t + \tau_t \nabla S(x_t)).
\end{align}
Let the step size choice $\interstep_t = \eta_t \|x_t -
\xorig\|$ with $\eta_t\defn t^{-q}$, we have $\tau_t = \eta_t \frac {\|x_t - \xorig\|}{\|\nabla
  S(x_t)\|}$.

The squared distance ratio is
\begin{align}
\label{eq:dist_ratio}
\frac{\|x_{t+1}-\xorig\|_2^2 }{\|x_t - \xorig\|_2^2} = 
\frac{\|(1-\alpha)(\tau_t \grad + x_t - \xorig)\|_2^2}{\|x_t - \xorig\|_2^2}.
\end{align}
By a second-order Taylor series, we have
\begin{align}
\label{eq:nonlinear}
0 = \inprod{\grad}{x_{t+1} - x_t} + \frac{1}{2} (x_{t+1} - x_t)^T H_t
(x_{t+1} - x_t),
\end{align}
where $H_t = \nabla^2 S(\beta x_{t+1} + (1 - \beta) x_t)$ for some
$\beta \in [0, 1]$. Plugging equation~\eqref{eq:update_l2_new} into
equation~\eqref{eq:nonlinear} yields
\begin{align}
&\inprod{\grad}{-\alpha v_t +\tau_t \grad} + \nonumber\\
&~~\frac{1}{2} (- \alpha v_t
+\tau_t \grad)^T H_t(- \alpha v_t +\tau_t \grad) = 0,
\end{align}
where we define $v_t \defn x_t - \xorig + \tau_t \grad$. This can be
rewritten as a quadratic equation with respect to $\alpha$:
\begin{align}
&v_t^TH_tv_t \alpha^2 - 2\grad^T (  I + \tau_t H_t)v_t\alpha \nonumber\\
&~~+ \grad^T(\tau_t^2 H_t + 2\tau_t I) \grad = 0.
\end{align}
Solving for $\alpha$ yields
\begin{align}
\alpha & \geq \frac{\grad^T(\tau_t^2 H_t+2\tau_t I)\grad} {2\grad^T (I+\tau_t H_t)v_t}.
\end{align}
In order to
simplify the notation, define $\shortgrad \defn \grad$ and $\diff
\defn x_t - \xorig$. Hence, we have 
\begin{align*}
(1-\alpha)^2
        & \leq \Big ( 
        \frac {r_t + \eta_t \cdot \frac 3 2 L\frac {\|\diff\|_2}{\|\nabla_t\|_2}}{r_t + \eta_t \cdot (1 + \frac 3 2 L\frac {\|\diff\|_2}{\|\nabla_t\|_2})}
        \Big)^2,
\end{align*}
where 
\begin{align}
r_t = \frac{\langle x_t - \xorig , \nabla S(x_t) \rangle} {\|x_t - \xorig\|_2\|\nabla S(x_t)\|_2} = \frac{\langle \diff , \nabla_t \rangle} {\|\diff\|_2\|\nabla_t\|_2}.
\end{align}
Let $\kappa_t: = \frac 3 2 L\frac {\|\diff\|_2}{\|\nabla_t\|_2}$. Then $\kappa_t$ is bounded when $\|\nabla_t\|_2\geq \tilde C$  and $q>\frac 1 2$. Equation~\eqref{eq:dist_ratio} and the bound on $(1-\alpha)^2$ yield
\begin{align}
\frac{\|x_{t+1}-\xorig\|_2^2 }{\|x_t-\xorig\|_2^2} &\leq 
 \Big (\frac {r_t + \eta_t \kappa_t}{r_t + \eta_t (1 + \kappa_t)}\Big)^2
 \cdot (\eta_t^2 + 2\eta_t r_t + 1) \label{eq:ratio_bound}.
\end{align}


Define $\theta_t \defn \Big (\frac {r_t + \eta_t \kappa_t}{r_t + \eta_t (1 + \kappa_t)}\Big)^2 \cdot (\eta_t^2 + 2\eta_t r_t + 1)$. We analyze $\theta_t$ in the following two different cases: $r_t < \eta_t$ and $r_t\geq \eta_t$. In the first case, we have 
\begin{align}
\theta_{t} 
&\leq \Big (\frac {1 + \kappa_t}{1 + (1 + \kappa_t)}\Big)^2 \cdot (\eta_t^2 + 2\eta_t^2 + 1).
\end{align}
As long as $\eta_t\to 0$ as $t\to\infty$, there exists a positive constant $c_2>0$ such that $\theta_t<1-c_2$ for $t$ large enough. 

In the second case, we have $r_t\geq \eta_t$. Define $\lambda_t:=\frac{\eta_t}{r_t}\leq 1$. We bound $\theta_t$ by
\begin{align}
\theta_t
&= \frac {(1 + 2\lambda_t \kappa_t + \lambda_t^2 \kappa_t^2)(\eta_t^2 + 2\eta_t r_t + 1)}
{1 + 2\lambda_t (1 + \kappa_t) + \lambda_t^2 (1 + \kappa_t)^2} \nonumber
\\
&\leq\frac{1 + 2\lambda_t \kappa_t + \lambda_t^2 \kappa_t^2 + 2\lambda_tr_t^2}
{1 + 2\lambda_t \kappa_t + \lambda_t^2 \kappa_t^2 + 2\lambda_t} +\nonumber\\
&~~~~  
\eta_t^2(4\kappa_t + (1 + \lambda_t \kappa_t)^2 + 2\lambda_t \kappa_t^2)\nonumber\\
&\leq 
1 - \frac{2\lambda_t(1-r_t^2)}
{1 + 2\lambda_t \kappa_t + \lambda_t^2 \kappa_t^2 + 2\lambda_t}
+ c\eta_t^2\nonumber\\
&\leq 1 - c_1\lambda_t (1 - r_t^2) + c_2\eta_t^2,\nonumber
\end{align}
where $c_1,c_2$ are fixed constants. As the product of $\theta_t$ over $t$ is positive, we have
\begin{align}\label{eq:theta_}
\sum_{t=1}^\infty \log \theta_t = \log \Pi_{t=1}^\infty \theta_t >-\infty.
\end{align}
Then we have that there are at most a finite number of $t$ that falls in the first case, $r_t<\eta_t$. In the second case, Equation~\eqref{eq:theta_} is equivalent to 
$$\sum_{t=1}^\infty c_1\eta_t \frac{1 - r_t^2}{r_t}-c_2\eta_t^2 <\infty,$$
which implies $c_1\eta_t \frac{1 - r_t^2}{r_t}-c_2\eta_t^2 = o(t^{-1})$. When $\eta_t = t^{-q}$ for some constant $\frac 1 2 <q<1$, we have 
$$
\frac{1 - r_t^2}{r_t} = o(t^{q-1}). 
$$
Hence we have $1 - r_t = o(t^{q-1})$. 



\subsection{Proof of Theorem~\ref{thm:unbiased}}\label{app:thm1}
Let $u$ be a random vector uniformly distributed on the sphere. By Taylor's theorem, for any $\delta\in(0,1)$, we have 
\begin{align}
S(x_t+\delta u) 
&= \delta \nabla S(x_t)^Tu +\frac 1 2 \delta^2u^T\nabla^2S(x')u.\label{eq:taylor2}
\end{align}
for some $x'$ on the line between $x_t$ and $x_t+\delta u$, where we have made use of the fact that $S(x_t)=0$. As the function $S$ has Lipschitz gradients,
we can bound the second-order term as
\begin{align}
|\frac 1 2 \delta^2u^T\nabla^2S(x')u|\leq \frac 1 2 L\delta^2.
\end{align}
Let $w\defn \frac 1 2 L \delta$. By the Taylor expansion and the bound on the second-order term by eigenvalues, when $\nabla S(x_t)^T u>w$, we have
\begin{align*}
S(x_t+\delta u) &\geq \delta \nabla S(x_t)^Tu +\frac 1 2 \delta^2u^T\nabla^2S(x')u\\
&\geq \delta (\nabla S(x_t)^Tu  - \frac{1}{2}L\delta) > 0.
\end{align*}
Similarly, we have $S(x_t+\delta u) < 0$ when $\nabla S(x_t)^T u<-w$. Therefore, we have
\begin{align*}
\phi_x(x_t+\delta u) = \begin{cases}
1 \text{ if } \nabla S(x_t)^T u>w,\\
-1 \text{ if } \nabla S(x_t)^T u<-w.
\end{cases}
\end{align*}
We expand the vector $\nabla S(x_t)$ to an orthogonal bases in $\mathbb R^d$: $v_1 = \nabla S(x_t)/\|\nabla S(x_t)\|_2, v_2,\dots,v_d$. The random vector $u$ can be expressed as 
$u = \sum_{i=1}^d \beta_i v_i$,
where $\beta$ is uniformly distributed on the sphere.
Denote the upper cap as $E_1\defn\{\nabla S(x_t)^T u > w\}$, the annulus as $E_2\defn\{|\nabla S(x_t)^T u|<w\}$, and the lower cap as $E_3\defn\{\nabla S(x_t)^T u<-w\}$. Let $p\defn \Prob(E_2)$ be the probability of event $E_2$. Thus we have $\Prob(E_1) = \Prob(E_3) = (1-p)/2$. By symmetry, for any $i\neq 1$, we have 
\begin{align*}
\Exp[\beta_i \mid E_1] = \Exp[\beta_i \mid E_3]=0.
\end{align*} 
Therefore, the expected value of the estimator is
\begin{align*}
\Exp[\phi_x(x_t + \delta u)u]
&= p\cdot \big(\Exp[\phi_x(x_t + \delta u)u \mid E_2]  \\
&~~~~~~- \frac{1}{2}\Exp[\beta_1v_1\mid E_1] - \frac{1}{2}\Exp[-\beta_1 v_1 \mid E_3]\big) \\
&~~~~~~~~+ \Exp[\beta_1v_1\mid E_1] + \Exp[-\beta_1 v_1 \mid E_3] 
\end{align*}
Exploiting the above derivation, we can bound the difference between $\Exp[|\beta_1|v_1] = \frac{\Exp|\beta_1|}{\|\nabla S(x_t)\|_2}\nabla S(x_t)$ and $\Exp[\phi_x(x_t + \delta u)u]$:
\begin{align*}
\|\Exp[\phi_x(x_t + \delta u)u] - \Exp[|\beta_1|v_1]\|_2&\leq 2p + p = 3p,
\end{align*}
which yields 
\begin{align}\label{eq:cosine_intermediate}
\cos \angle \left( \Exp[\phi_x(x_t + \delta u)u], \nabla S(x_t) \right)\geq 
1 - \frac 1 2 \Big(\frac {3p}{\Exp|\beta_1|}\Big)^2.
\end{align}
We can bound $p$ by observing that $\langle \frac{\nabla S(x_t)}{\|\nabla S(x_t)\|_2}, u\rangle^2$ is a Beta distribution $\mathcal B(\frac 1 2, \frac {d-1} 2)$:
\begin{align*}
p&=\Prob \Big(\mathcal \langle \frac{\nabla S(x_t)}{\|\nabla S(x_t)\|_2}, u\rangle^2\leq 
\frac {w^2} {\|\nabla S(x_t)\|_2^2}\Big) \\
&\leq \frac {2w}{\mathcal B(\frac 1 2, \frac {d-1} 2)\|\nabla S(x_t)\|_2}. 
\end{align*}
Plugging into Equation~\eqref{eq:cosine_intermediate}, we get 
\begin{align*}
\lefteqn{\cos \angle \left( \Exp[\phi_x(x_t + \delta u)u], \nabla S(x_t) \right)} \\
&\geq 
1 - \frac {18w^2}{(\Exp|\beta_1|)^2\mathcal B(\frac 1 2, \frac {d-1} 2)^2\|\nabla S(x_t)\|_2^2}\\
&= 1 - \frac {9L^2\delta^2(d-1)^2}{8\|\nabla S(x_t)\|_2^2}.
\end{align*}
We also observe that 
\begin{align*}
\mathbb E \widetilde{\nabla S}(x_t, \delta) = \Exp[\phi_x(x_t + \delta u)u].
\end{align*}
As a consequence, we have established
\begin{align*}
\cos \angle \left( \Exp[\widetilde{\nabla S}(x_t, \delta)], \nabla S(x_t) \right)
\geq 1 - \frac {9L^2\delta^2(d-1)^2}{8\|\nabla S(x_t)\|_2^2}.
\end{align*}
Taking $\delta\to 0$, we get 
\begin{align*}
\lim_{\delta\to 0}
\mathbb \cos \angle \left( \Exp[\widetilde{\nabla S}(x_t, \delta)], \nabla S(x_t) \right) = 1.
\end{align*}

\subsection{Proof of Theorem~\ref{thm:var}}\label{app:thm2}

\begin{proof}
For notational simplicity, we denote $\xi_b\defn \phi_x(x_t + \delta
u_b)$, and $\bar\xi =\frac{1}{B}\sum_{b=1}^B \xi_b
= \overline{\phi_x}$. We use $\xi,u$ to denote i.i.d. copies of
$\xi_b$ and $u_b$ respectively. By exploiting independence of $u_a,u_b$ and independence of $\xi_a u_a, \xi_b u_b$, the variance of the estimate with the baseline can be expressed as
\begin{align}
\lefteqn{\text{Var}(\widehat{\nabla S}(x_t, \delta))} \nonumber\\
 &= \frac{1}{(B-1)^2}\sum_{a=1}^B \Big(\Exp\Big\| \xi_a u_a - \Exp [\xi u]\Big\|_2^2 - 
2 \Exp [\bar\xi\xi_a] + 
 \nonumber\\
&~~~~~~
\Exp\bar\xi^2 +(\frac{2}{B}-\frac 1 {B^2}) \|\Exp [\xi u] \|^2\Big) + \frac{\|\Exp\xi u\|_2^2}{B(B-1)} \nonumber
\\
&= \frac{B^2\text{Var}(\widetilde{\nabla S}(x_t, \delta))}{(B-1)^2} - 
\frac{B\Exp[\bar\xi^2]}{(B-1)^2}  + \frac{(3B-2)\|\Exp [\xi u]\|_2^2}{B(B-1)^2} \nonumber\\
&\leq \frac{B^2\text{Var}(\widetilde{\nabla S}(x_t, \delta))}{(B-1)^2} - 
\frac{B\Exp[\bar\xi^2]}{(B-1)^2}  + \frac{3B-2}{B(B-1)^2} \label{eq:variance_expression}.
\end{align}
The middle term can be expanded as
\begin{align*}
- \frac{B}{(B-1)^2} \Exp[\bar\xi^2] &= - \frac{1}{(B-1)^2} - \frac{4}{B-1}(\Exp \xi-\frac 1 2)^2.
\end{align*}
Plugging into Equation~\eqref{eq:variance_expression}, we get
\begin{align*}
\text{Var}(\widehat{\nabla S}(x_t, \delta))
&= \text{Var}(\widetilde{\nabla S}(x_t, \delta)) \Big\{1 + \frac{2B-1}{(B-1)^2} - \\
&~~~~~~\frac {2}{\sigma^2 (B-1)}\big(2B(\Exp[\xi]-\frac{1}{2})^2-1\big)\Big\}.
\end{align*}
When $\Exp[\xi]$ satisfies $(\Exp[\xi]-\frac 1 2)^2 > \frac{1}{2B}(1 + \frac{2B-1}{2B-2}\sigma^2)$, we have
\begin{align*}
\frac{2B-1}{(B-1)^2} < \frac {2}{\sigma^2 (B-1)}(2B(\Exp[\xi]-\frac{1}{2})^2-1),
\end{align*}
which implies $\text{Var}(\widehat{\nabla S}(x_t, \delta))
< \text{Var}(\widetilde{\nabla S}(x_t, \delta))$.
\end{proof}

\section{Sensitivity analysis}
\label{app:sensitivity}
In this section, we carry out experiments to evaluate the hyper-parameters suggested by our theoretical analysis. We use a $20$-layer ResNet~\cite{he2016identity} trained over
CIFAR-10~\cite{krizhevsky2009learning}. We run the $\ell_2$-optimized HopSkipJumpAttack over a subset of randomly sampled images.
{\paragraph{Choice of step size} We compare several schemes of choosing step size at each step. The first scheme is suggested by Theorem~\ref{thm:convergence}: at the $t$-th step, we set $\xi_t = \|x_t - \xorig\|_2/\sqrt{t}$, which we call ``Scale with Distance (Sqrt. Decay).'' We include the other two scales which scale with distance, ``Scale with Distance (Linear Decay)'' with $\xi_t = \|x_t - \xorig\|_2/t$ and ``Scale with Distance (No Decay)'' with $\xi_t = \|x_t - \xorig\|_2$. We then include ``Grid Search,'' which searchs step sizes over a log-scale grid, and chooses the step size that best controls the distance with the original sample after projecting the updated sample back to the boundary via binary search. Finally, we include constant stepsizes at $\xi_t = 0.01,0.1,1.0$. For all schemes, we always use geometric progression to decrease the step size by half until $\phi_\xorig(\tilde x_t)= 1$ before the next binary search step. }

\begin{figure}[!bt]
\centering
\includegraphics[width=1.0\linewidth]{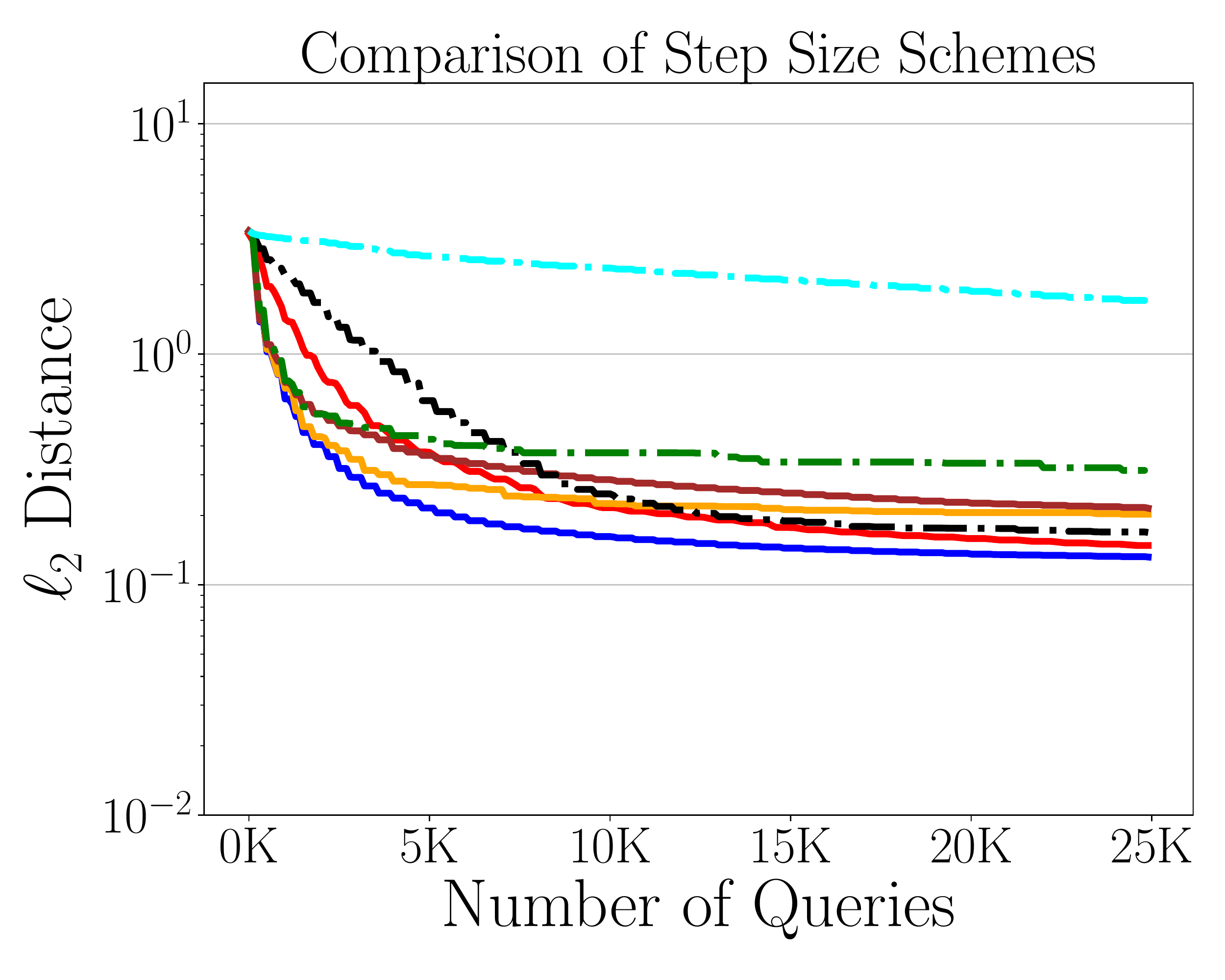} 
\includegraphics[width=1.0\linewidth]{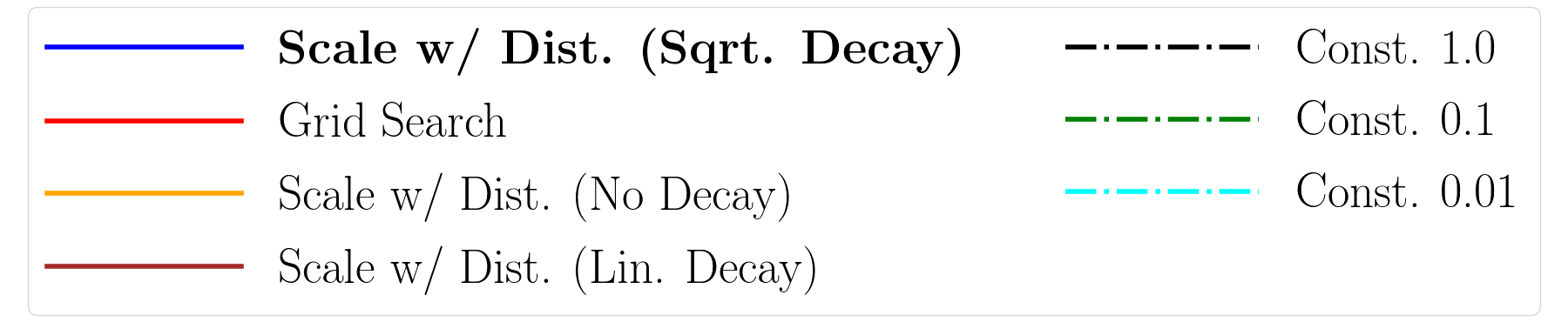} 
\caption{{Comparison of various choices of step size.}}
\label{fig:sensitivity1}
\end{figure}

{Figure~\ref{fig:sensitivity1} plots the median distance against the number of queries for all schemes. We observe that the scheme suggested by Theorem~\ref{thm:convergence} achieves the best performance in this experiment. Grid search costs extra query budget initially but eventually achieves a comparable convergence rate. When the step size scales with the distance but with inappropriately chosen decay, the algorithm converges slightly slower. The performance of the algorithm suffers from a constant step size.}

\paragraph{Choice of perturbation size and introduction of baseline} We now study the effectiveness of {the proposed perturbation size and baseline} for estimating gradient direction when the sample deviates from the boundary. In particular, we focus on the choice of $\delta_t$ and the introduction of baseline analyzed in Section~\ref{sec:gradient}. Gradient direction estimation is carried out at perturbed images at the $i$th iteration, for $i=10,20,30,40,50,60$. We use the {cosine of the angle} between the gradient-direction estimate and the truth gradient of the model as a metric.

\begin{figure}[!bt]
\centering
\includegraphics[width=1.0\linewidth]{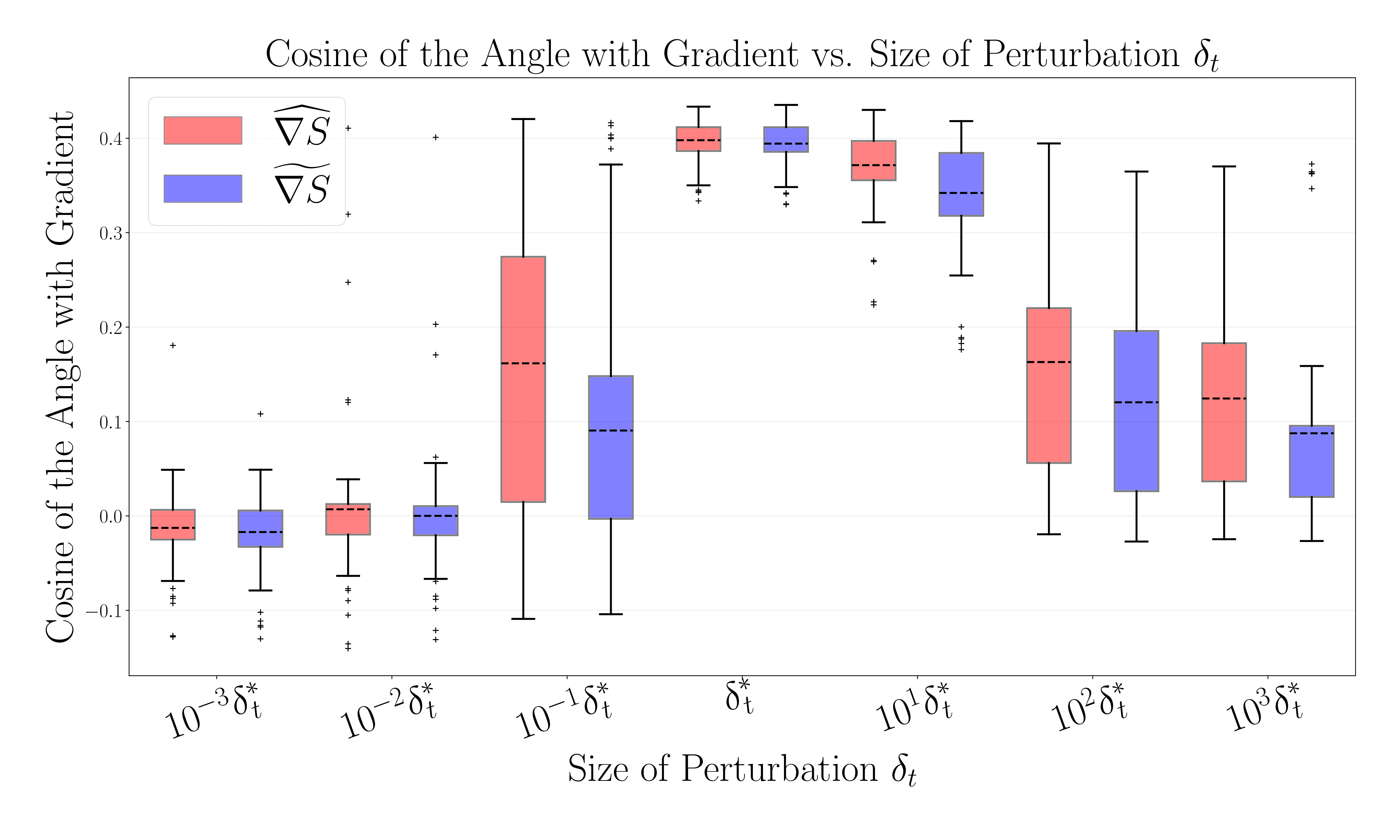}
\caption{{Box plots of the cosine of the angle between the proposed estimates and the true gradient.}}
\label{fig:sensitivity2}
\end{figure}

Figure~\ref{fig:sensitivity2} shows the box plots of two gradient-direction estimates as $\delta_t$ varies among
$0.01\delta_t^*,0.1\delta_t^*,\delta_t^*,10\delta_t^*,100\delta_t^*$,
where $\delta_t^* = 10 \sqrt{d}\theta\|\tilde x_{t-1} - \xorig\|_2$ is our
proposed choice. We observe that our proposed choice of $\delta_t$
yields the highest {cosine of the angle} on average. Also, the baseline in
$\widehat{\nabla S}$ further improves the performance, in particular
when $\delta_t$ is not chosen optimally so that there is severe unevenness in the distribution of perturbed images.

\section{Model without gradients}\label{app:nondiff}
{In this section, we evaluate HopSkipJumpAttack on a model without gradients. We aim to show HopSkipJumpAttack is able to craft adversarial examples under weaker conditions, such as
non-differentiable models, or even discontinuous input transform.} 

{Concretely, we implement input binarization followed by a random forest on MNIST. Binarization transforms an input image to an array of $\{0, 1\}$, but transforming all pixels larger than a given threshold to $1$, and all pixels smaller than the threshold to $0$. The algorithm for training random forests applies bootstrap aggregating to tree learners. We implement the random forest with default parameters in scikit-learn~\cite{scikit-learn}, using the Gini impurity as split criterion. For each split, $\sqrt{d}$ randomly selected features are used, where $d=28\times28$ is the number of pixels. We evaluate two random forests with different thresholds for binarization: $0.1$ and $0.5$. With the first threshold, the model achieves the highest accuracy, $96\%$, on natural test data. The second threshold yields the most robust performance under adversarial perturbation, with accuracy $94.5\%$ on natural test data.}

{For both Boundary Attack and HopSkipJumpAttack, we adopt the same initialization and hyper-parameters as in Section~\ref{sec:efficiency}. 
The original image (with real values) is used as input to both attacks for model queries. When an image is fed into the model by the attacker, the model processes the image with binarization first, followed by the random forest. Such a design preserves the black-box assumption for decision-based attacks. We only focus on untargeted $\ell_2$ attack here. Note that over $91\%$ of the pixels on MNIST are either greater than $0.9$ or less than $0.1$, and thus require a perturbation of size at least $0.4$ to change their outputs after being thresholded by $0.5$. This fact makes $\ell_\infty$ perturbation inappropriate for crafting adversarial examples.}

\begin{figure}[!bt]
\centering
\includegraphics[width=0.48\linewidth]{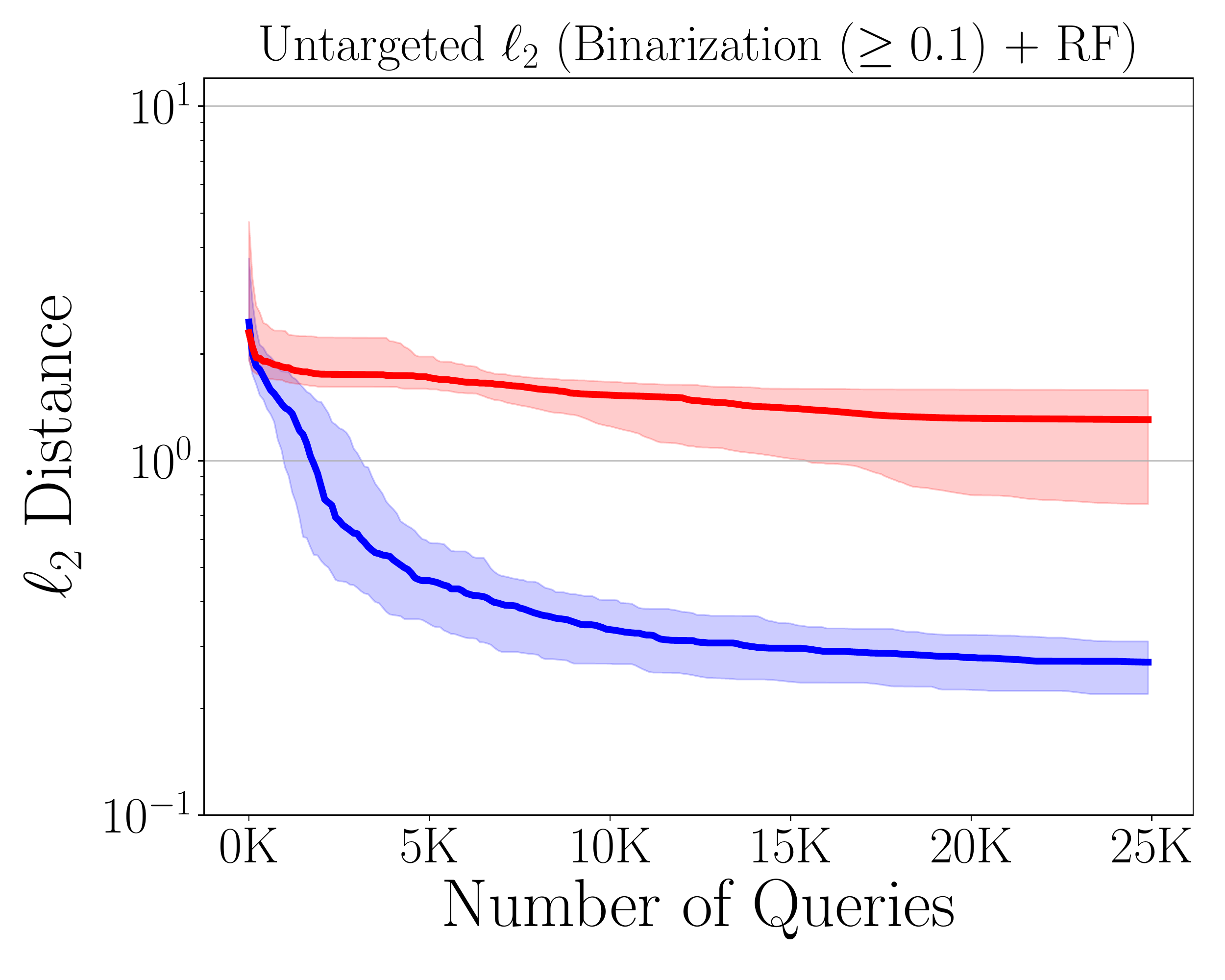} 
\includegraphics[width=0.48\linewidth]{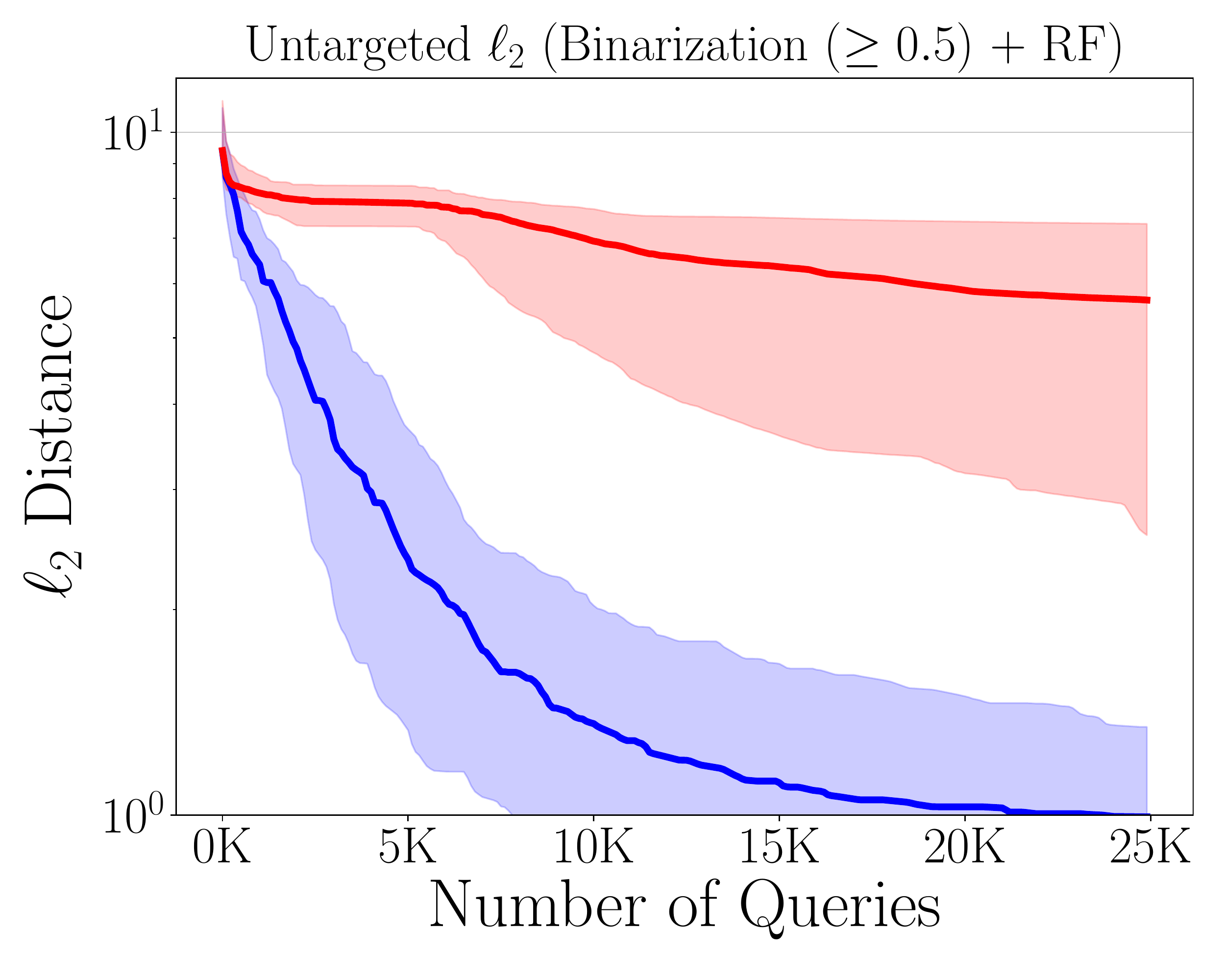}
\includegraphics[width=0.98\linewidth]{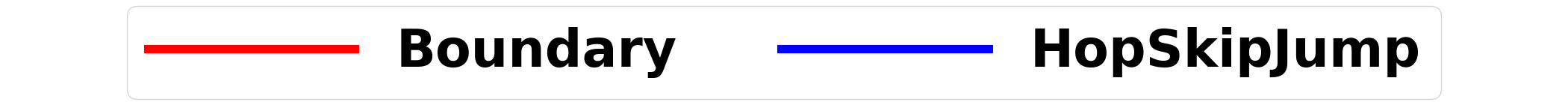}
\caption{{Median $\ell_2$ distance versus number of model queries on MNIST with binarization + random forest. The threshold of binarization is set to be $0.1$ and $0.5$ respectively.}}
\label{fig:discrete}
\end{figure} 
\begin{figure}[!bt]
\includegraphics[width=0.48\linewidth]{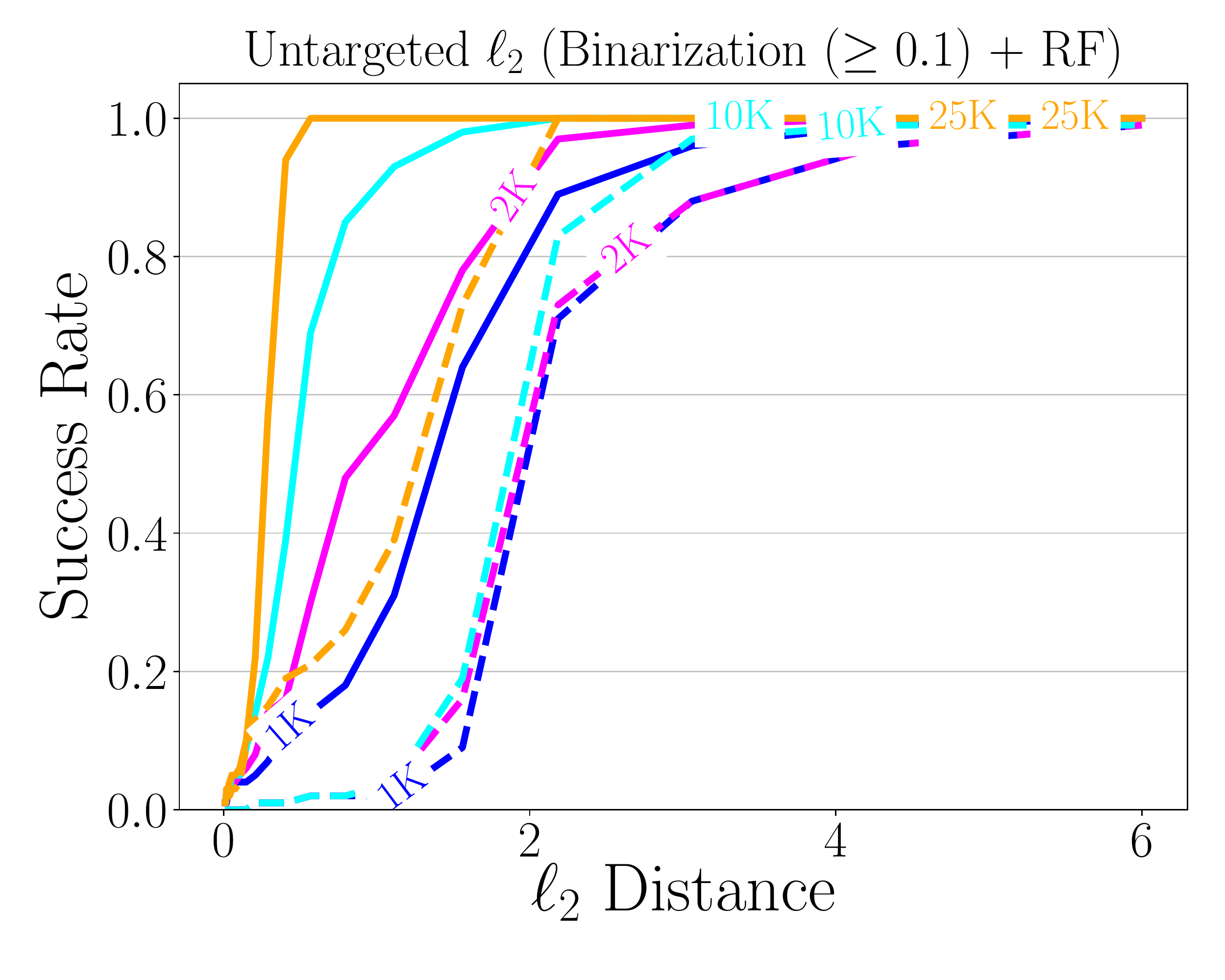}  
\includegraphics[width=0.48\linewidth]{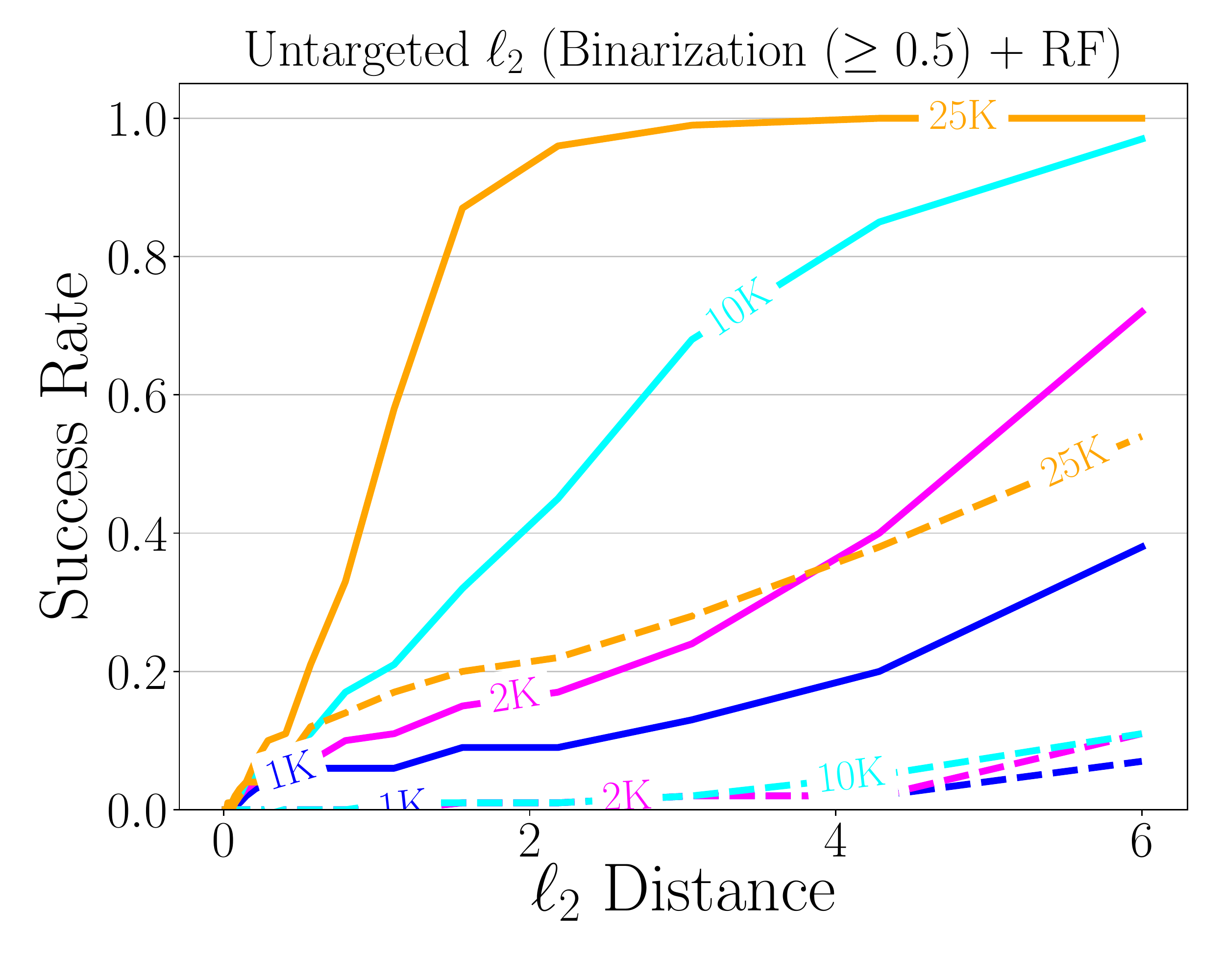} 
\includegraphics[width=0.98\linewidth]{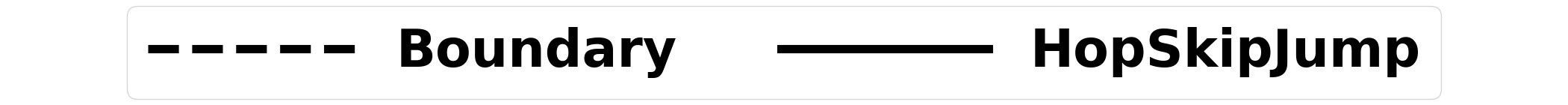}
\caption{{Success rate versus distance threshold on MNIST with binarization + random forest. The threshold of binarization is set to be $0.1$ and $0.5$ respectively.}}
\label{fig:discrete2}
\end{figure}

{Figure~\ref{fig:discrete} shows the median distance (on a log scale) against the queries, with the first and third quartiles used as lower and upper error bars. Figure~\ref{fig:discrete2} shows the success rate against the distance threshold. }

{When the threshold is set to be $0.1$, the random forest with binarization becomes extremely vulnerable to adversarial examples. Around $96\%$ adversarial examples fall into the size-$3$ $\ell_2$-neighborhood of the respective original examples with 1K model queries of HopSkipJumpAttack. The vulnerability is caused by the ease of activating pixels through increasing the strength by $0.1$. It also indicates HopSkipJumpAttack and Boundary Attack are able to craft adversarial examples without smooth decision boundaries.}

{When the threshold is set to be $0.5$, we have a more robust model. A median $\ell_2$distance of $3$ is achieved by HopSkipJumpAttack through 3K model queries. It takes 25K queries to achieve $99\%$ success rate at an $\ell_2$ distance of $3$ for HopSkipJumpAttack. On the other hand, we observe that Boundary Attack only achieves a median distance of $5$ even with 25K model queries. This might result from the inefficiency in spending queries on random walk instead of ``gradient direction'' estimation step in HopSkipJumpAttack. We remark that the concept of ``gradient direction'' requires an alternative definition in the current setting, such as a formulation via subgradients.}


\end{document}